\documentclass[%
paper=letter,%
pagesize,%
numbers=noendperiod,%
captions=nooneline,%
fontsize=11pt,%
DIV=14
]{scrartcl}

\usepackage[left=1in, right=1in, bottom=1in, top=1in, footskip=11mm]{geometry}%

\usepackage[T1]{fontenc}%
\usepackage{lmodern}%
\usepackage[american]{babel}%
\usepackage{microtype}%

\usepackage[
hyperref,%
table%
]{xcolor}%

\usepackage{scrlayer-scrpage}%

\usepackage{eso-pic}%
\usepackage{rotating}%
\usepackage{amsmath}%
\usepackage{mathtools}%
\usepackage{amssymb}%
\usepackage{amsthm}%
\usepackage{etoolbox}%
\usepackage{bm}%
\usepackage{bbm}%
\usepackage{enumitem}%
\usepackage{graphicx}%
\usepackage{siunitx} %
\usepackage{grffile}%
\usepackage{tikz}%
\usetikzlibrary{shapes, decorations.pathreplacing}
\usepackage{wrapfig}%
\usepackage{tabularx}%
\usepackage{bbm}
\usepackage{siunitx}%
\usepackage{booktabs}%
\usepackage{multirow}%
\usepackage{fancyvrb}%
\usepackage{listings}%
\usepackage{csquotes}%
\usepackage[
style=authoryear,%
dashed=false,%
hyperref=true,%
useprefix=true,%
maxnames=2,%
maxbibnames=6,%
uniquename=false,%
seconds=true,%
date=iso,%
urldate=iso%
]{biblatex}%
\usepackage[
hypertexnames=false,%
setpagesize=false,%
pdfborder={0 0 0},%
pdfstartview=Fit,%
bookmarksopen=true,%
bookmarksnumbered=true%
]{hyperref}%
\xdefinecolor{lightgray}{RGB}{247, 247, 247}%
\xdefinecolor{semilightgray}{RGB}{240, 240, 240}%
\xdefinecolor{middlegray}{RGB}{127, 127, 127}%
\xdefinecolor{blue}{RGB}{58, 95, 205}%
\xdefinecolor{deepskyblue}{RGB}{0, 154, 205}%
\xdefinecolor{chocolate}{RGB}{205, 102, 29}%

\setkomafont{pageheadfoot}{\normalfont\normalcolor\sffamily}%
\setkomafont{pagenumber}{\normalfont\normalcolor\sffamily}%
\automark{section}%
\setkomafont{captionlabel}{\normalfont\normalcolor\sffamily\bfseries}%

\setlist{%
  align=left,%
  labelsep=*,%
  leftmargin=*,%
  topsep=1mm,%
  itemsep=0mm%
}
\newcommand*{\mysquare}{\rule[0.18em]{0.36em}{0.36em}}
\newcommand*{\mytriangle}{\raisebox{0.12em}{\resizebox{0.48em}{0.48em}{$\blacktriangleright$}}}
\newcommand*{\mybar}{\rule[0.32em]{0.62em}{0.08em}}
\newcommand*{\mydot}{\raisebox{0.14em}{\resizebox{0.44em}{!}{$\bullet$}}}
\setlist[itemize,1]{label={\mysquare\ }}%
\setlist[itemize,2]{label={\mytriangle\ }}%
\setlist[itemize,3]{label={\mybar\ }}%
\setlist[itemize,4]{label={\mydot\ }}%
\setlist[enumerate,1]{label=\arabic*)}%
\setlist[enumerate,2]{label=\arabic{enumi}.\arabic*)}%
\setlist[enumerate,3]{label=\arabic{enumi}.\arabic{enumii}.\arabic*)}%

\makeatletter
\newcommand\myisodate{\number\year-\ifcase\month\or 01\or 02\or 03\or 04\or 05\or 06\or 07\or 08\or 09\or 10\or 11\or 12\fi-\ifcase\day\or 01\or 02\or 03\or 04\or 05\or 06\or 07\or 08\or 09\or 10\or 11\or 12\or 13\or 14\or 15\or 16\or 17\or 18\or 19\or 20\or 21\or 22\or 23\or 24\or 25\or 26\or 27\or 28\or 29\or 30\or 31\fi}%
\makeatother
\newcommand*{\abstractnoindent}{}%
\let\abstractnoindent\abstract
\renewcommand*{\abstract}{\let\quotation\quote\let\endquotation\endquote
  \abstractnoindent}
\deffootnote[1em]{1em}{1em}{\textsuperscript{\thefootnotemark}}%
\lstdefinestyle{input}{
  backgroundcolor=\color{semilightgray},%
  commentstyle=\itshape\color{chocolate},%
  keywordstyle=\color{blue},%
  stringstyle=\color{blue},%
  numbers=left,%
  numbersep=4.8pt,%
  numberstyle=\color{darkgray!80}\tiny%
}
\lstdefinestyle{output}{
  backgroundcolor=\color{lightgray}%
}

\lstdefinestyle{Lstyle}{
  language=[LaTeX]TeX,%
  texcs={},%
  otherkeywords={}%
}

\lstnewenvironment{Linput}[1][]{%
  \lstset{style=input, style=Lstyle}
  #1%
}{\vspace{-0.25\baselineskip}}%

\lstnewenvironment{Loutput}[1][]{%
  \lstset{style=output, style=Lstyle}
  #1%
}{\vspace{-0.25\baselineskip}}%

\lstdefinestyle{Rstyle}{
  language=R,%
  keywords={if, else, repeat, while, function, for, in, next, break},%
  otherkeywords={}%
}

\lstnewenvironment{Rinput}[1][]{%
  \lstset{style=input, style=Rstyle}
  #1%
}{\vspace{-0.25\baselineskip}}%

\lstnewenvironment{Routput}[1][]{%
  \lstset{style=output, style=Rstyle}
  #1%
}{\vspace{-0.25\baselineskip}}%

\newcommand*{\code}{\lstinline[
  basicstyle=\upshape\ttfamily,
  style=Rstyle,
  literate={~}{{$\sim$}}1
  ]}

\DeclareNameAlias{sortname}{last-first}%
\DefineBibliographyExtras{american}{\DeclareQuotePunctuation{}}%
\renewbibmacro*{volume+number+eid}{%
  \setunit*{\addcomma\space}%
  \printfield{volume}%
  \printfield{number}}
\DeclareFieldFormat*{number}{(#1)}
\DeclareFieldFormat*{title}{#1}%
\DeclareFieldFormat{doi}{%
  \ifhyperref
  {\href{http://dx.doi.org/#1}{\nolinkurl{doi:#1}}}%
  {\nolinkurl{doi:#1}}}%
\renewbibmacro*{in:}{}%
\DeclareFieldFormat{isbn}{ISBN #1}%
\DeclareFieldFormat{pages}{#1}%
\DeclareFieldFormat{url}{\url{#1}}%
\DeclareFieldFormat{urldate}{\mkbibparens{#1}}%
\renewcommand*{\cite}[2][]{\textcite[#1]{#2}}%

\newif\ifstarttheorem
\newtheoremstyle{mythmstyle}%
{0.5em}%
{0.5em}%
{}%
{}%
{\sffamily\bfseries\global\starttheoremtrue}%
{}%
{\newline}%
{\thmname{#1}\ \thmnumber{#2}\ \thmnote{(#3)}}%
\theoremstyle{mythmstyle}%
\newtheorem{definition}{Definition}[section]%
\newtheorem{proposition}[definition]{Proposition}

\newtheorem{remark}[definition]{Remark}

\newtheorem{algorithm}[definition]{Algorithm}

\makeatletter
\preto\itemize{%
  \if@inlabel
  \ifstarttheorem
  \mbox{}\par\nobreak\vskip\glueexpr-\parskip-\baselineskip+0.25em\relax\hrule\@height\z@
  \fi%
  \fi%
  \global\starttheoremfalse%
  \def\tempa{proof}%
  \ifx\tempa\mycurrenvir
  \ifstarttheorem
  \mbox{}\par\nobreak\vskip\glueexpr-\parskip-\baselineskip+0.25em\relax\hrule\@height\z@
  \fi%
  \fi%
  \global\starttheoremfalse%
}
\preto\enditemize{\global\starttheoremfalse}
\makeatother

\makeatletter
\preto\enumerate{%
  \if@inlabel
  \ifstarttheorem
  \mbox{}\par\nobreak\vskip\glueexpr-\parskip-\baselineskip+0.25em\relax\hrule\@height\z@
  \fi%
  \fi%
  \global\starttheoremfalse%
  \def\tempa{proof}%
  \ifx\tempa\mycurrenvir
  \ifstarttheorem
  \mbox{}\par\nobreak\vskip\glueexpr-\parskip-\baselineskip+0.25em\relax\hrule\@height\z@
  \fi%
  \fi%
  \global\starttheoremfalse%
}
\preto\endenumerate{\global\starttheoremfalse}
\makeatother

\newcommand{\ou}[3]{%
  \mathrel{%
    \vcenter{\offinterlineskip
      \ialign{##\cr$#1$\cr\noalign{\kern-#3}$#2$\cr}%
    }%
  }%
}
\newcommand*{\omu}[3]{\underset{#3}{\overset{#1}{#2}}}
\newcommand*{\T}{^{\top}}

\newcommand*{\isim}{\omu{\text{\tiny{ind.}}}{\sim}{}}

\newcommand*{\IN}{\mathbbm{N}}

\newcommand*{\IR}{\mathbbm{R}}

\newcommand*{\U}{\operatorname{U}}

\newcommand*{\N}{\operatorname{N}}

\newcommand*{\rd}{\mathrm{d}}

\renewcommand*{\P}{\mathbbm{P}}

\newcommand*{\R}{\textsf{R}}

\newcommand*{\ntrn}{n_{\text{trn}}}
\newcommand*{\nbat}{n_{\text{bat}}}

\newcommand*{\ngen}{n_{\text{gen}}}

\newcommand*{\nrep}{n_{\text{rep}}}
\newcommand*{\nprj}{n_{\text{prj}}}

\begin{document}
\thispagestyle{plain}
\begin{center}
  \sffamily
  {\bfseries\LARGE Dependence model assessment and selection with DecoupleNets\par}
  \bigskip\smallskip
  {\Large Marius Hofert\footnote{Department of Statistics and Actuarial Science, University of
      Waterloo, 200 University Avenue West, Waterloo, ON, N2L
      3G1,
      \href{mailto:marius.hofert@uwaterloo.ca}{\nolinkurl{marius.hofert@uwaterloo.ca}}. The
      author acknowledges support from NSERC (Grant RGPIN-2020-04897).},
    Avinash Prasad\footnote{Department of Statistics and Actuarial Science, University of
      Waterloo, 200 University Avenue West, Waterloo, ON, N2L
      3G1,
      \href{mailto:a2prasad@uwaterloo.ca}{\nolinkurl{a2prasad@uwaterloo.ca}}.},
    Mu Zhu\footnote{Department of Statistics and Actuarial Science, University of
      Waterloo, 200 University Avenue West, Waterloo, ON, N2L
      3G1,
      \href{mailto:mu.zhu@uwaterloo.ca}{\nolinkurl{mu.zhu@uwaterloo.ca}}. The
      author acknowledges support from NSERC (RGPIN-2016-03876).}
    \par\bigskip
    \myisodate\par}
\end{center}
\par\smallskip
\begin{abstract}
  Neural networks are suggested for learning a map from $d$-dimensional samples
  with any underlying dependence structure to multivariate uniformity in $d'$
  dimensions. This map, termed DecoupleNet, is used for dependence model
  assessment and selection. If the data-generating dependence model was known,
  and if it was among the few analytically tractable ones, one such
  transformation for $d'=d$ is Rosenblatt's transform. DecoupleNets have
  multiple advantages. For example, they only require an available sample and
  are applicable to $d'<d$, in particular $d'=2$. This allows for simpler model
  assessment and selection, both numerically and, because $d'=2$, especially
  graphically. A graphical assessment method has the advantage of being able to
  identify why, or in which region of the domain, a candidate model does not
  provide an adequate fit, thus leading to model selection in particular regions
  of interest or improved model building strategies in such
  regions. Through simulation studies with data from various copulas,
  the feasibility and validity of this novel DecoupleNet approach is
  demonstrated. Applications to real world data illustrate its usefulness for
  model assessment and selection.
\end{abstract}
\minisec{Keywords}
Neural networks, copulas, Rosenblatt transformation, model assessment, model
selection, graphical approach.
\minisec{MSC2010}
62H99, 65C60, 60E05, 62M45, 00A72, 65C10,
62M10. %

\section{Introduction}\label{sec:intro}
Copula modeling is well established by now, be it for parameter estimation in
statistical applications to engineering or hydrology, or for model building in
applications to finance, insurance or risk management, to name a few. The quest
to find an adequate copula for the modeling task at hand is omnipresent,
especially in higher-dimensional applications, where the application of interest
defines what constitutes ``higher-dimensional''; see, for example,
\cite{hofertoldford2018a} where even the most commonly applied copula models
fail to adequately capture the dependence found in basic log-return data. In
this paper, we present an approach to help the dependence modeler in
that quest. Moreover, and in contrast to other assessment methods -- for
example, purely numerically in terms of single numbers based on test statistics
\parencite[for example,][]{genestremillardbeaudoin2009} or graphically through pairs
only \parencite[for example,][]{hofertmaechler2014a, hofertoldford2018a}, if a copula
model is not deemed adequate, our approach can give guidance why, for example in
which tail region, the model fails to capture dependence properly.

If $\bm{U}\sim C$ for a $d$-dimensional copula $C$, the central idea of our
paper is to introduce a \emph{DecoupleNet}, a neural network to be specified
later, which maps $\bm{U}$ to $\bm{U}'\sim\U(0,1)^{d'}$ for $d'\le d$. The
flexibility of DecoupleNets allows us to learn transformations not only from any
(non-tractable) parametric copula $C$, but also from any underlying empirical
copula of a given dataset. A DecoupleNet is thus a natural tool for answering
the question
\begin{center}
  ``How can we assess and select copulas that best fit given data?''
\end{center}

\begin{figure}[htbp]
  \includegraphics[width=0.48\textwidth]{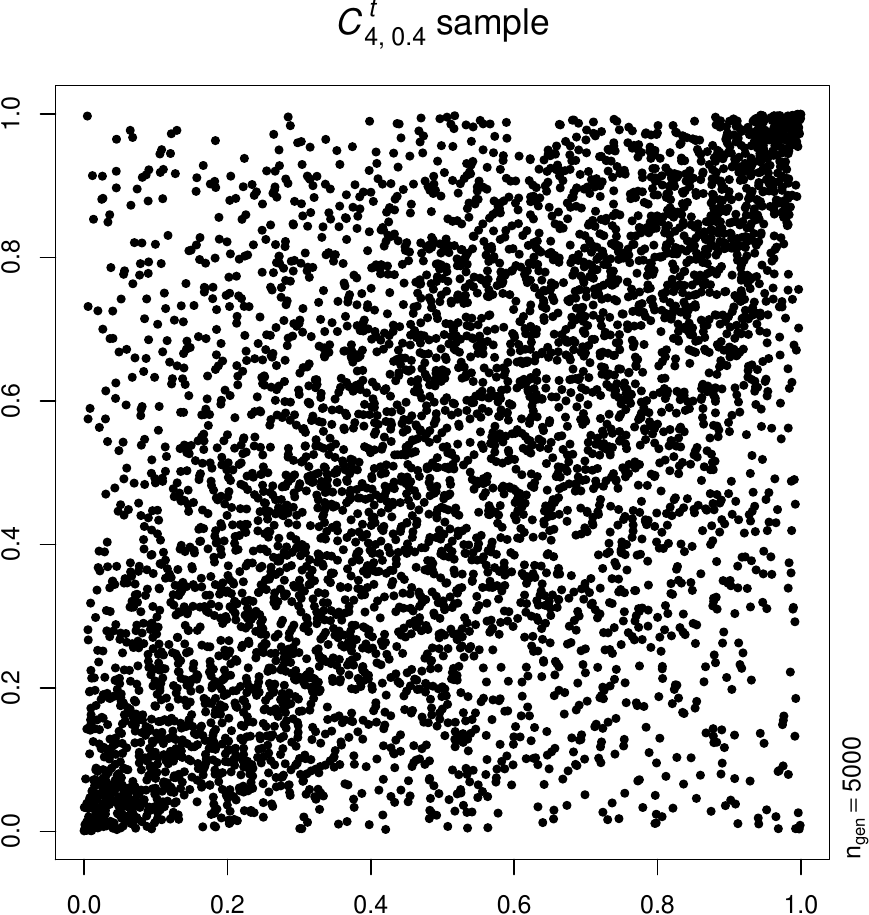}\hfill
  \includegraphics[width=0.48\textwidth]{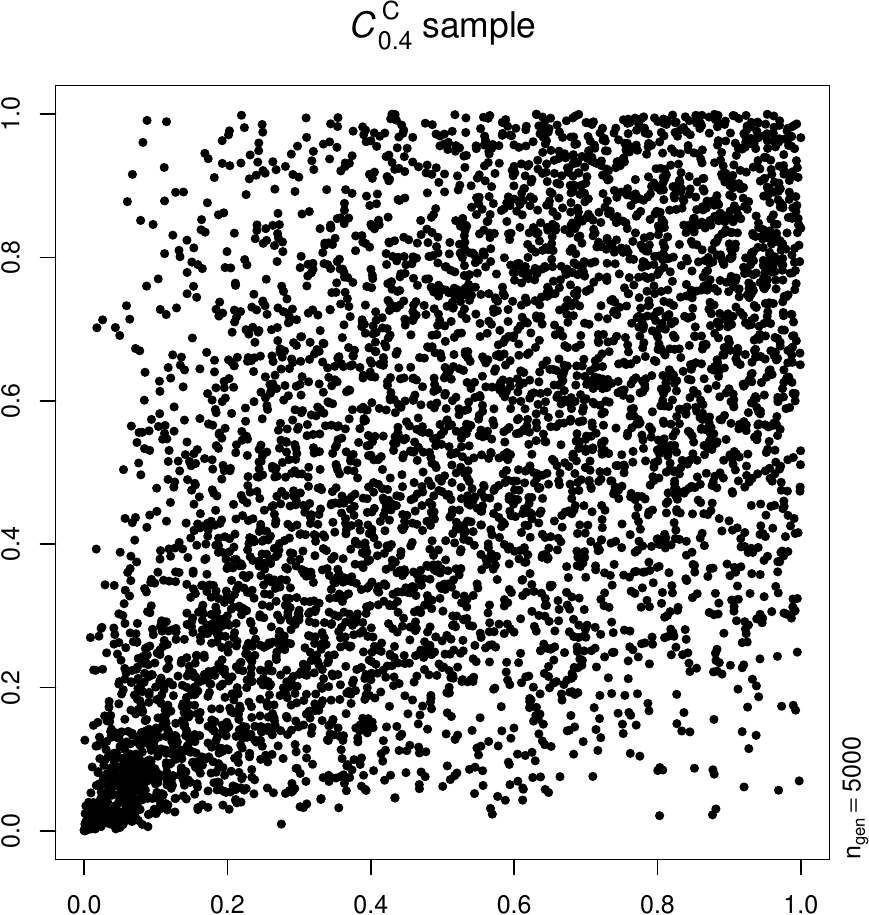}
  \\[4mm]
  \includegraphics[width=0.48\textwidth]{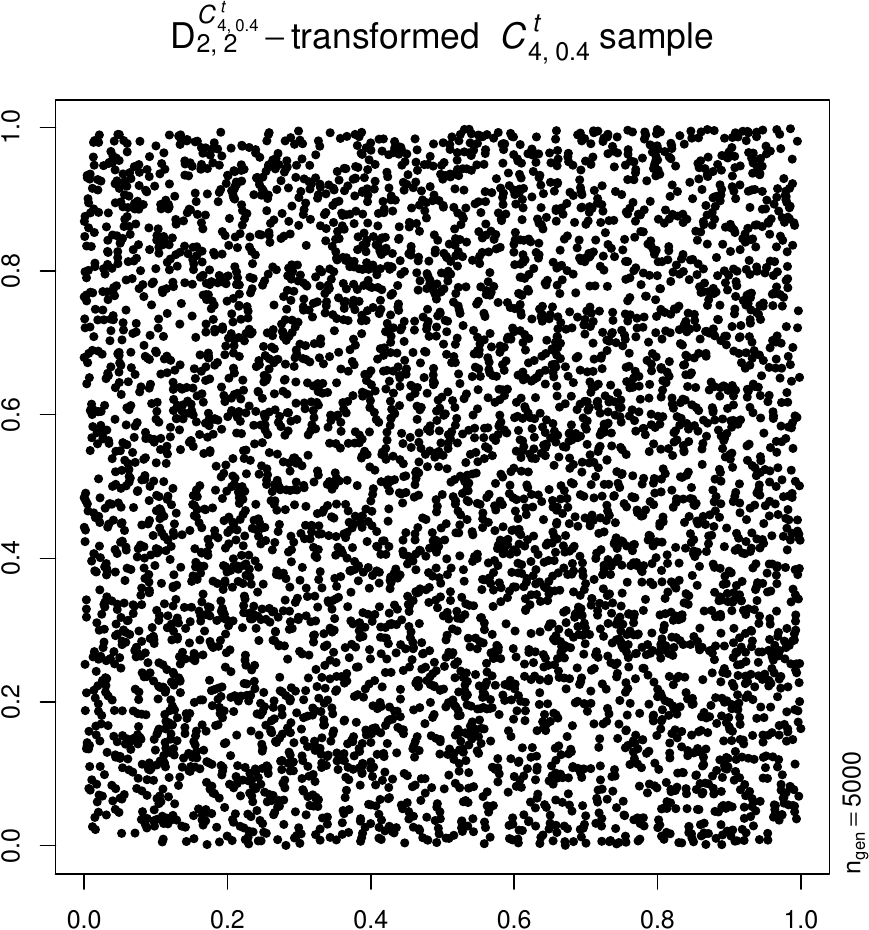}\hfill
  \includegraphics[width=0.48\textwidth]{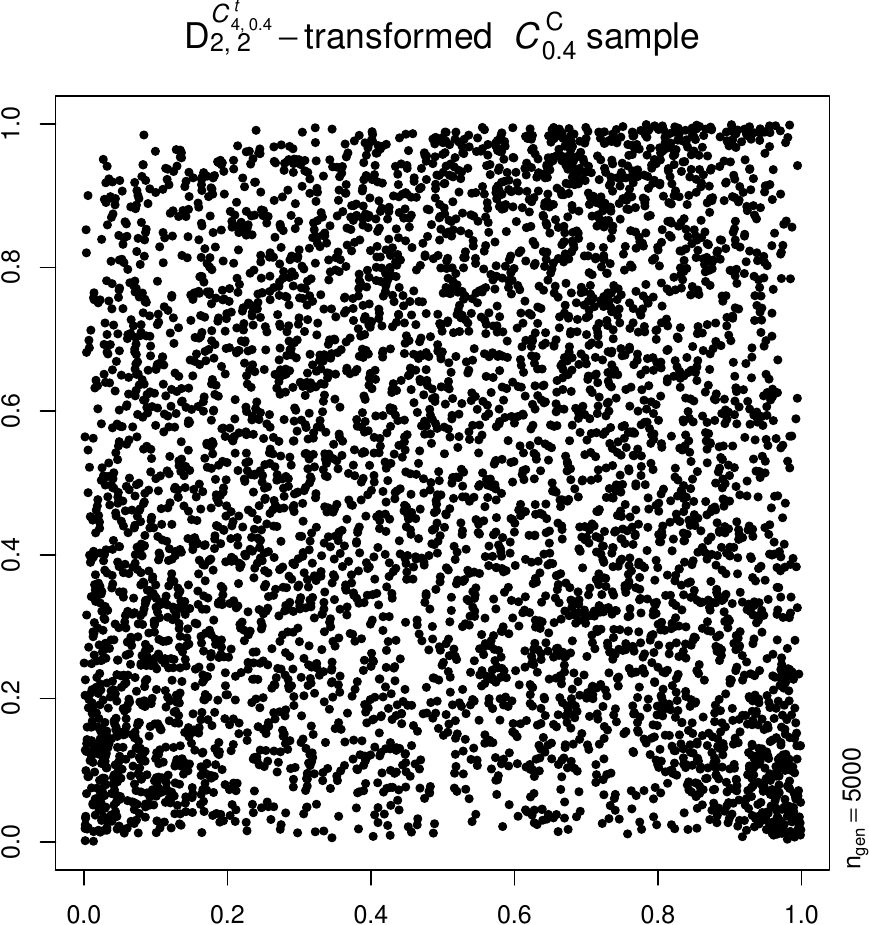}
  \caption{Samples of size $\ngen=5000$ from a bivariate $C^t_{4,0.4}$ copula (top left) and
    a bivariate $C^{\text{C}}_{0.4}$ copula (top right), with corresponding
    $D^{C^t_{4,0.4}}_{2,2}$-transformed samples (bottom row).}
  \label{fig:basic:ex}
\end{figure}

In Section~\ref{sec:decouplenets} we introduce DecoupleNets and our approach for
dependence model assessment and selection. As a high-level and easy to grasp
graphical example for $d=d'=2$ in this introduction, we trained a DecoupleNet,
denoted by $D^{C^t_{4,0.4}}_{2,2}$, on a sample of size 50\,000 from a bivariate
$t$ copula with $\nu=4$ degrees of freedom and Kendall's tau being $\tau=0.4$;
in short, $C=C^t_{\nu,\tau}=C^t_{4,0.4}$ (for notational ease, we omit
incorporting the dimension in the notation of copulas in this work, as it should
be clear from the context). The top left plot of Figure~\ref{fig:basic:ex} shows
a (new) sample of size $\ngen=5000$ from this copula.
Passing this sample through $D^{C^t_{4,0.4}}_{2,2}$ leads to the bottom left
plot whose uniformity confirms training quality. In the top right plot, we see a
sample of size 5000 from some candidate model -- here, a Clayton copula with the
same Kendall's tau $\tau=0.4$ -- whose adequacy for the training data we want to
assess. Applying $D^{C^t_{4,0.4}}_{2,2}$ to this sample leads to the plot on the
bottom right. We clearly see departure from uniformity suggesting that this
Clayton copula is not an adequate model for our data; we will later
  also color points belonging to specific regions of interest so that model
  assessment and selection can also be focused on specific regions.
Repeating this procedure for several candidate models allows us to assess them
and select the most suitable one. As we will see later, the same holds true if
$d>d'=2$, which makes this graphical assessment and selection approach feasible
in higher dimensions. To complement the graphical assessment, the quality of
  overall (non-)uniformity can also be summarized
  numerically. Section~\ref{sec:assess} investigates the details of the
graphical approach and the numerical summary in terms of simulated data. Another
advantage of DecoupleNets is that they can capture the dependence of any real
world data; see Section~\ref{sec:applications}. Moreover, the perceived
  computational burden of having to train a neural network can become
  negligible in comparison to having to find parametric candidate models,
  estimate their parameters and compute quantities such as the Rosenblatt
  transform for model assessment and selection (if available at all).
Section~\ref{sec:concl} concludes with a summary and outlook.

\section{DecoupleNets for model assessment and selection}\label{sec:decouplenets}
\subsection{Transformation}
Let $C$ be any $d$-dimensional copula. A \emph{DecoupleNet} $D^C_{d,d'}$ is a
neural network that maps $\bm{U} \sim C$ to $\bm{U}'\sim\U(0,1)^{d'}$, so
$D^C_{d,d'}(\bm{U})=\bm{U}'$ with the goal of model assessment and selection.
We specify this map to be $D^C_{d,d'} = T^C \circ \bm{\Phi}^{-1}$, where
$\bm{\Phi}^{-1}(\bm{u}) = (\Phi^{-1}(u_1),\dots,\Phi^{-1}(u_d))$ is a
componentwise transformation with the standard normal quantile function
$\Phi^{-1}$ and $T^C$ is a trained neural network. The initial map
$\bm{\Phi}^{-1}$ to standard normal margins acts as a pre-processing step that
helps facilitate the training of the neural network $T^C$.

\begin{remark}[Rosenblatt's transformation]
  Another transformation from $\bm{U}\sim C$ to $\bm{U}'\sim\U(0,1)^{d'}$, but
  limited to $d'=d$, is the transformation of \cite{rosenblatt1952}. It is the
  (only known) general such transformation from $\bm{U}\sim C$ (the ``general''
  referring to the fact that it applies to any $d$-dimensional copula $C$) to
  $\bm{U}'\sim\U(0,1)^d$; for specific $C$, there may be other transformations,
  for example the one of \cite{wuvaldezsherris2007} for Archimedean
  copulas. Having to rely on such transformations has several main drawbacks in
  comparison to DecoupleNets. First, for any $d$-dimensional copula $C$,
  Rosenblatt's transformation is given by
  $\bm{U}'=R^C_{d}(\bm{U})$ with first component $R^C_{d}(\bm{U})_1=U_1$
  and $j$th component
  \begin{align*}
    R^C_{d}(\bm{U})_j= C_{j|1,\dots,j-1}(U_j\,|\,U_1,\dots,U_{j-1}),\quad j=2,\dots,d;
  \end{align*}
  here $C_{j|1,\dots,j-1}(u_j\,|\,u_1,\dots,u_{j-1})=\P(U_j\le u_j\,|\,U_1=u_1,\dots,U_{j-1}=u_{j-1})$.
  Under differentiability assumptions on $C$, these conditional distributions
  can be expressed as
  \begin{align}
    C_{j|1,\dots,j-1}(u_j\,|\,u_1,\dots,u_{j-1})=\frac{\frac{\partial^{j-1}}{\partial x_{j-1}\dots\partial x_1} C^{(1,\dots,j)}(x_1,\dots,x_j)\bigr|_{(x_1,\dots,x_j)=(u_1,\dots,u_j)}}{\frac{\partial^{j-1}}{\partial x_{j-1}\dots\partial x_1} C^{(1,\dots,j-1)}(x_1,\dots,x_{j-1})\bigr|_{(x_1,\dots,x_{j-1})=(u_1,\dots,u_{j-1})}}\label{eq:cond:distr:der}
  \end{align}
  For most copulas, \eqref{eq:cond:distr:der} is not available analytically, nor
  tractable numerically. Notable exceptions where \eqref{eq:cond:distr:der} is
  available are normal, $t$ and Clayton copulas. However, these copulas are
  typically not flexible enough to fit real world data well, the second
  drawback. This especially applies to higher dimensions where, additionally,
  the fact that $d'=d$ makes computing \eqref{eq:cond:distr:der} numerically and
  computationally intractable, the third drawback. Despite these drawbacks,
  Rosenblatt's transformation is applied in copula modeling; see, for example,
  \cite{genestremillardbeaudoin2009}.

  As we will see, DecoupleNets have none of these drawbacks. Moreover, although
  run time is not a focus here, note that the perceived computational burden of
  having to train a neural network is well compensated by considering the only
  available (but largely limited) alternative, such as the Rosenblatt transform
  (for $d'=d$). In virtually all applications, we do \emph{not} know the true
  underlying copula, so we would first need to estimate various candidate models
  and then compute their (implied) Rosenblatt transforms, etc. In this light,
  having to train just one neural network is actually orders of
  magnitudes faster. Furthermore, the training of the rather simple neural
  networks we use is by no means very time-consuming (especially also with the
  rather small sample sizes one often faces in practice).

Furthermore, the copula $C$ underlying $D^C_{d,d'}$ is typically not known
analytically and only specified through a given sample.

\end{remark}

\subsection{Optimization}
To train a DecoupleNet, we make use of a generative neural network modeling
technique introduced by \cite{liswerskyzemel2015} and
\cite{dziugaiteroyghahramani2015}. We work with a family $\mathcal{T}$ of
feedforward neural networks with a pre-specified architecture, where a network
$T^C\in\mathcal{T}$ is characterized by weights $\bm{W}$. Given a sample
$\{\bm{U}_i\}_{i=1}^{\ntrn}$ from $C$ and a sample $\{\bm{U}'_i\}_{i=1}^{\ntrn}$ from $\U(0,1)^{d'}$, we minimize
\begin{align}
  &\phantom{{}={}}\mathcal{L}\Bigl(\{T^C(\bm{\Phi}^{-1}(\bm{U}_{i}))\}_{i=1}^{\ntrn},\{\bm{U}'_i\}_{i=1}^{\ntrn}\Bigr)\notag\\
  &=\frac{1}{\ntrn^2}\sum_{i=1}^{\ntrn}\sum_{i'=1}^{\ntrn} \Bigl(K\bigl(T^C(\bm{\Phi}^{-1}(\bm{U}_{i})),T^C(\bm{\Phi}^{-1}(\bm{U}_{i'}))\bigr) - 2K(T^C(\bm{\Phi}^{-1}(\bm{U}_{i'})), \bm{U}'_i) + K(\bm{U}'_i,\bm{U}'_{i'}) \Bigr)\label{eq:opt:kernel}
\end{align}
over all $T^C\in\mathcal{T}$ by a version of stochastic gradient descent, where
$K(\cdot,\cdot)$ is a kernel function. Minimizing~\eqref{eq:opt:kernel}
ensures that the distribution of the DecoupleNet output
$\{D^C_{d,d'}(\bm{U}_i)\}_{i=1}^{\ntrn}$ is as close as possible to $\U(0,1)^{d'}$.
This is due to the fact that the loss function $\mathcal{L}$ being minimized is equal to
\begin{align}
  \left\|\frac{1}{\ntrn}\sum_{i'=1}^{\ntrn} \varphi\bigl(T^C(\bm{\Phi}^{-1}(\bm{U}_{i'}))\bigr)
  -\frac{1}{\ntrn}\sum_{i=1}^{\ntrn} \varphi(\bm{U}'_i)\right\|^2,\label{eq:opt:phi}
\end{align}
where $\varphi$ is the implied feature map of $K$, such that
$K(\bm{u},\bm{v}) = \varphi(\bm{u})\T\varphi(\bm{v})$. By selecting $K$ to be a
Gaussian kernel $K(\bm{u},\bm{v})=\exp(-\|\bm{u}-\bm{v}\|^2/\sigma)$, where
$\sigma>0$ denotes the bandwidth parameter, the two terms in \eqref{eq:opt:phi}
will contain all empirical moments of
$\{D^C_{d,d'}(\bm{U}_{\ell})\}_{\ell=1}^{\ntrn}$ and
$\{\bm{U}'_i\}_{i=1}^{\ntrn}$, respectively, thus ensuring that the DecoupleNet
output matches the $\U(0,1)^{d'}$ distribution. For more details about the types
of generative neural networks we use (that is, $T^C$ in \eqref{eq:opt:kernel})
and their capabilities for learning maps between uniformity and (possibly
empirically) specified dependencies, see \cite{hofertprasadzhu2021a}. As in this
reference, we follow the suggestion of \cite{liswerskyzemel2015} and work with a
mixture of Gaussian kernels with different bandwidth parameters in order to
avoid selecting a single optimal bandwidth parameter.

\subsection{Training}
Directly performing the optimization in~\eqref{eq:opt:kernel}, also known as
\emph{batch optimization}, would involve all $\binom{\ntrn}{2}$ pairs of
observations which is memory-prohibitive even for moderately large
$\ntrn$. Instead, we adopt a \emph{mini-batch optimization} procedure, where the
training dataset is partitioned into \emph{batches} of size $\nbat$ and the
batches are used sequentially to update the weights $\bm{W}$ with the Adam
optimizer of \cite{kingma2014b} (a ``memory-sticking gradient'' procedure, that
is a weighted combination of the current gradient and past gradients from
earlier iterations). After a pass through the entire training data, that is,
after roughly $(\ntrn/\nbat)$-many gradient steps, one \emph{epoch} of the
neural network training is completed. The trade-off in utilizing mini-batches,
particularly with a smaller batch size $\nbat$, is that the objective function
is computed only with partial information for each gradient step in the
optimization. For relatively small datasets however batch optimization can still
be used, and conceptually we can view it as a special case of the mini-batch
procedure (for $\nbat=\ntrn$).

\subsection{Understanding DecoupleNets and how to use them for dependence model
  assessment and selection}
We now briefly revisit the example of Section~\ref{sec:intro} to illustrate the
nature of a trained DecoupleNet transform and why it is useful for model
assessment and selection.

By construction, given an input sample $\{\bm{U}_i\}_{i=1}^{\ngen}$ from a known
copula (or pseudo-observations of an unknown copula) $C$, the trained DecoupleNet $D^C_{d,d'}$ generates an output sample
$\{D^C_{d,d'}(\bm{U}_i)\}_{i=1}^{\ngen}$ that is approximately $\U(0,1)^{d'}$. On the
other hand, for an input sample $\{\tilde{\bm{U}}_i\}_{i=1}^{\ngen}$ from some
candidate copula $\tilde{C}$ with $\tilde{C}\neq C$, the DecoupleNet output
$\{D^C_{d,d'}(\tilde{\bm{U}}_i)\}_{i=1}^{\ngen}$ should exhibit departures from
$\U(0,1)^{d'}$.

\begin{figure}[htbp]
  \includegraphics[width=0.48\textwidth]{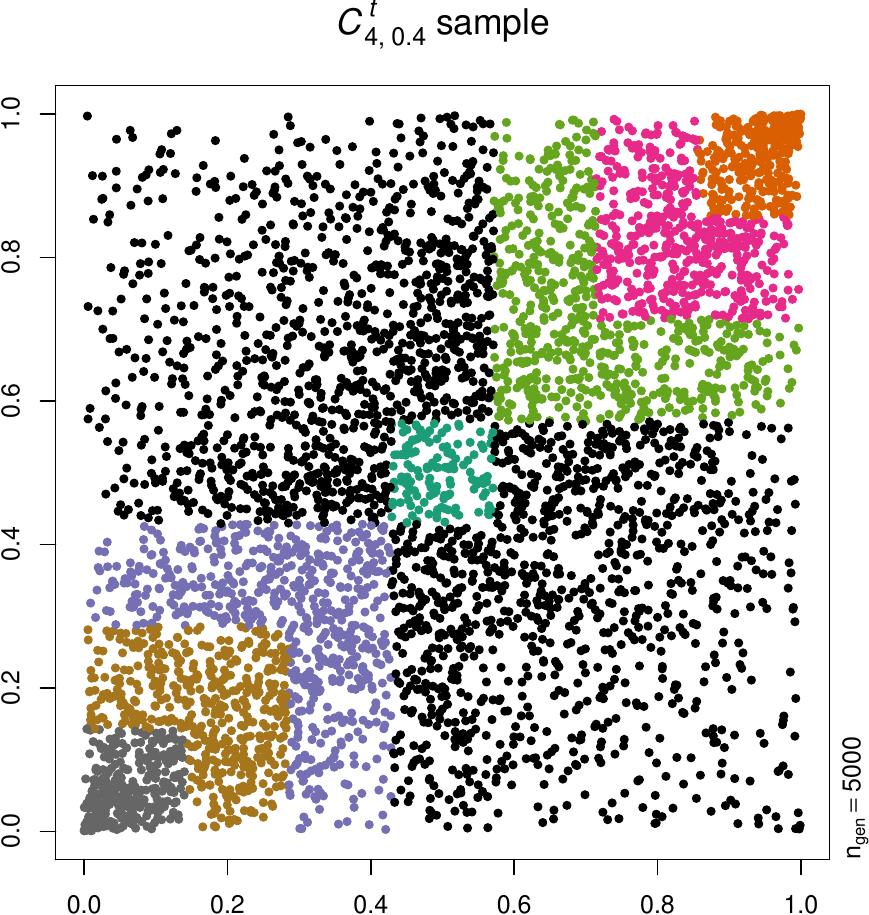}\hfill
  \includegraphics[width=0.48\textwidth]{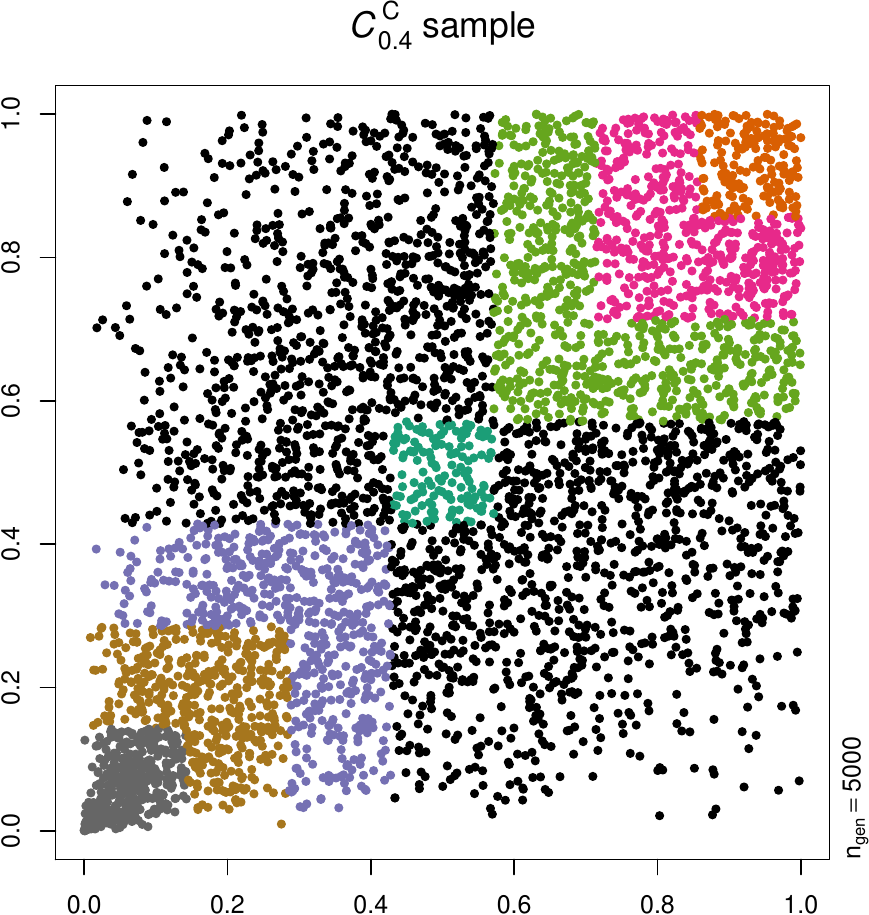}
  \\[4mm]
  \includegraphics[width=0.48\textwidth]{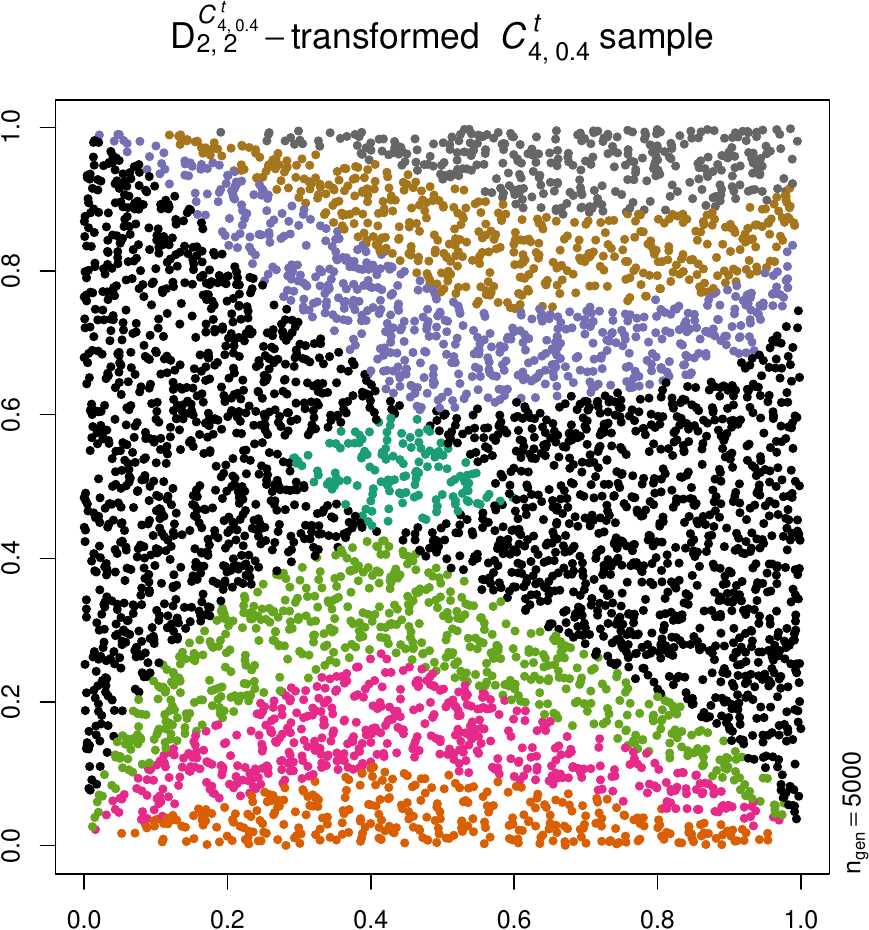}\hfill
  \includegraphics[width=0.48\textwidth]{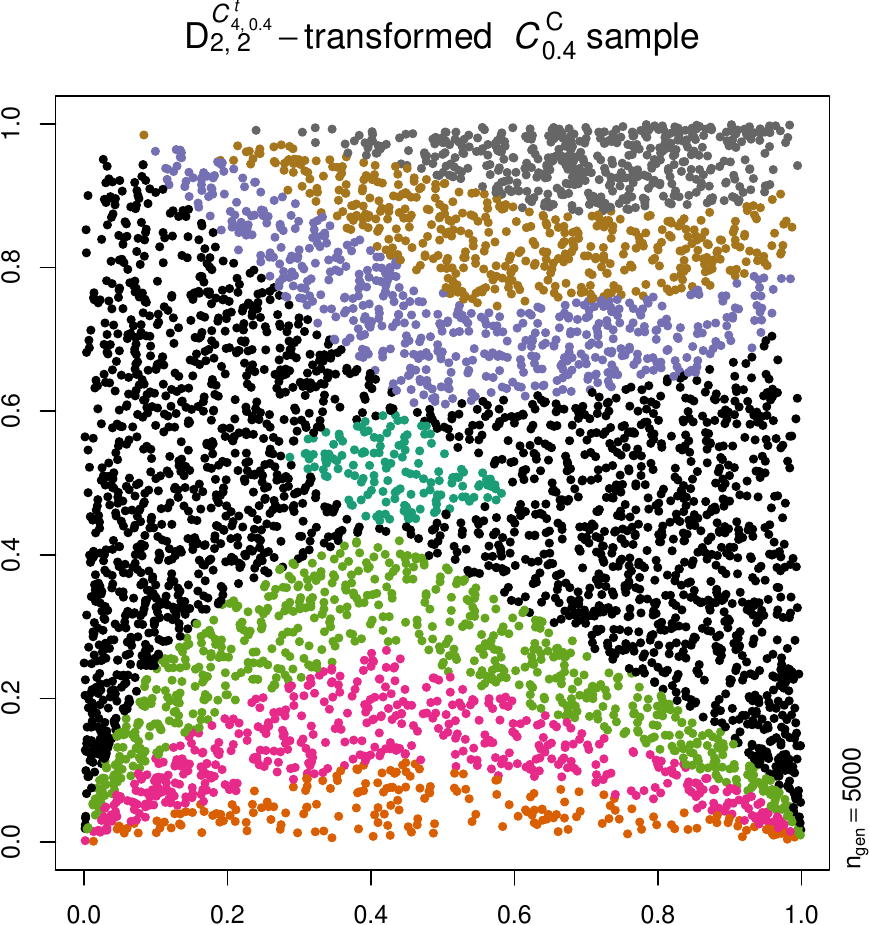}
  \caption{Colored samples of size $\ngen=5000$ from a bivariate $C^t_{4,0.4}$
    copula (top left) and a bivariate $C^{\text{C}}_{0.4}$ copula (top right),
    with corresponding $D^{C^t_{4,0.4}}_{2,2}$-transformed samples (bottom
    row).}
  \label{fig:col:ex}
\end{figure}

To demonstrate this idea, Figure~\ref{fig:col:ex} shows the same data as
Figure~\ref{fig:basic:ex} but we now colored different regions of the input
samples $\{\bm{U}_i\}_{i=1}^{\ngen}$ and, correspondingly, the
output samples $\{D^{C^t_{4,0.4}}_{2,2}(\bm{U}_i)\}_{i=1}^{\ngen}$.
Comparing the plot on the bottom left with the one on the top left, we see from
the colored regions that samples $\{\bm{U}_i\}_{i=1}^{\ngen}$ in the joint right
tail of $C^t_{4,0.4}$ are (here) mapped to samples
$\{D^{C^t_{4,0.4}}_{2,2}(\bm{U}_i\}_{i=1}^{\ngen})\}$ that concentrate near the
bottom (small second component), and similarly for the joint left
tail. Comparing the plot on the bottom right with the one on the bottom left, we
see that the region at the bottom (with samples from the joint right tail) is
underrepresented, so there must have been too few input samples in the upper
right region -- indeed what we see in the plot at the top right in comparison to
the one on the top left; one can also verify this numerically, the probability
to fall in $[6/7,1]^2$ is about $0.0673$ under $C^t_{4,0.4}$ and about
$0.0400$ under $C^{\text{C}}_{0.4}$.  Similarly, the region at the top with
samples from the joint left tail is overrepresented, so there must have been too
many input samples in the lower left region -- indeed what we see in the plot at
the top right in comparison to the one on the top left; and again one can verify
this numerically, the probability to fall in $[0,1/7]^2$ is about
$0.0673$ under $C^t_{4,0.4}$ and about $0.0874$ under $C^{\text{C}}_{0.4}$. In
short, the colors indicate to which regions input samples are transformed and
thus allow us to assess and select copulas that well capture specific regions of
interest.

\begin{figure}[htbp]
  \includegraphics[width=0.48\textwidth]{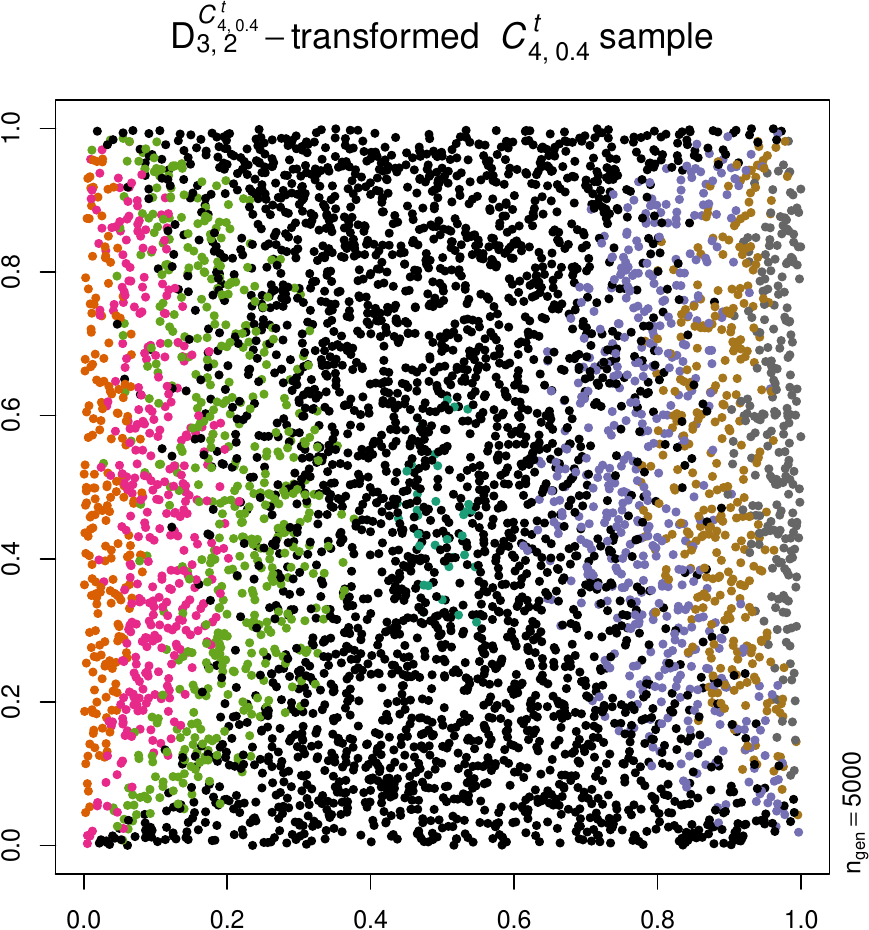}\hfill
  \includegraphics[width=0.48\textwidth]{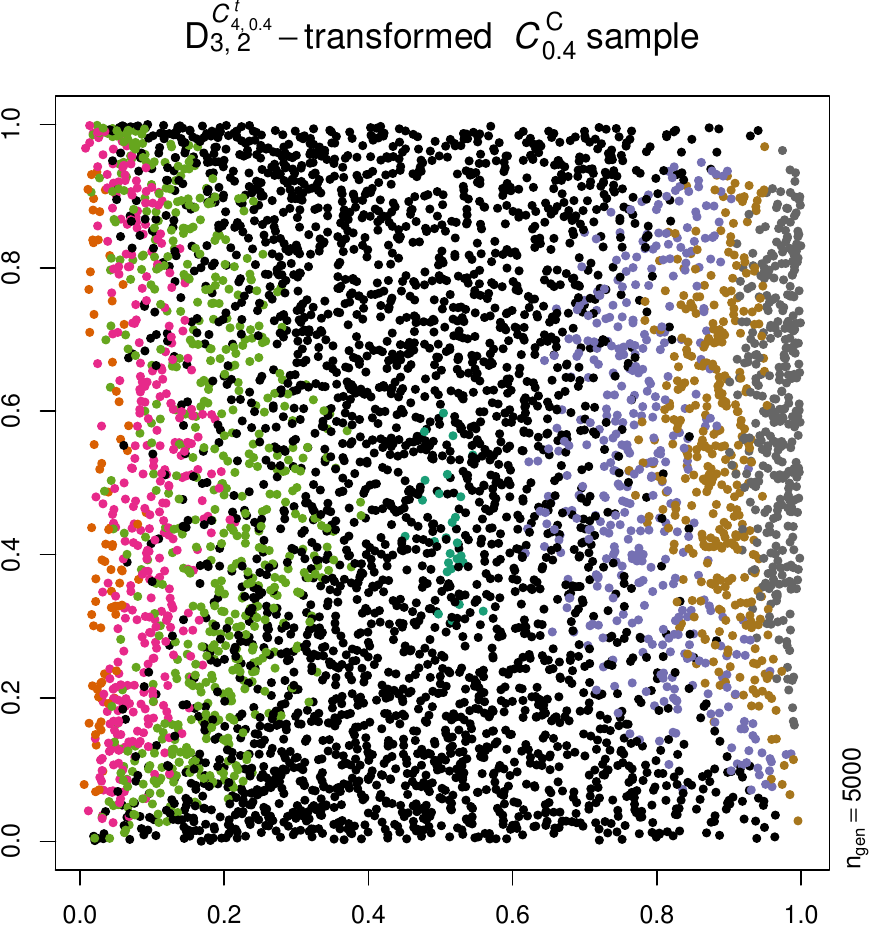}
  \caption{$D^{C^t_{4,0.4}}_{3,2}$-transformed colored
    samples of size $\ngen=5000$ from trivariate $C^t_{4,0.4}$ (left)
    and $C^{\text{C}}_{0.4}$ (right) copulas.}
  \label{fig:col:ex:3d}
\end{figure}

Figure~\ref{fig:col:ex:3d} shows $D^{C^t_{4,0.4}}_{3,2}$-transformed colored samples
from a $C^t_{4,0.4}$ and a $C^{\text{C}}_{0.4}$ copula.
This is an example where the DecoupleNet maps from $d$ to $d'$ with $3=d>d'=2$,
and we still see from the overrepresented dark color (joint left tail) and
underrepresented bright color (joint right tail) which regions
$C^{\text{C}}_{0.4}$ fails to capture.

These examples already demonstrates how DecoupleNets can be utilized for
graphical model assessment of copulas. For additional bivariate and
higher-dimensional examples of graphical assessments in simulated and
real-world settings, see Sections~\ref{sec:graph:approach} and
\ref{sec:applications}, respectively.

A question one may have is whether the trick with colors always works.
  For example, it would be much harder (or rather impossible) to interpret the
  under- or over-representation of colors if they were distributed all over the
  place (instead of within topologically connected regions) after the
  DecoupleNet transformation is applied. According to the following result
  based on the notion of topological connectedness, this cannot happen.
\begin{proposition}[Connected colored regions]
  A DecoupleNet maps connected colored regions to connected colored regions.
\end{proposition}
\begin{proof}
  By \cite[Theorem~23.5]{munkres2000}, the image of a connected space under a
  continuous map is connected. The claim follows by realizing that DecoupleNets
  are continuous maps.
\end{proof}

Suppose we are given data $\{\bm{X}_i\}_{i=1}^{\ntrn}$ in $\IR^d$, assumed to
come from a joint distribution with continuous marginal distribution
functions. Since our primary focus is on modeling the underlying dependence
structure, we first compute the pseudo-observations
$\hat{U}_{i,j} = \hat{R}_{i,j}/(\ntrn + 1)$, $i=1,\dots,\ntrn$, $j=1,\dots,d$,
where $\hat{R}_{i,j}$ denotes the rank of $X_{i,j}$ among
$X_{1,j},\dots,X_{\ntrn,j}$. Let $\hat{C}_{\ntrn}$ denote the empirical copula
of $\{\hat{\bm{U}}_i\}_{i=1}^{\ntrn}$. Now suppose we are interested in
selecting the best copula from a collection $\mathcal{C}$ of candidate
models. We denote an element of $\mathcal{C}$ as $C_{\bm{\theta}}$ for a
parameter vector $\bm{\theta}$; note however that $C_{\bm{\theta}}$ could very
well be a copula without any parameter vector to estimate, for example, if specified
by an expert. For each parametric candidate model
$C_{\bm{\theta}} \in \mathcal{C}$, we proceed by first fitting $\bm{\theta}$ to
the pseudo-observations $\{\hat{\bm{U}}_i\}_{i=1}^{\ntrn}$. Next, we learn a
DecoupleNet $D^{\hat{C}_{\ntrn}}_{d,d'}$ from the pseudo-observations
$\{\hat{\bm{U}}_i\}_{i=1}^{\ntrn}$ to $\U(0,1)^{d'}$. By passing samples from
each fitted candidate copula $C_{\hat{\bm{\theta}}}$ through
$D^{\hat{C}_{\ntrn}}_{d,d'}$, we can use the resulting DecoupleNet-transformed
samples to rank the fit of the candidate copulas to the pseudo-observations,
that is, the closer the DecoupleNet-transformed sample is to $\U(0,1)^{d'}$, the
better. Formulated as an algorithm, our proposed model selection procedure is
summarized in Algorithm~\ref{algo:model:selection}.

We can also numerically summarize how close a DecoupleNet output
$\{D^C_{d,d'}(\bm{U}_i)\}_{i=1}^{\ngen}$ is to $\U(0,1)^{d'}$ using a score of the
Cram\'{e}r-von-Mises (CvM) type,
\begin{align}
  S_{\ngen,d'} = \int_{[0,1]^{d'}} \ngen \Biggl( C_{\ngen}(\bm{u})-\prod_{j=1}^{d'}u_j\Biggr)^2\,\rd C_{\ngen}(\bm{u}), \label{CvM:score}
\end{align}
where $C_{\ngen}$ is the empirical copula of the pseudo-observations of $\{D^C_{d,d'}(\bm{U}_i)\}_{i=1}^{\ngen}$,
so the empirical copula of $\{\bm{R}_i/(\ngen + 1)\}_{i=1}^{\ngen}$ with $\bm{R}_i=(R_{i,1},\dots,R_{i,d'})$, where
$R_{i,j}$ denotes the rank of the $j$th among all components of $D^C_{d,d'}(\bm{U}_i)$.

The following algorithm describes both the graphical approach and the numerical summary of
our proposed model selection procedure.
\begin{algorithm}[Model assessment and selection with DecoupleNets]\label{algo:model:selection}
  \begin{enumerate}
  \item Given data $\{\bm{X}_i\}_{i=1}^{\ntrn}$, construct the
    pseudo-observations $\{\hat{\bm{U}}_{i}\}_{i=1}^{\ntrn}$. Their
    empirical copula is denoted by $\hat{C}_{\ntrn}$.
  \item Train the DecoupleNet $D^{\hat{C}_{\ntrn}}_{d,d'}$ based on the pseudo-observations
    $\{\hat{\bm{U}}_i\}_{i=1}^{\ntrn}$ and the desired output
    $\{\bm{U}'_i\}_{i=1}^{\ntrn}$ from $\U(0,1)^{d'}$. %
  \item For each parametric candidate copula $C_{\bm{\theta}} \in \mathcal{C}$,
    estimate the parameter $\bm{\theta}$ of $C_{\bm{\theta}}$ using the
    pseudo-observations $\{\hat{\bm{U}}_i\}_{i=1}^{\ntrn}$ to obtain
    $C_{\bm{\hat{\theta}}}$. This leaves us with a finite number of candidate
    copulas, fitted or fixed; the latter refers to copulas with fixed parameters
    where no estimation is necessary. We denote a generic candidate copula by
    $\tilde{C}$.
  \item For each candidate copula $\tilde{C}$ do:
    \begin{enumerate}
    \item Generate a sample $\{\tilde{\bm{U}}_i\}_{i=1}^{\ngen}$ from $\tilde{C}$.
    \item Pass $\{\tilde{\bm{U}}_i\}_{i=1}^{\ngen}$ through the
      DecoupleNet $D^{\hat{C}_{\ntrn}}_{d,d'}$ to obtain $\{D^{\hat{C}_{\ntrn}}_{d,d'}(\tilde{\bm{U}}_i)\}_{i=1}^{\ngen}$.
    \item\label{step:last}
      For a graphical approach ($d'=2$), create a scatter plot of the
      DecoupleNet-transformed sample
      $\{D^{\hat{C}_{\ntrn}}_{d,d'}(\tilde{\bm{U}}_i)\}_{i=1}^{\ngen}$.
      Determine the color of sample points $\tilde{\bm{U}}_i$ according to
      regions of interest; then, color the sample
      $\{D^{\hat{C}_{\ntrn}}_{d,d'}(\tilde{\bm{U}}_i)\}_{i=1}^{\ngen}$
      accordingly and create a colored scatter plot. For the numerical summary,
      compute the Cram\'{e}r-von-Mises score $S_{\ngen,d'}$ of~\eqref{CvM:score}
      for the DecoupleNet-transformed sample
      $\{D^{\hat{C}_{\ntrn}}_{d,d'}(\tilde{\bm{U}}_i)\}_{i=1}^{\ngen}$.
    \end{enumerate}
  \item For the graphical approach, compare the two types of scatter
    plots created in Step~\ref{step:last} for all candidate copulas
    $\tilde{C}$ and select the candidate copula that shows least
    non-uniformity overall or in the region of interest. For the numerical
    summary, compare the Cram\'{e}r-von-Mises scores for all candidate copulas
    $\tilde{C}$ and select the candidate copula that yields the
    lowest Cram\'{e}r-von-Mises score.
  \end{enumerate}
\end{algorithm}

In what follows we consider $d'=2$ which allows us to investigate both the
graphical approach and the numerical summary for dependence model assessment and
selection. We also investigated the numerical summary for $d'>2$ (results
not presented) and found no advantage over $d'=2$. Moreover, the case $d'=2$
has the advantage of reduced run time when training a DecoupleNet.

\begin{remark}[About the importance of graphical assessments]\label{rem:graph:vs:num}
  In many areas of statistics, graphical tools are preferred over summary
  statistics (single numbers); see, for example, the popularity of Q-Q plots for
  (univariate) model assessment. The problem with a numerical assessment through a
  summary statistic like~\eqref{CvM:score} is that, if one deems a model not
  adequate based on a single number, one does not gain much information about
  \emph{why} it is not adequate. As we mentioned in the beginning of Section~\ref{sec:intro},
  it is typically hard to find an adequate dependence model; most
  will be deemed inadequate. In these cases one needs to know \emph{why} the
  model is inadequate and then make a decision about changing the model
  accordingly or whether to keep working with the model. In many applications,
  copula models are not necessarily used as overall models, but only in specific
  regions of interest. For example, if only the joint right tail is of interest, a
  copula model deemed overall inadequate but which provides a good fit in the
  joint right tail may very well be adequate to work with. As we have demonstrated in this section,
  graphical applications of DecoupleNets are
  useful for model assessment and selection based on regions of interest.
\end{remark}

\section{Model assessment and selection based on simulated data}\label{sec:assess}
In this section we investigate our model assessment and selection procedure
based on simulated data. Section~\ref{sec:graph:approach} considers the
graphical approach, and Section~\ref{sec:num:approach} the numerical summary.

\subsection{Graphical approach}\label{sec:graph:approach}
\begin{figure}[htbp]
  \includegraphics[width=0.19\textwidth]{fig_scatter_cop_dim_2_out_dim_2_ngen_5000_t4_tau_0.4_t4_tau_0.4.pdf}\hfill
  \includegraphics[width=0.19\textwidth]{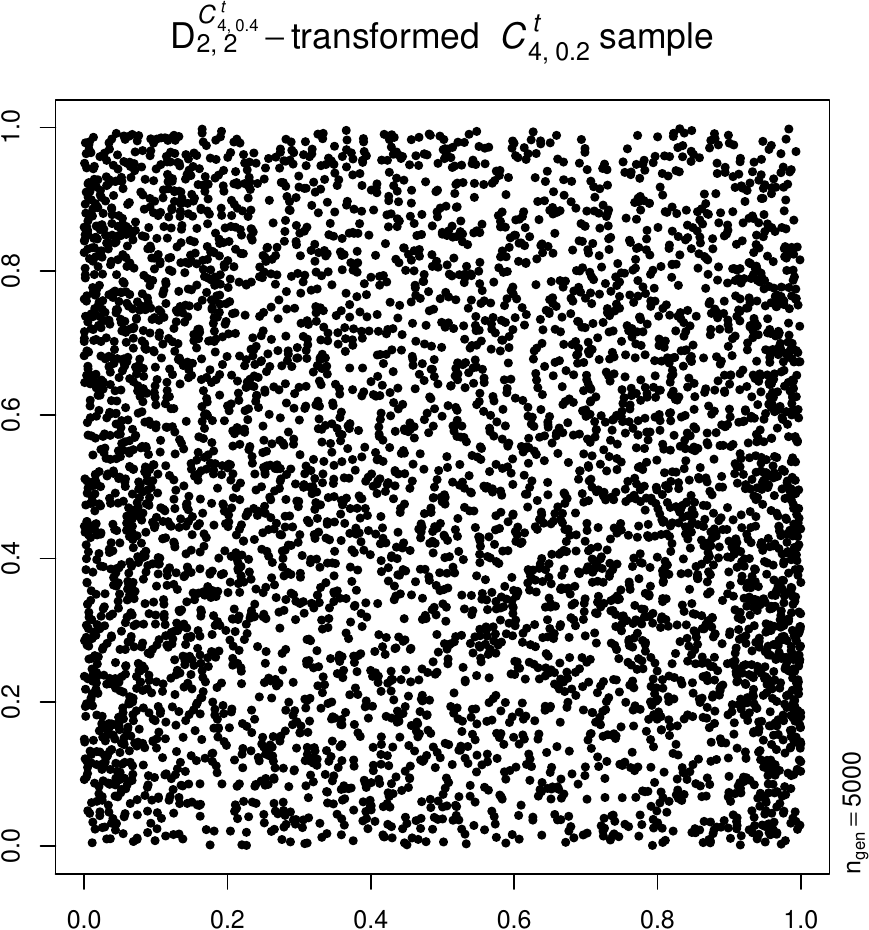}\hfill
  \includegraphics[width=0.19\textwidth]{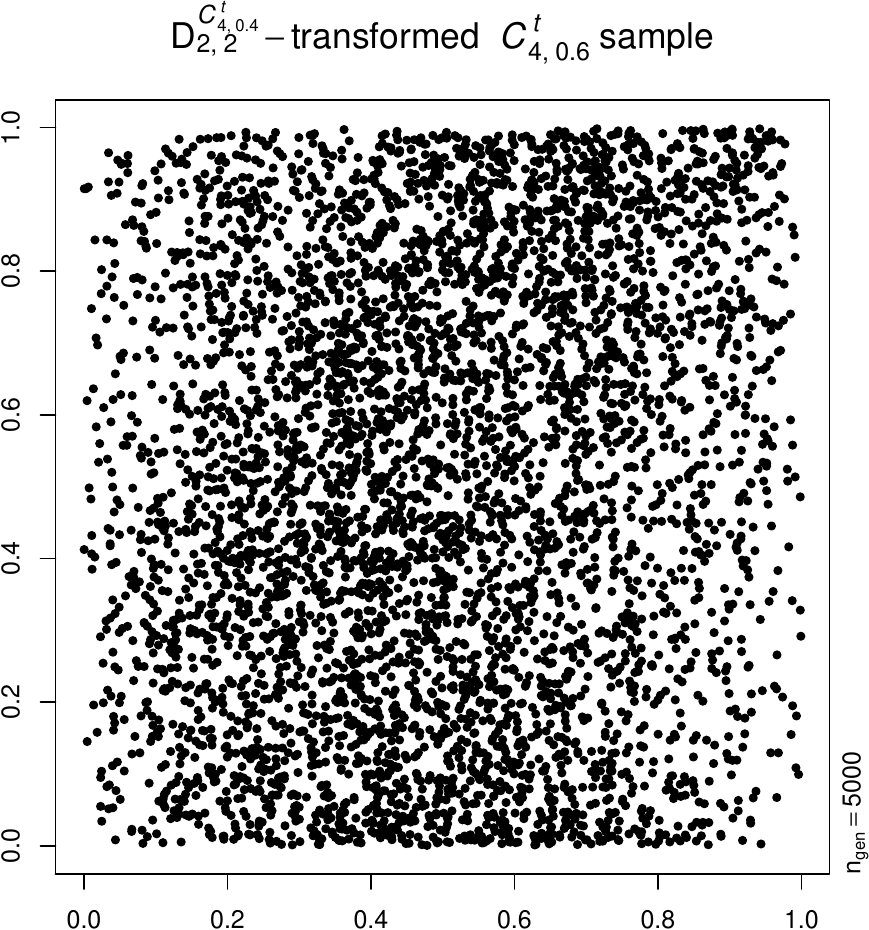}\hfill
  \includegraphics[width=0.19\textwidth]{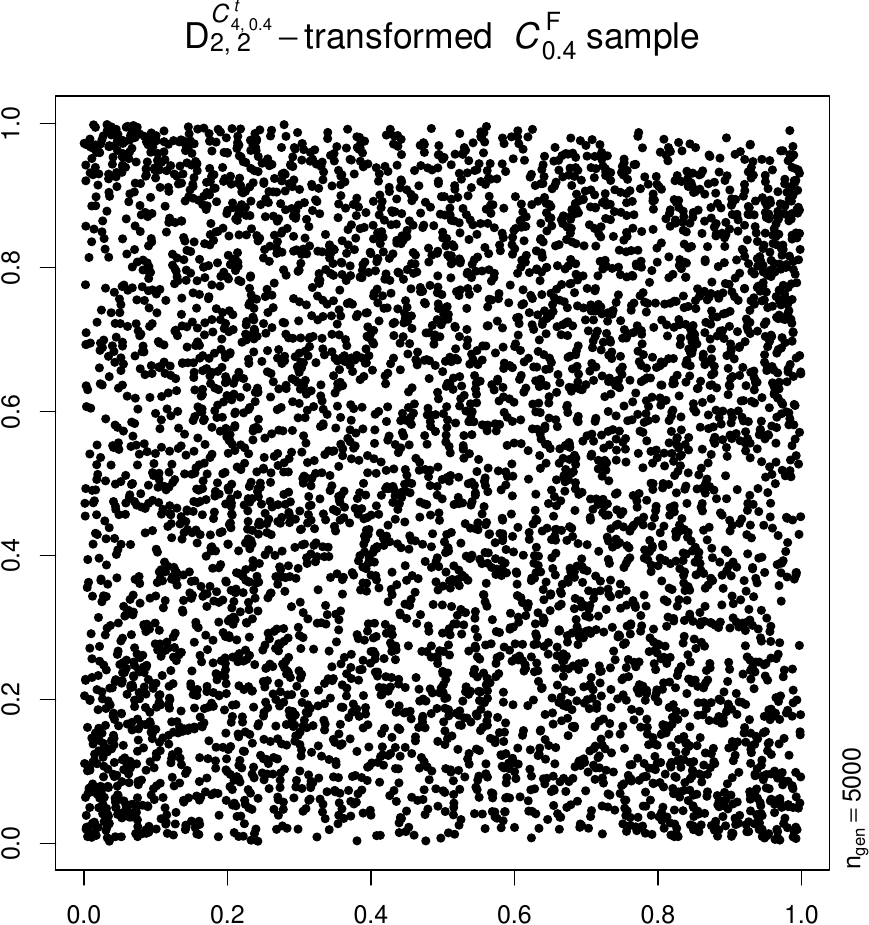}\hfill
  \includegraphics[width=0.19\textwidth]{fig_scatter_cop_dim_2_out_dim_2_ngen_5000_t4_tau_0.4_C_tau_0.4.pdf}
  \\[2mm]
  \includegraphics[width=0.19\textwidth]{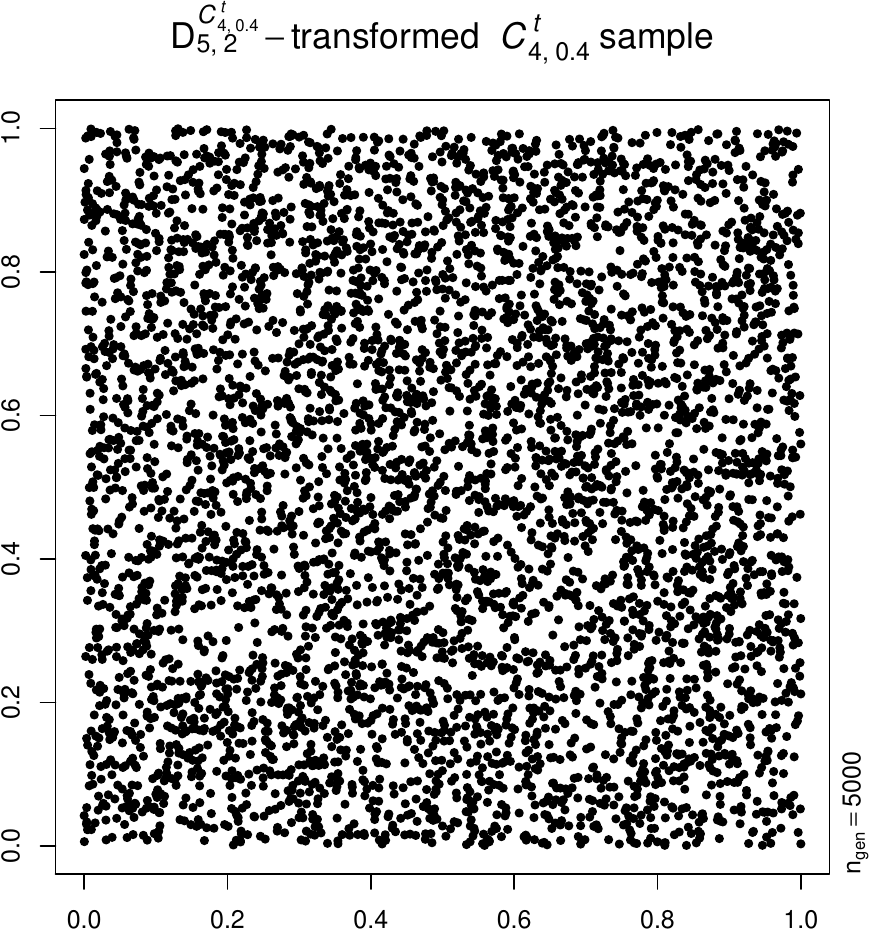}\hfill
  \includegraphics[width=0.19\textwidth]{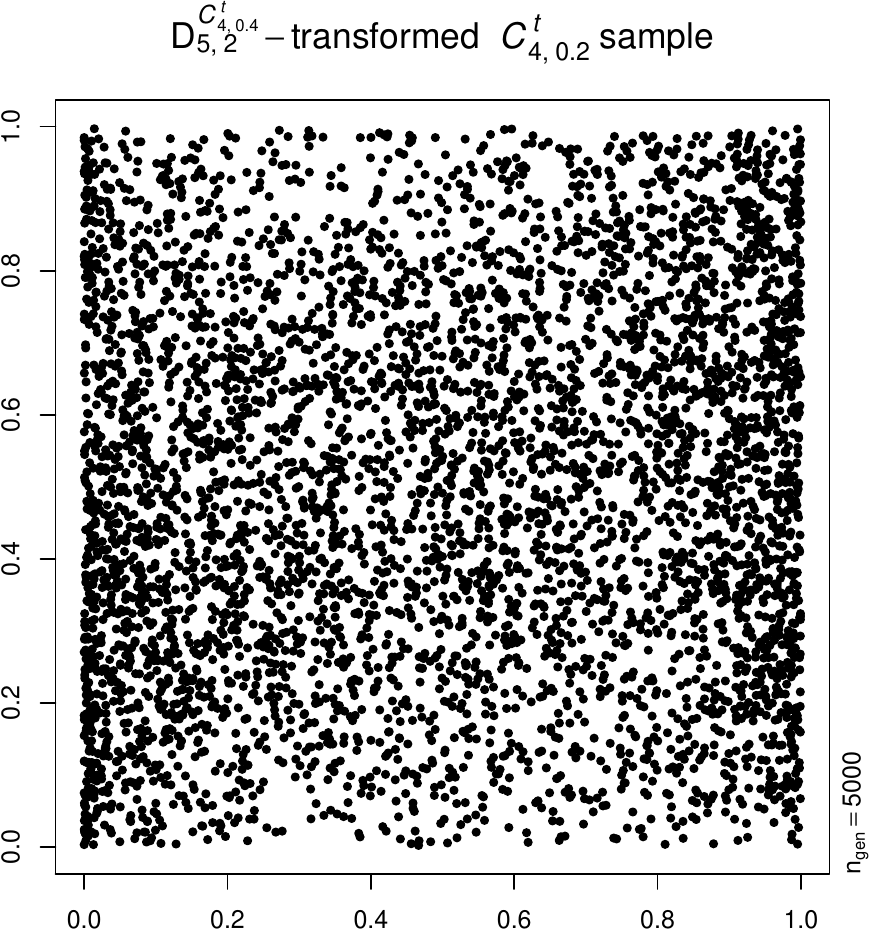}\hfill
  \includegraphics[width=0.19\textwidth]{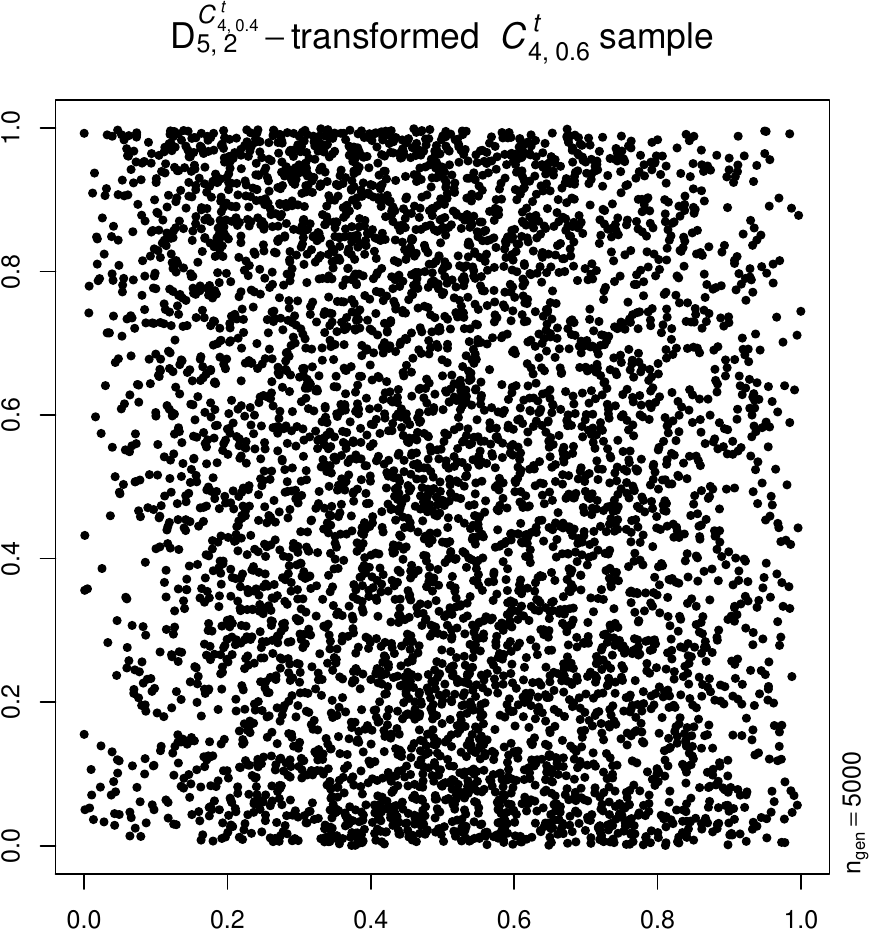}\hfill
  \includegraphics[width=0.19\textwidth]{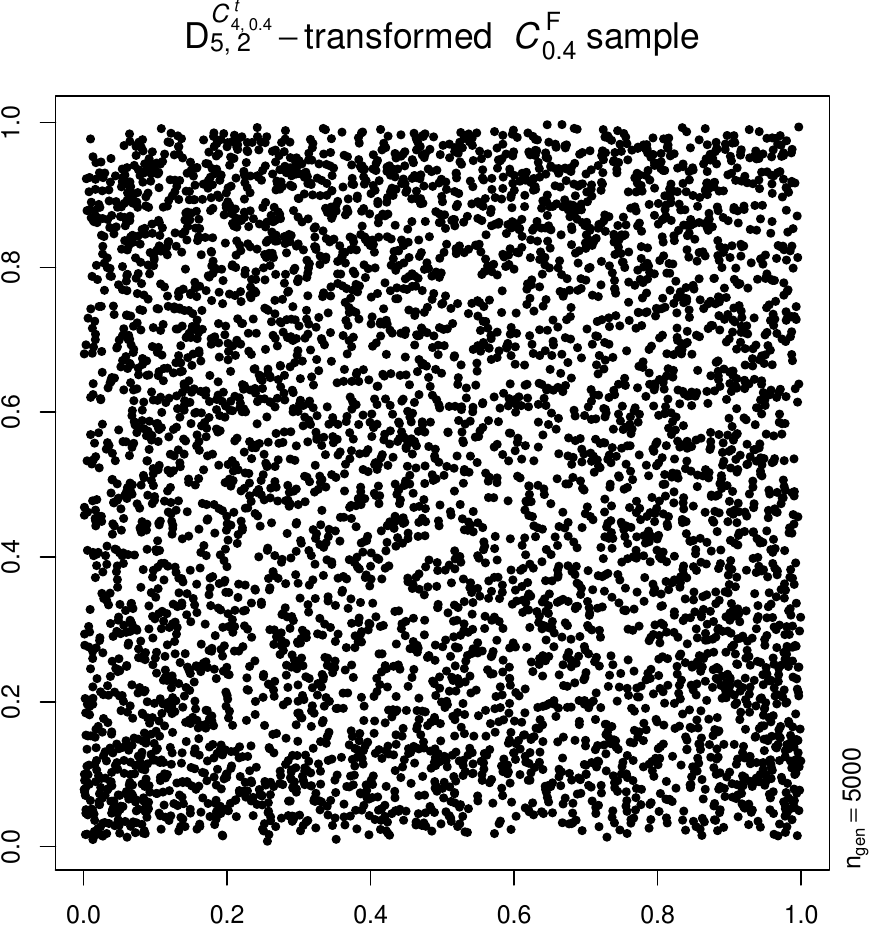}\hfill
  \includegraphics[width=0.19\textwidth]{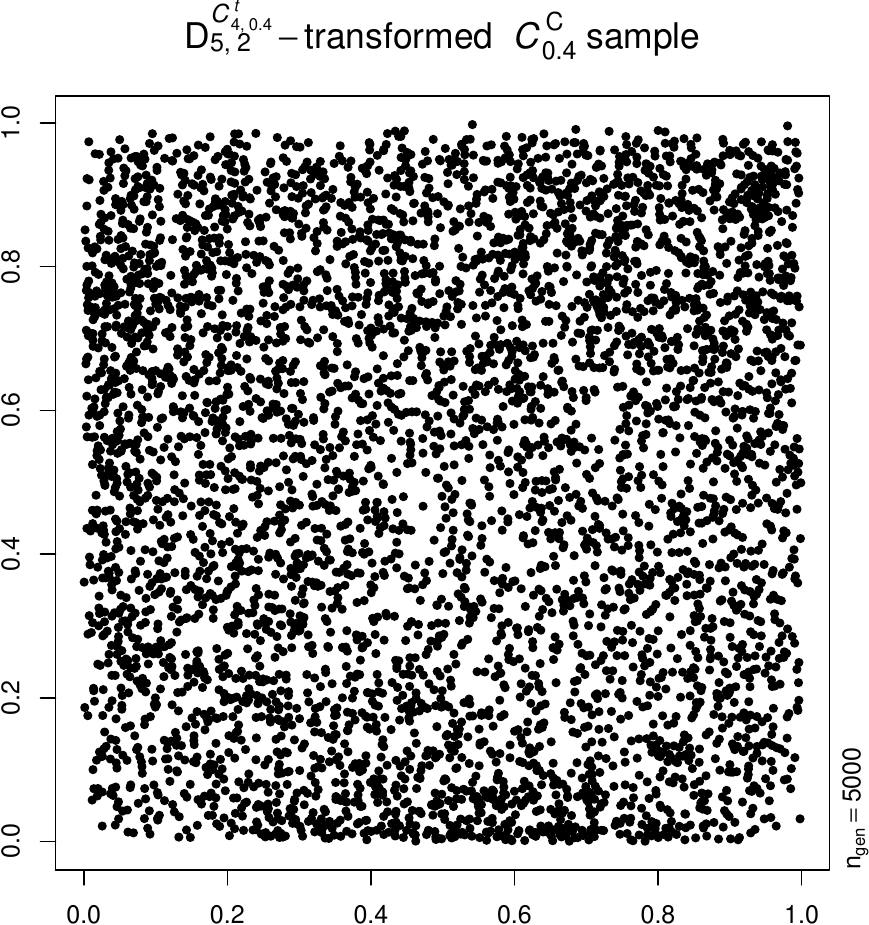}
  \\[2mm]
  \includegraphics[width=0.19\textwidth]{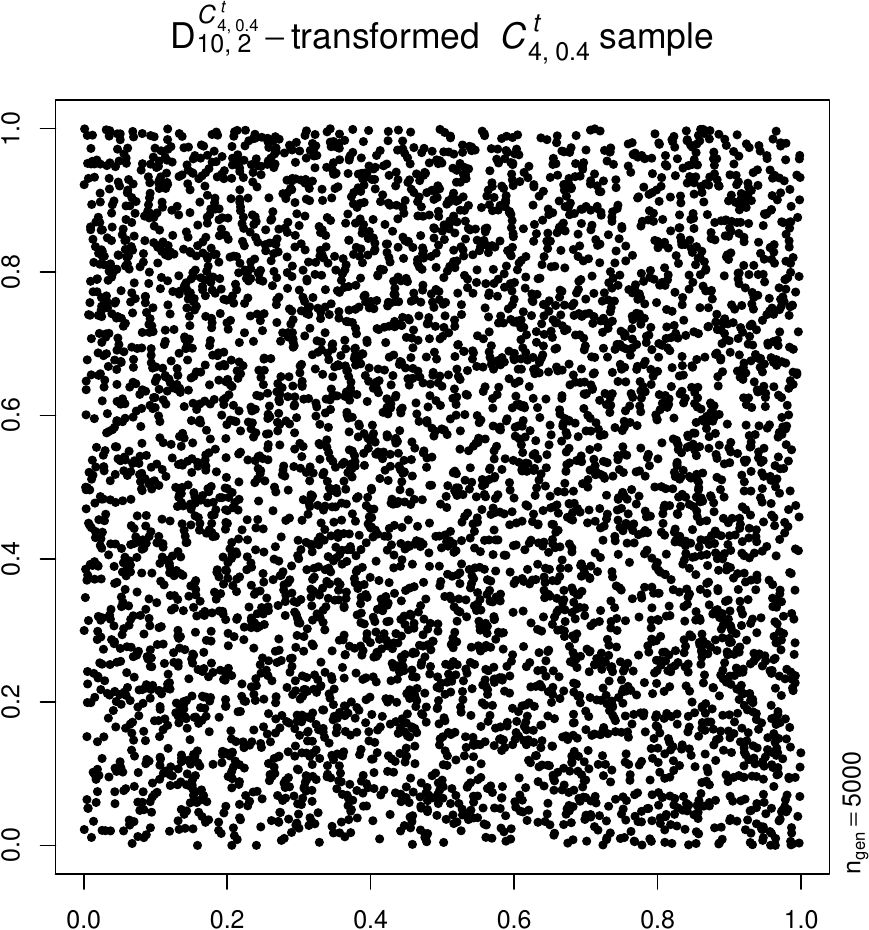}\hfill
  \includegraphics[width=0.19\textwidth]{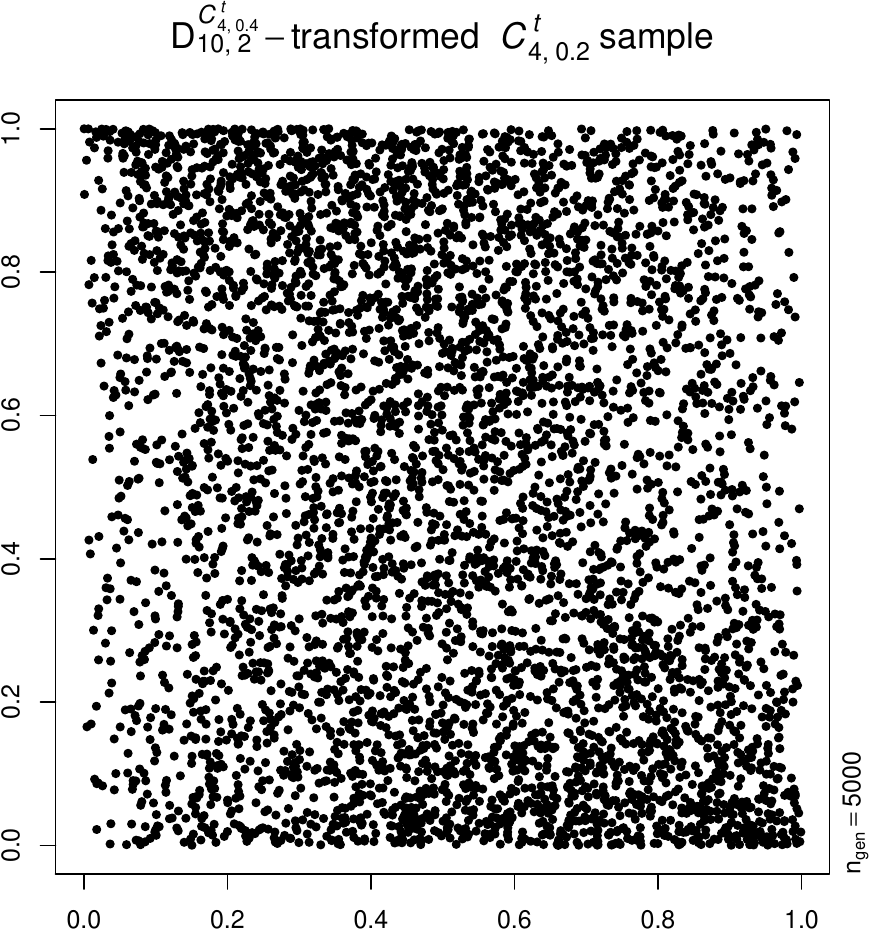}\hfill
  \includegraphics[width=0.19\textwidth]{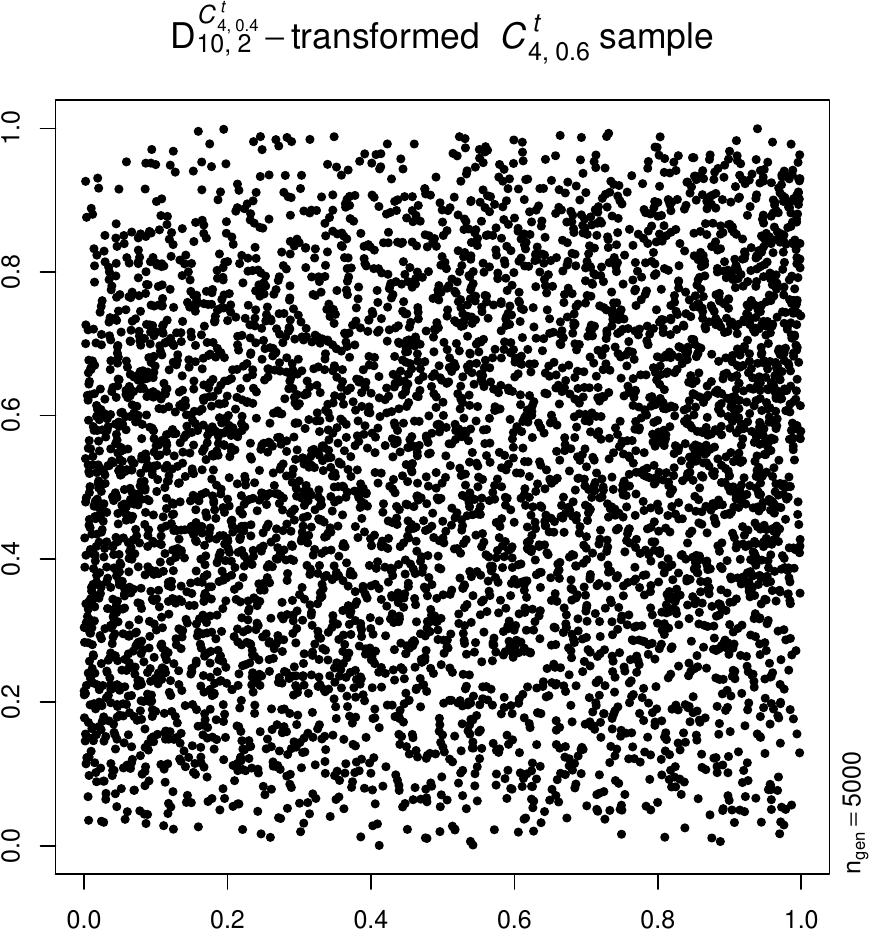}\hfill
  \includegraphics[width=0.19\textwidth]{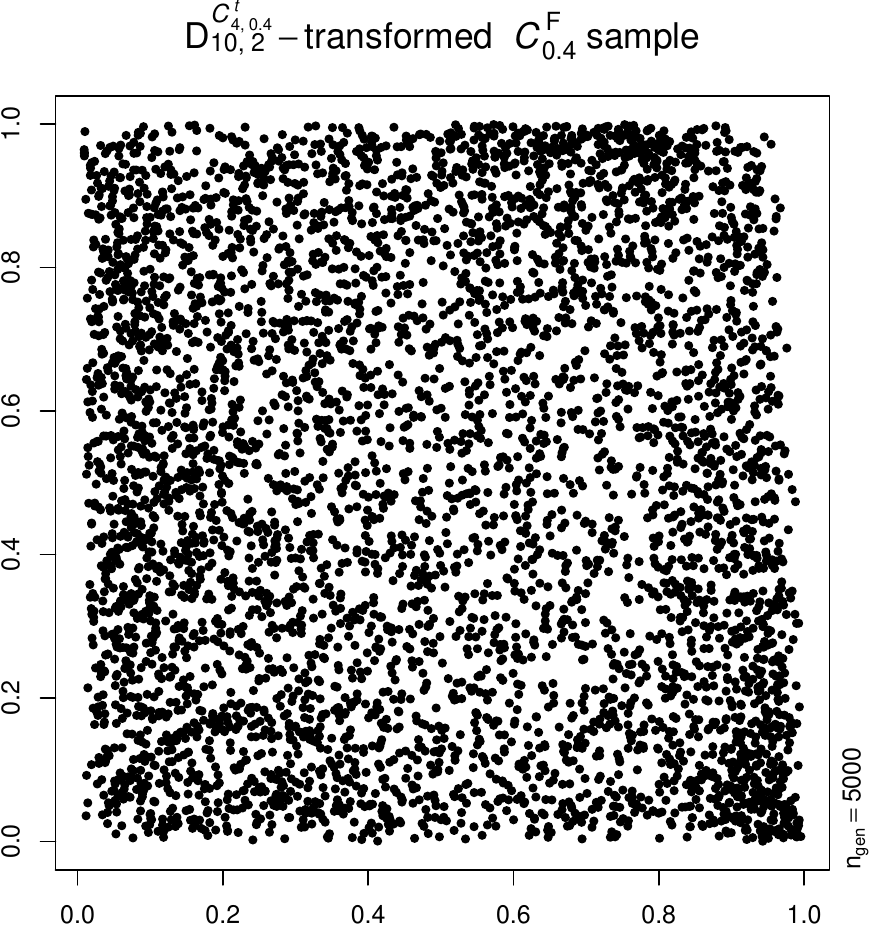}\hfill
  \includegraphics[width=0.19\textwidth]{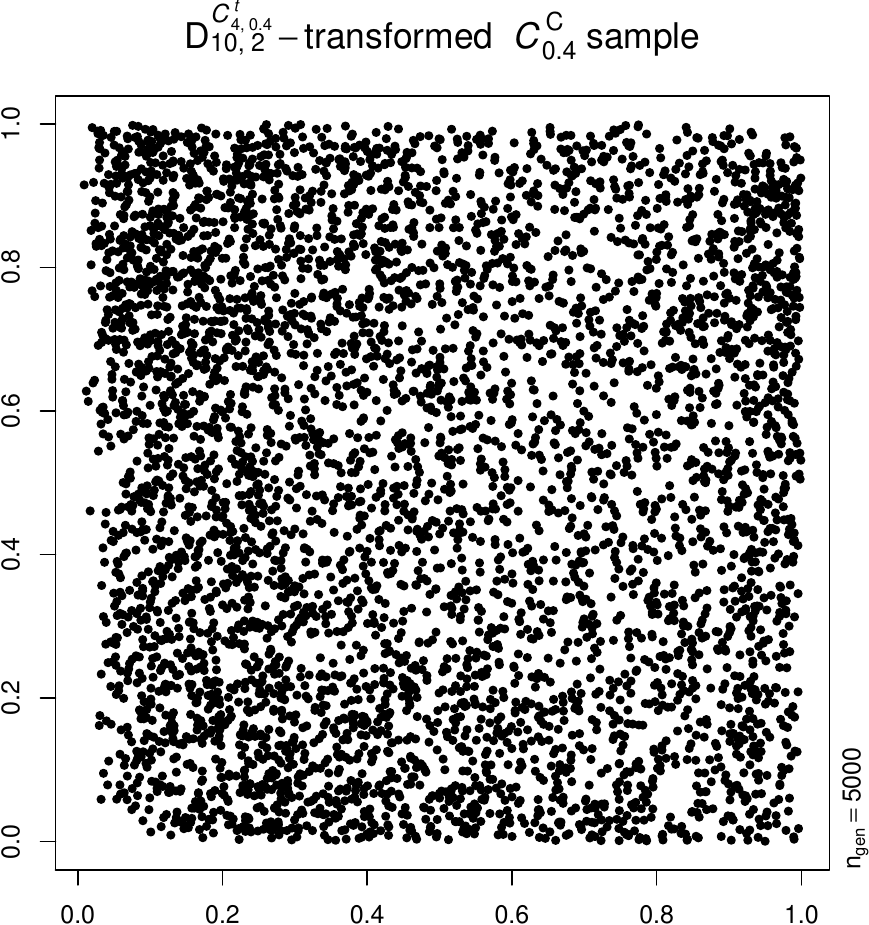}
  \caption{$D^{C^t_{4,0.4}}_{2,2}$-transformed samples of size $\ngen=5000$ from
    bivariate $C^t_{4,0.4}$, $C^t_{4,0.2}$, $C^t_{4,0.6}$, $C^{\text{F}}_{0.4}$
    and $C^{\text{C}}_{0.4}$ copulas (top row, from left to right),
    $D^{C^t_{4,0.4}}_{5,2}$-transformed samples of the same size and from the same
    type of copulas but five-dimensional (middle row), and
    $D^{C^t_{4,0.4}}_{10,2}$-transformed samples of the same size and from the same
    type of copulas but ten-dimensional (bottom row).}
  \label{fig:scatter:D:2:5:10:to:2}
\end{figure}

We first focus on the graphical assessment and selection approach.
Figure~\ref{fig:scatter:D:2:5:10:to:2} shows DecoupleNet-transformed samples
from different copulas (columns) and from different dimensions (rows).
Let us start by focusing on the first row. Here a DecoupleNet was trained on a
sample of size $\ntrn=50\,000$ from a bivariate $C^t_{4,0.4}$ copula. The
resulting DecoupleNet is $D^{C^t_{4,0.4}}_{2,2}$.  Then samples of size
$\ngen=5000$ from $C^t_{4,0.4}$ (so the same copula as what the DecoupleNet was trained
on, referred to as the \emph{true} copula), from $C^t_{4,0.2}$, $C^t_{4,0.6}$
(so also $t$ copulas with the same degrees of freedom but different Kendall's
tau), from $C^{\text{F}}_{0.4}$ (the Archimedean Frank copula with Kendall's tau
$0.4$) and from $C^{\text{C}}_{0.4}$ copulas are generated and each is passed
through the DecoupleNet $D^{C^t_{4,0.4}}_{2,2}$ and then plotted in the first
row of Figure~\ref{fig:scatter:D:2:5:10:to:2} (from left to right). For the true
copula, so the sample from $C^t_{4,0.4}$, the
$D^{C^t_{4,0.4}}_{2,2}$-transformed samples look uniform as they should. And for
all other candidate copulas, we clearly see non-uniformity in the
$D^{C^t_{4,0.4}}_{2,2}$-transformed samples.  The samples in the second and
third row of Figure~\ref{fig:scatter:D:2:5:10:to:2} are constructed similarly,
using the same candidate copulas but now in $d=5$ (middle row) and $d=10$
(bottom row) dimensions; the corresponding DecoupleNets trained are denoted by
$D^{C^t_{4,0.4}}_{5,2}$ (middle row) and $D^{C^t_{4,0.4}}_{10,2}$ (bottom row).
We come to the same conclusion as in the first row -- namely, that we correctly
observe uniformity in the first column and non-uniformity in all others. From
all plots showing departures from uniformity in
Figure~\ref{fig:scatter:D:2:5:10:to:2}, we can even see that, across the
dimensions $d\in\{2,5,10\}$, the type of non-uniformity remains roughly the same
within each column -- up to rotation by a multiple of 90 degrees, an
insignificant artifact stemming from the stochastic nature of our training
procedure. This observation shows that we do not lose much information when
mapping from $d>2$ to $d'=2$ for the purpose of model assessment and selection.

\begin{figure}[htbp]
  \includegraphics[width=0.19\textwidth]{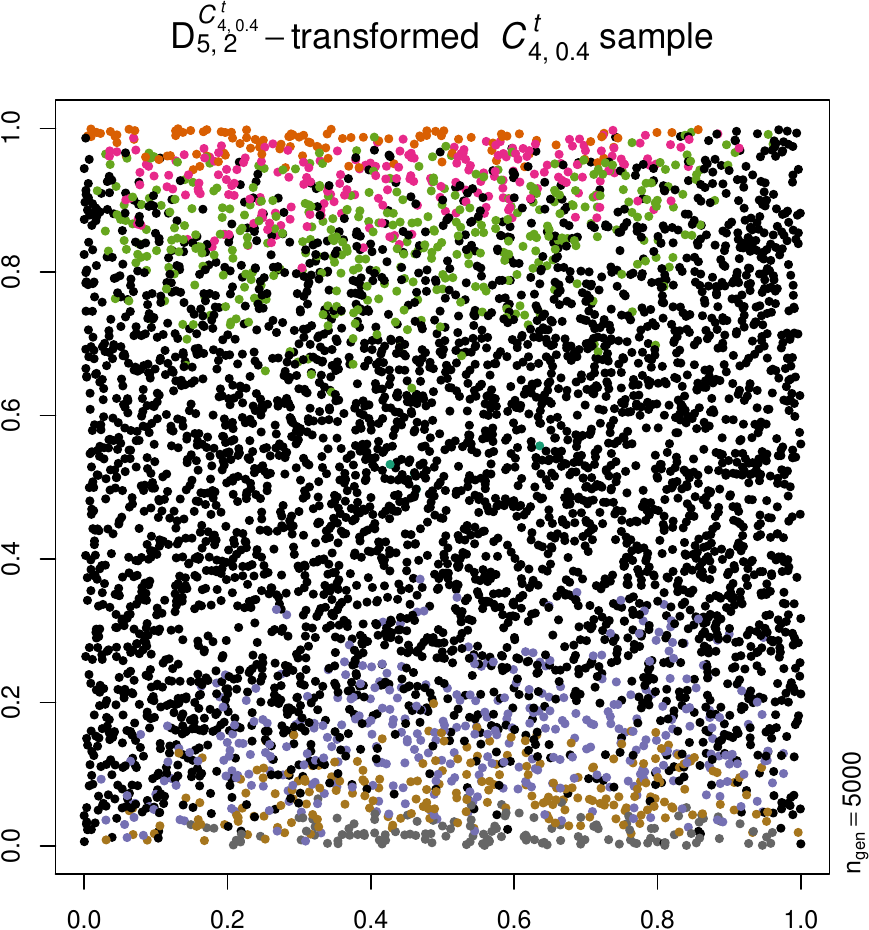}\hfill
  \includegraphics[width=0.19\textwidth]{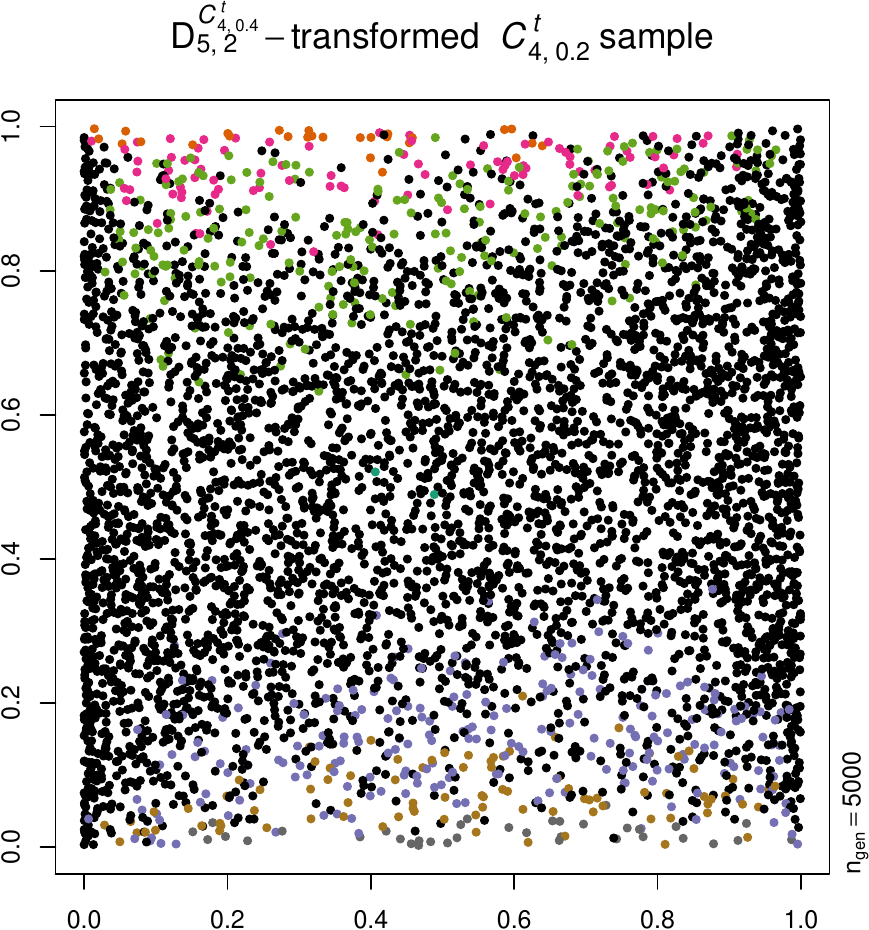}\hfill
  \includegraphics[width=0.19\textwidth]{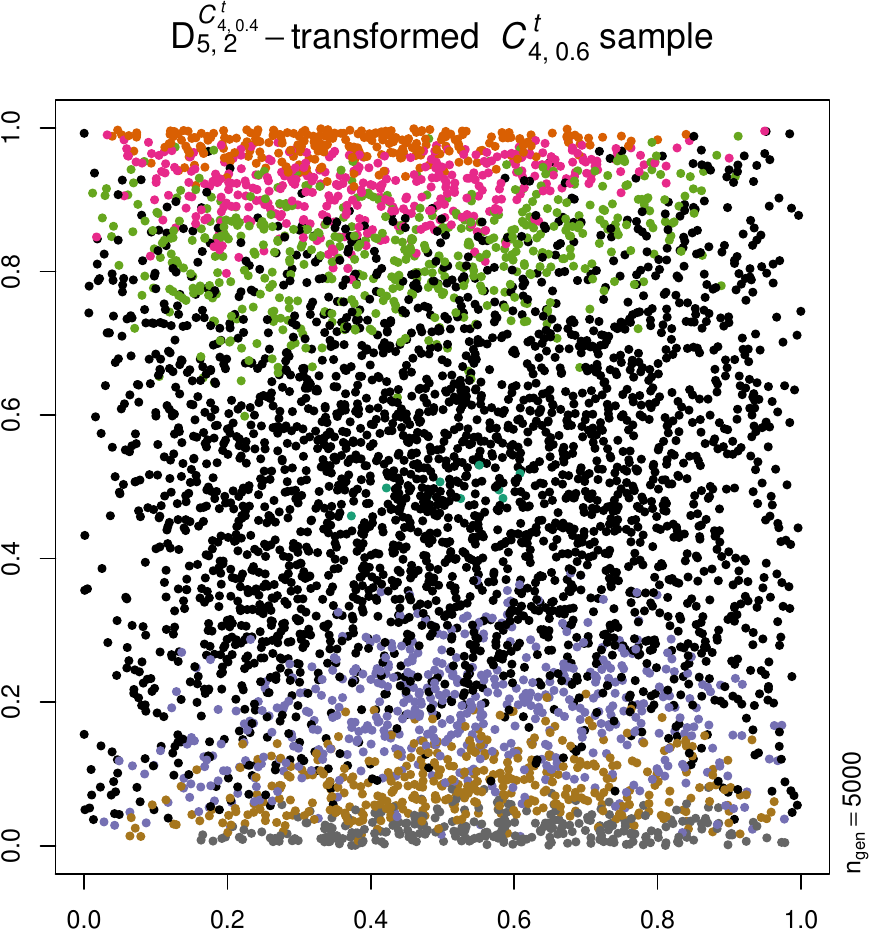}\hfill
  \includegraphics[width=0.19\textwidth]{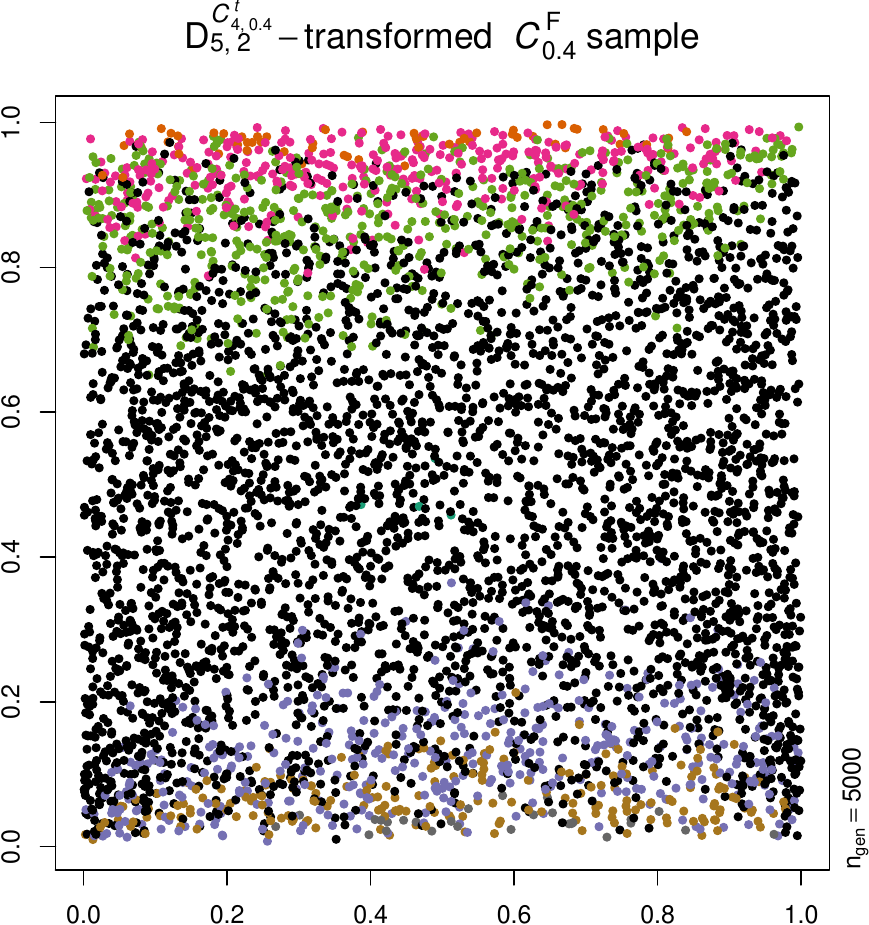}\hfill
  \includegraphics[width=0.19\textwidth]{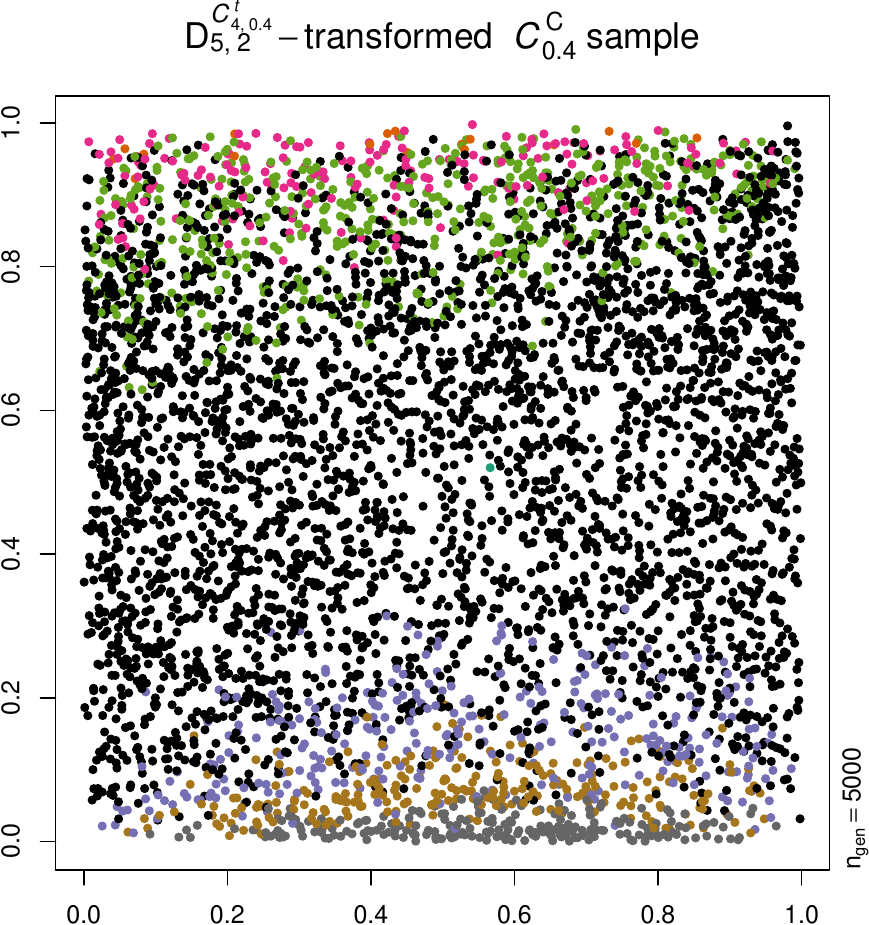}
  \caption{$D^{C^t_{4,0.4}}_{5,2}$-transformed colored samples of size
    $\ngen=5000$ from five-dimensional $C^t_{4,0.4}$, $C^t_{4,0.2}$,
    $C^t_{4,0.6}$, $C^{\text{F}}_{0.4}$ and $C^{\text{C}}_{0.4}$ copulas
    (from left to right). The plots correspond to the middle row of
    Figure~\ref{fig:scatter:D:2:5:10:to:2} but with colors.}
  \label{fig:scatter:D:5:to:2:col}
\end{figure}

Next, Figure~\ref{fig:scatter:D:5:to:2:col} shows the middle row of
Figure~\ref{fig:scatter:D:2:5:10:to:2} but with colored samples.
As also in the rest of the paper, we used the same color scheme here as we have
already seen in Figure~\ref{fig:col:ex}, so darker colors correspond to the
joint left tail and brighter colors to the joint right tail of the input sample
or copula. This allows us to assess the different five-dimensional candidate
copulas, and ultimately to select one of them, according to their ability to
properly capture, say, the joint right tail. For example, the
$D^{C^t_{4,0.4}}_{5,2}$-transformed samples from $C^t_{4,0.2}$ (second plot) and
$C^{\text{C}}_{0.4}$ (last plot) show too few bright points in the top region and
thus underestimate the joint right tail; this can also be verified numerically,
the probability to fall in $[6/7,1]^5$ is about $0.0673$ under $C^t_{4,0.4}$ but only
$0.0457$ under $C^t_{4,0.2}$ and $0.0400$ under $C^{\text{C}}_{0.4}$.
Similarly, the $D^{C^t_{4,0.4}}_{5,2}$-transformed sample from $C^t_{4,0.6}$
(third plot) shows too many points in the top region and thus overestimates the
joint right tail; the probability to fall in $[6/7,1]^5$ is about $0.0912$ under
$C^t_{4,0.6}$.
Selecting a model based on only the joint right tail region (an important region
for risk management applications, for example), we select $C^{\text{F}}_{0.4}$
(fourth plot); again this can be confirmed numerically, the probability to fall in
$[6/7,1]^5$ under $C^{\text{F}}_{0.4}$ is $0.0548$, which is closest to the
probability $0.0673$ of the true model among all other candidate models.

\begin{figure}[htbp]
  \includegraphics[width=0.32\textwidth]{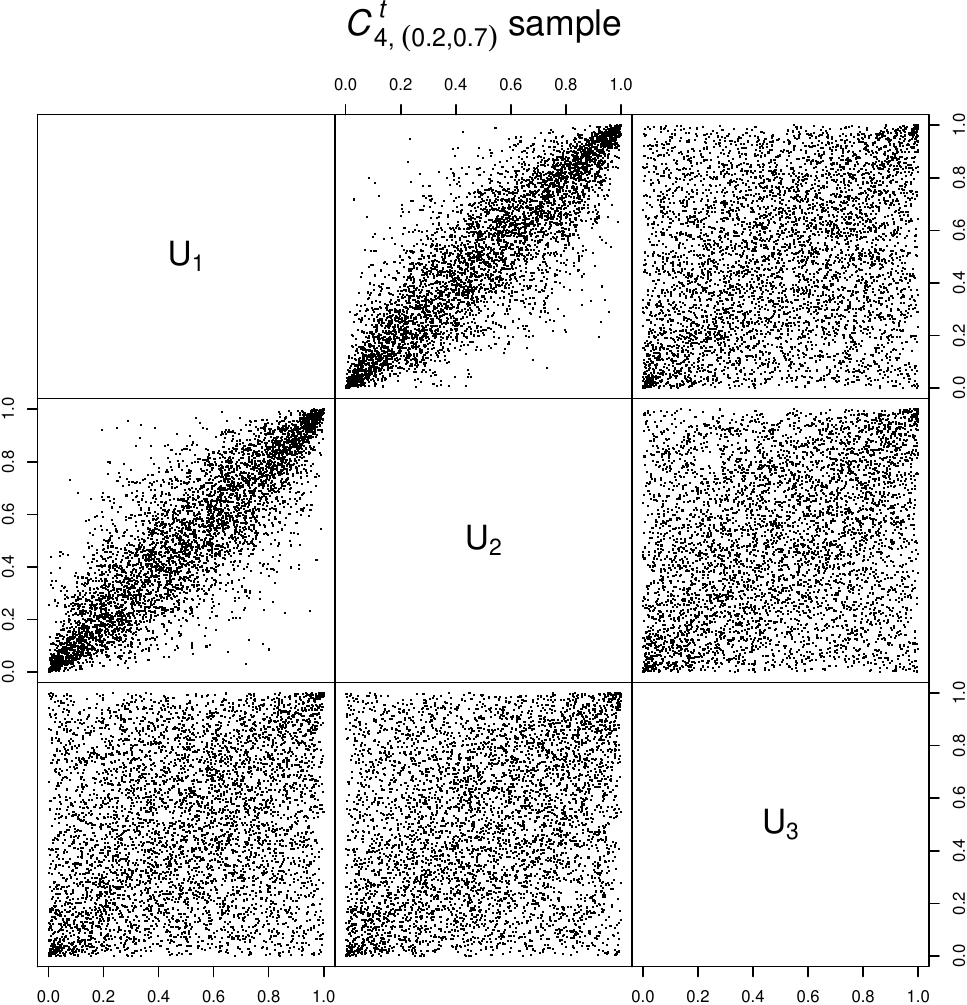}\hfill
  \includegraphics[width=0.32\textwidth]{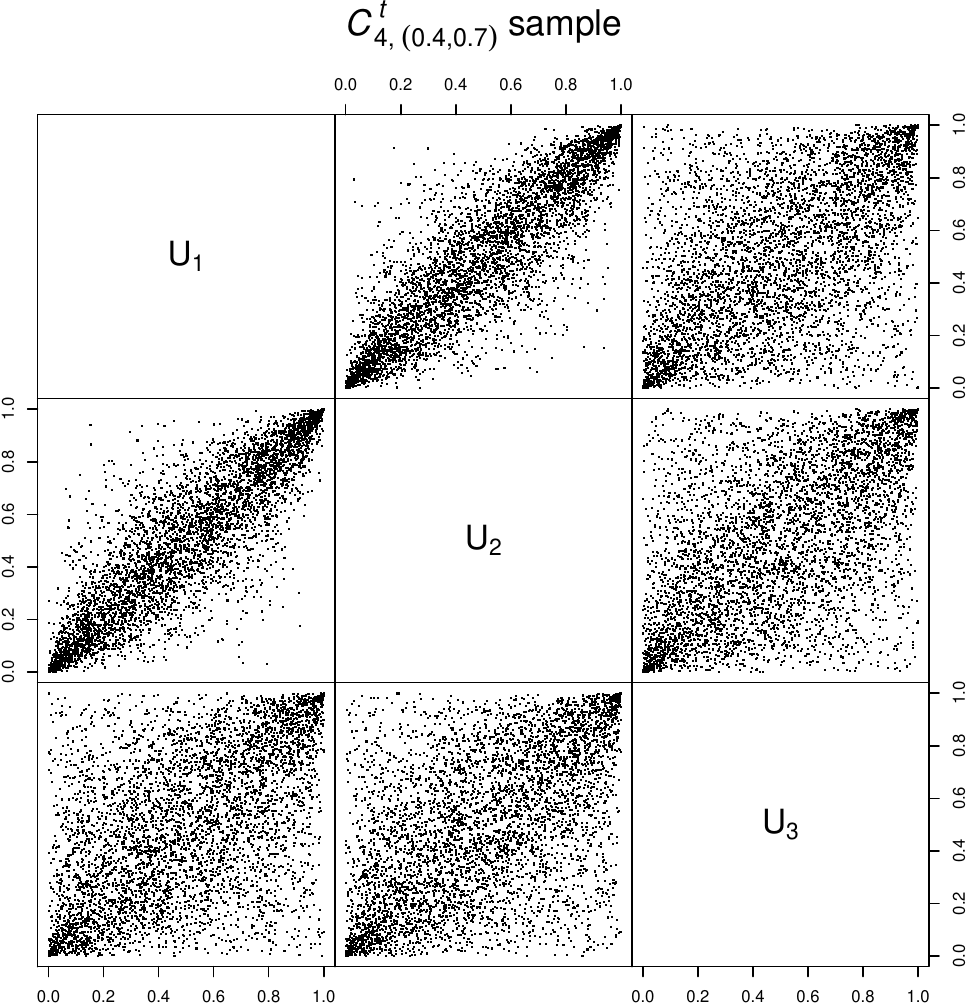}\hfill
  \includegraphics[width=0.32\textwidth]{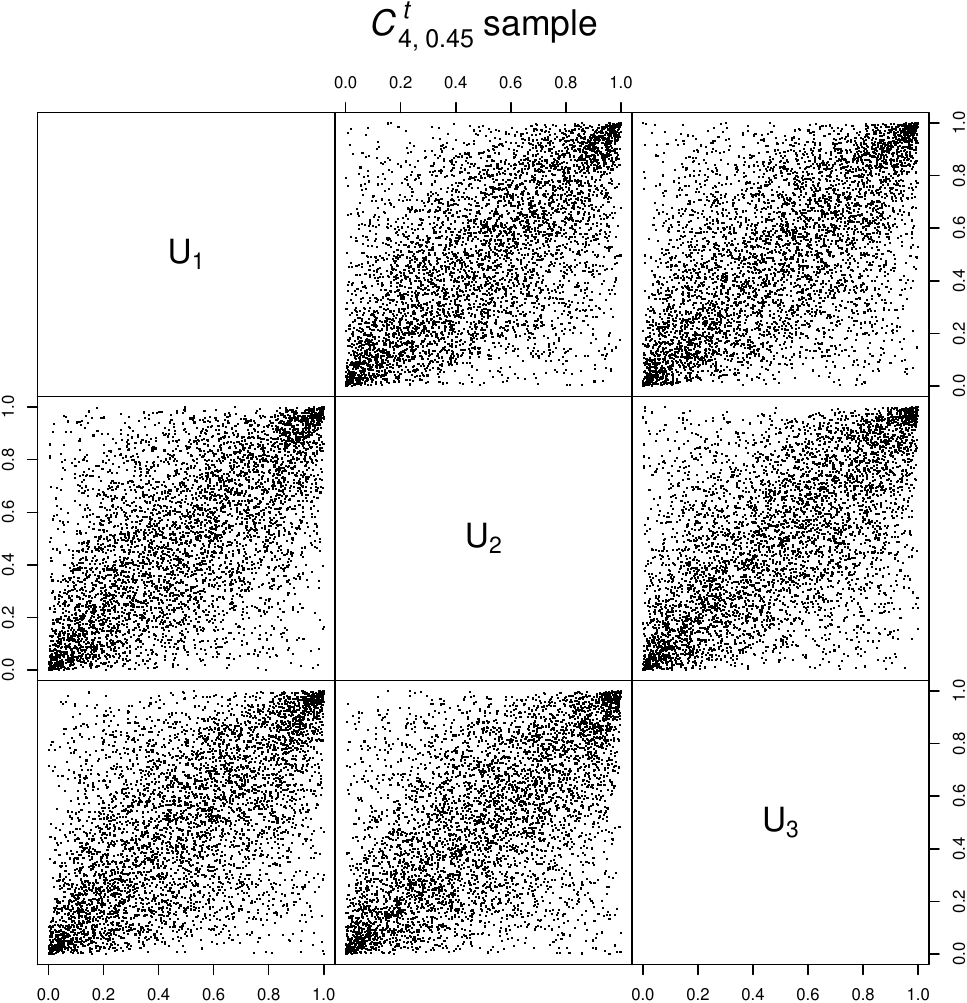}
  \\[4mm]
  \includegraphics[width=0.32\textwidth]{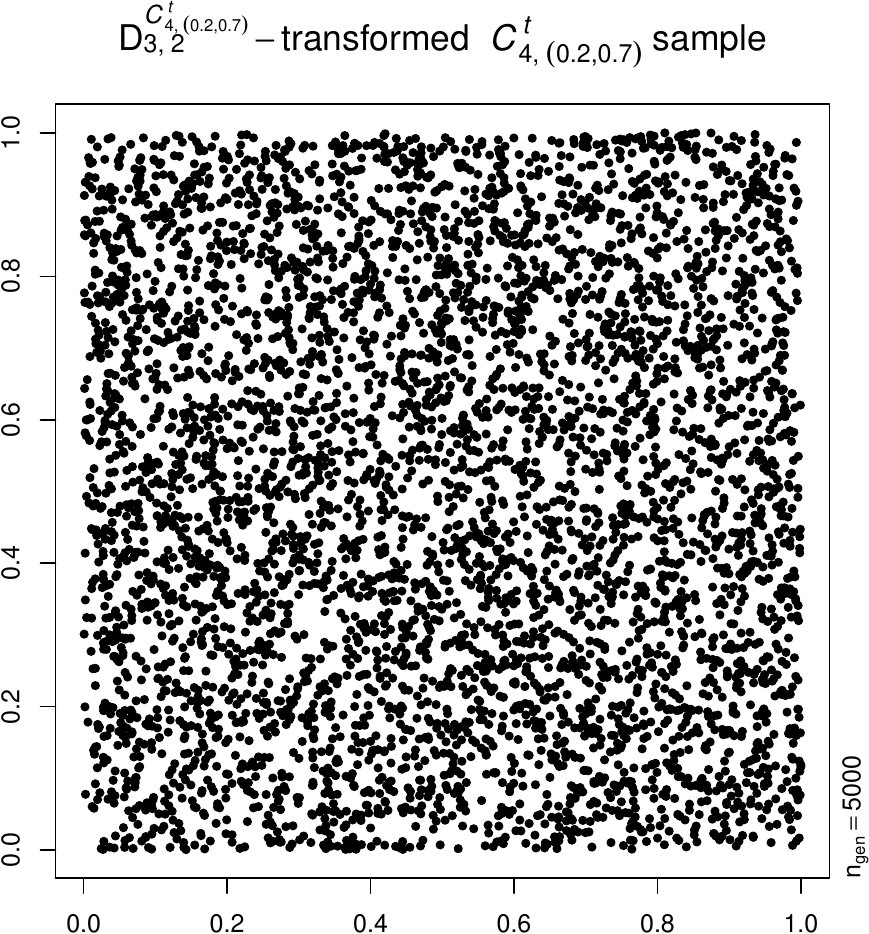}\hfill
  \includegraphics[width=0.32\textwidth]{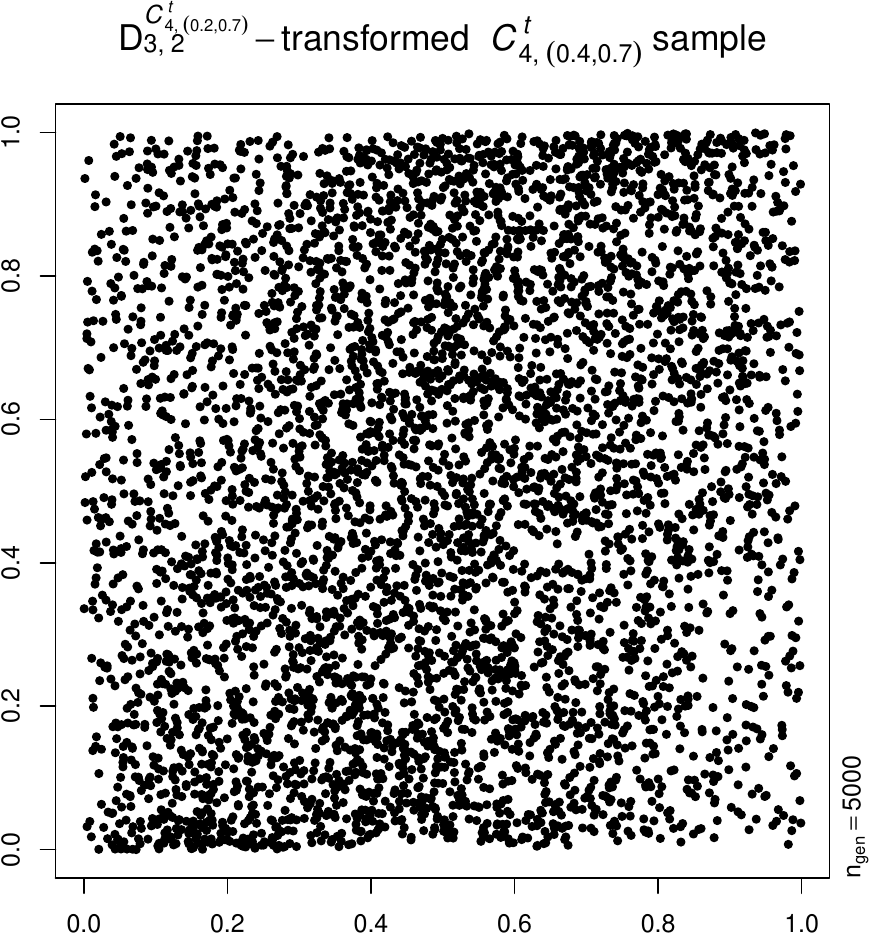}\hfill
  \includegraphics[width=0.32\textwidth]{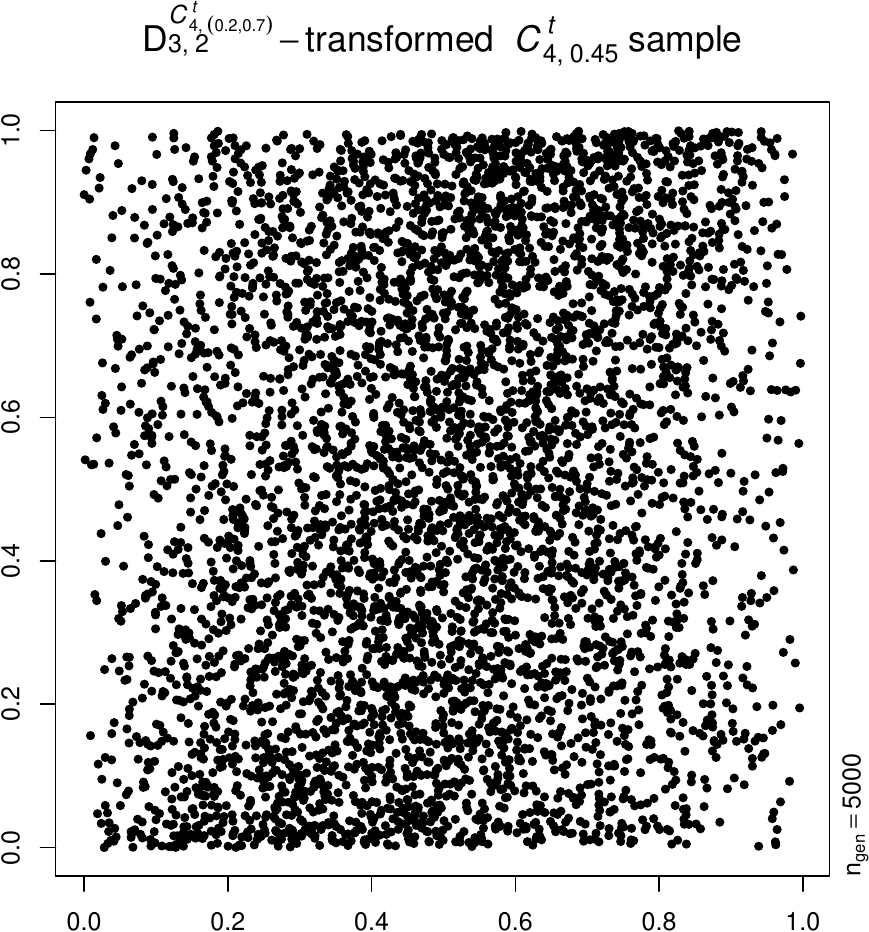}
  \caption{Samples of size $\ngen=5000$ from trivariate $C^t_{4,(0.2,0.7)}$,
    $C^t_{4,(0.4,0.7)}$ and $C^t_{4,0.45}$ copulas (top row, from left to right),
    with corresponding $D^{C^t_{4,(0.2,0.7)}}_{3,2}$-transformed samples (bottom
    row).}
  \label{fig:scatter:nested:3d:D:3:to:2}
\end{figure}

In our next example we consider deviations from the true $t$ copula in some entries
of the underlying correlation matrix $P$. To this end we use trivariate $t$ copulas
with $\nu=4$ degrees of freedom and correlation matrices $P$ of hierarchical
nature. The top row of Figure~\ref{fig:scatter:nested:3d:D:3:to:2} shows scatter
plot matrices of the trivariate samples with size $\ngen=5000$ from these models,
denoted by $C^t_{4,(0.2, 0.7)}$, $C^t_{4,(0.4, 0.7)}$ and $C^t_{4, 0.45}$ (from
left to right).
The notation for the former two models is $C^t_{4,(\tau_0, \tau_1)}$, where
$\tau_0$ is the Kendall's tau corresponding to the entries $\rho_{13},\rho_{23}$
or the correlation matrix $P$ of the $t$ copula, whereas $\tau_1$ corresponds to
the entry $\rho_{12}$ of $P$; note that for $t$ copulas, one has
$\rho=\sin(\tau\pi/2)$. The bottom row of
Figure~\ref{fig:scatter:nested:3d:D:3:to:2} shows scatter plots of the
$D^{C^t_{4,(0.2,0.7)}}_{3,2}$-transformed samples of $C^t_{4,(0.2, 0.7)}$ (the
true copula here), $C^t_{4,(0.4, 0.7)}$ (deviating in $\tau_0$, so in
$\rho_{13},\rho_{23}$) and $C^t_{4, 0.45}$ (deviating in all entries of $P$ but
capturing the average Kendall's tau $(0.2+0.7)/2$). As before, also here we
correctly see uniformity in the first, and non-uniformity in the other two
plots.

\subsection{Numerical summary}\label{sec:num:approach}
Despite the drawbacks of using just a numerical summary for model assessment and
  selection (Remark~\ref{rem:graph:vs:num}), in this section we still investigate it further,
  largely because it is much easier to report replications for single numeric summaries than it is
  for graphical assessments. The following algorithm summarizes what we do
in this section for various dependence models to be specified later.
\begin{algorithm}[Numerical model assessment and selection based on simulated
  data]\label{algo:num:assess}
  \begin{enumerate}
  \item Fix a $d$-dimensional copula $C$ and a number $B\in\IN$ of replications.
  \item For $b=1,\dots,B$ do:
    \begin{enumerate}
    \item Generate a sample of size $\ntrn$ from $C$ and compute its
      pseudo-observations $\{\hat{\bm{U}}^{(b)}_{i}\}_{i=1}^{\ntrn}$; we use
      pseudo-observations here to mimic a realistic scenario as would be the
      case for real world data.
    \item Train the DecoupleNet $D^{C}_{d,2}$ on the pseudo-observations
      $\{\hat{\bm{U}}^{(b)}_{i}\}_{i=1}^{\ntrn}$.
    \item For the true copula $C$ and each candidate copula $\tilde{C}$, do:
      \begin{enumerate}
      \item If the copula contains unknown parameters, estimate them using the
        pseudo-observations $\{\hat{\bm{U}}^{(b)}_{i}\}_{i=1}^{\ntrn}$. %
      \item Generate a sample $\{\tilde{\bm{U}}_i\}_{i=1}^{\ngen}$ of size $\ngen$ from the (fitted) copula.
      \item Pass $\{\tilde{\bm{U}}_i\}_{i=1}^{\ngen}$ through the trained DecoupleNet
        $D^{C}_{d,2}$ and obtained the decoupled output sample.
      \item Evaluate the decoupled output sample by computing the CvM
        score~\eqref{CvM:score}.
      \end{enumerate}
    \end{enumerate}
  \item Create box plots of the computed CvM scores.
  \end{enumerate}
\end{algorithm}
Where applicable, we also include a comparison with
  Rosenblatt-transformed data and box plots of sliced Wasserstein distances (introduced below).

We apply Algorithm~\ref{algo:num:assess} in three settings. In all three we
consider $d\in\{3,5,10\}$, $B=25$, $\ntrn=50\,000$ and $\ngen=10\,000$.  In the
first and third setting, the copulas were chosen among the few with analytically
available Rosenblatt transform to allow for a comparison.

\begin{figure}[htbp]
  \hspace*{\fill}
  \includegraphics[width=0.30\textwidth]{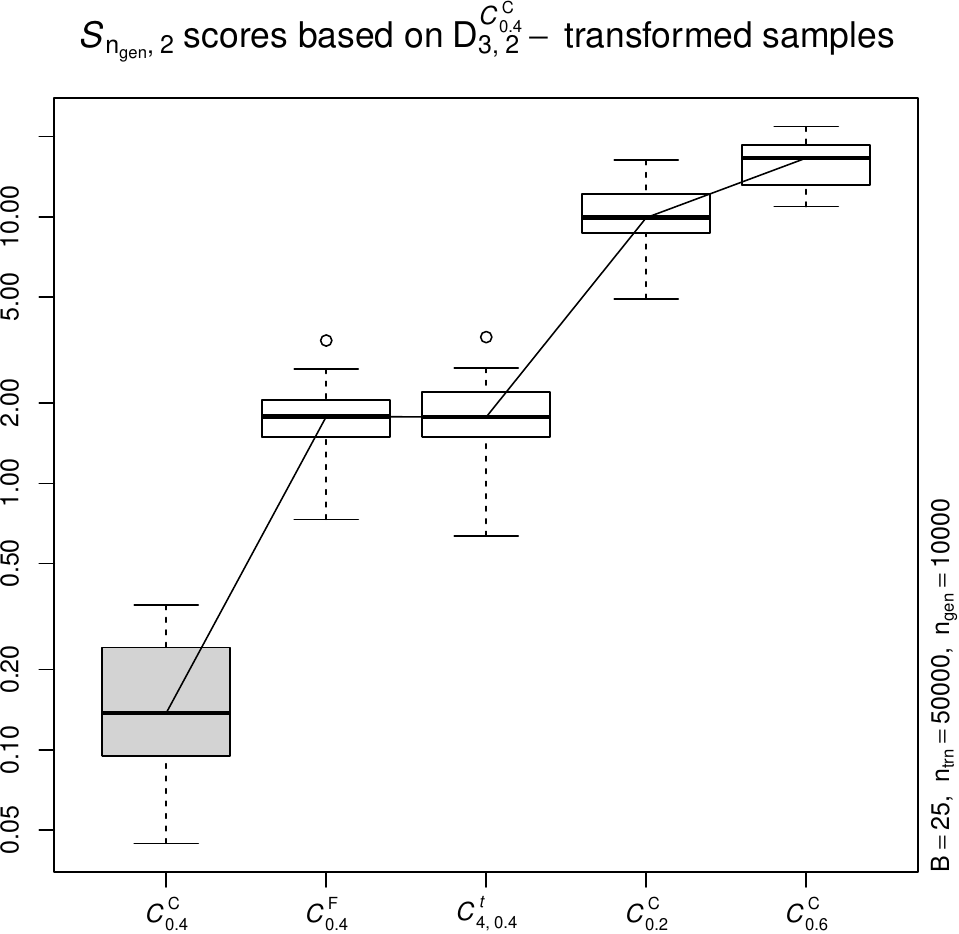}\hfill
  \includegraphics[width=0.30\textwidth]{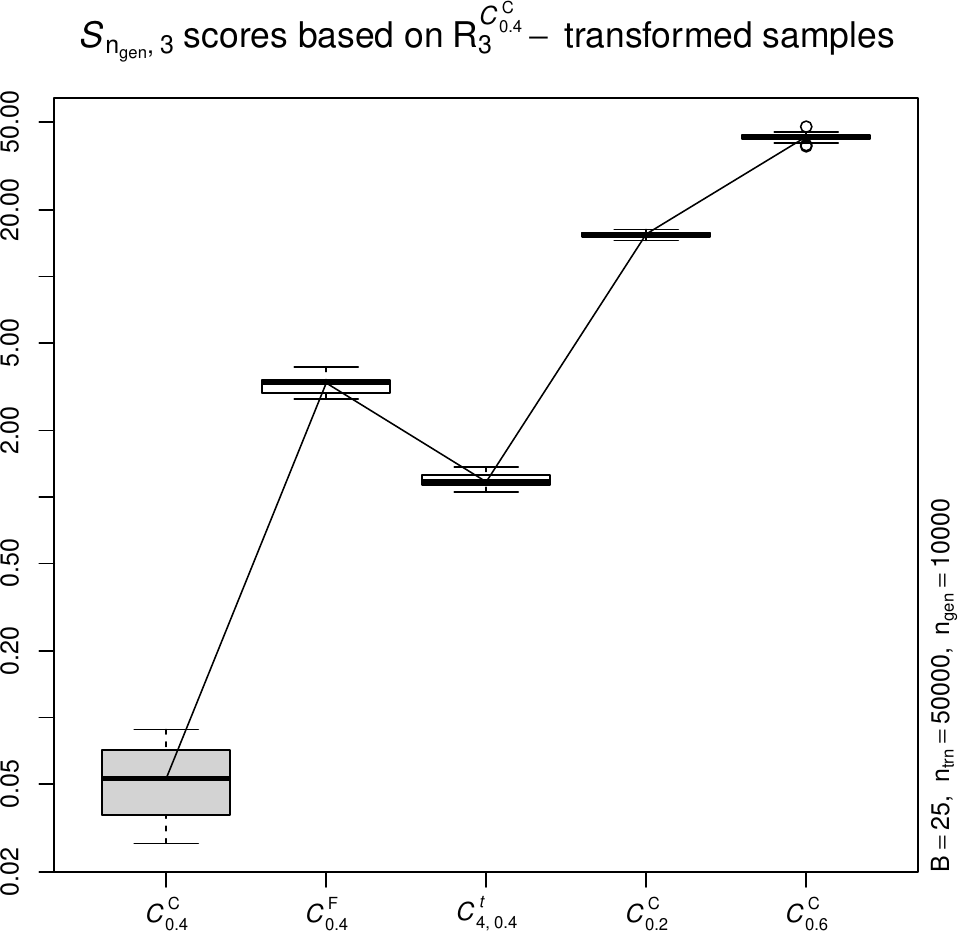}\hfill
  \includegraphics[width=0.30\textwidth]{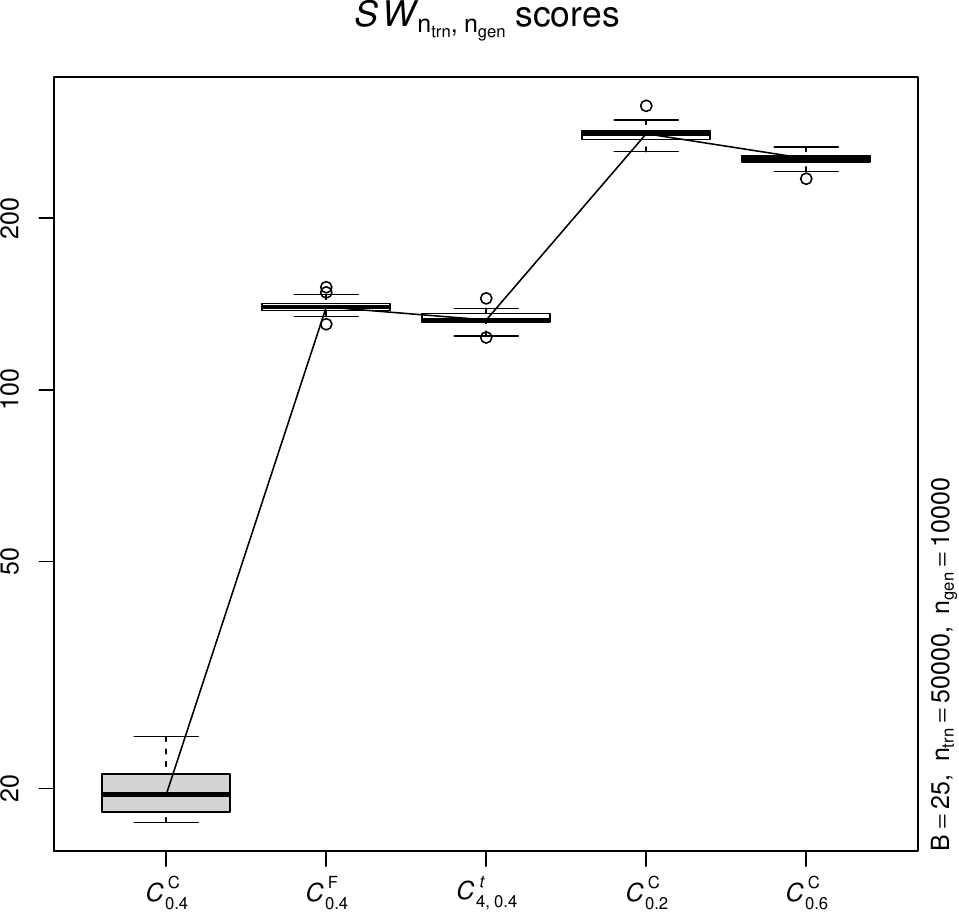}\hspace*{\fill}
  \\[4mm]
  \hspace*{\fill}
  \includegraphics[width=0.30\textwidth]{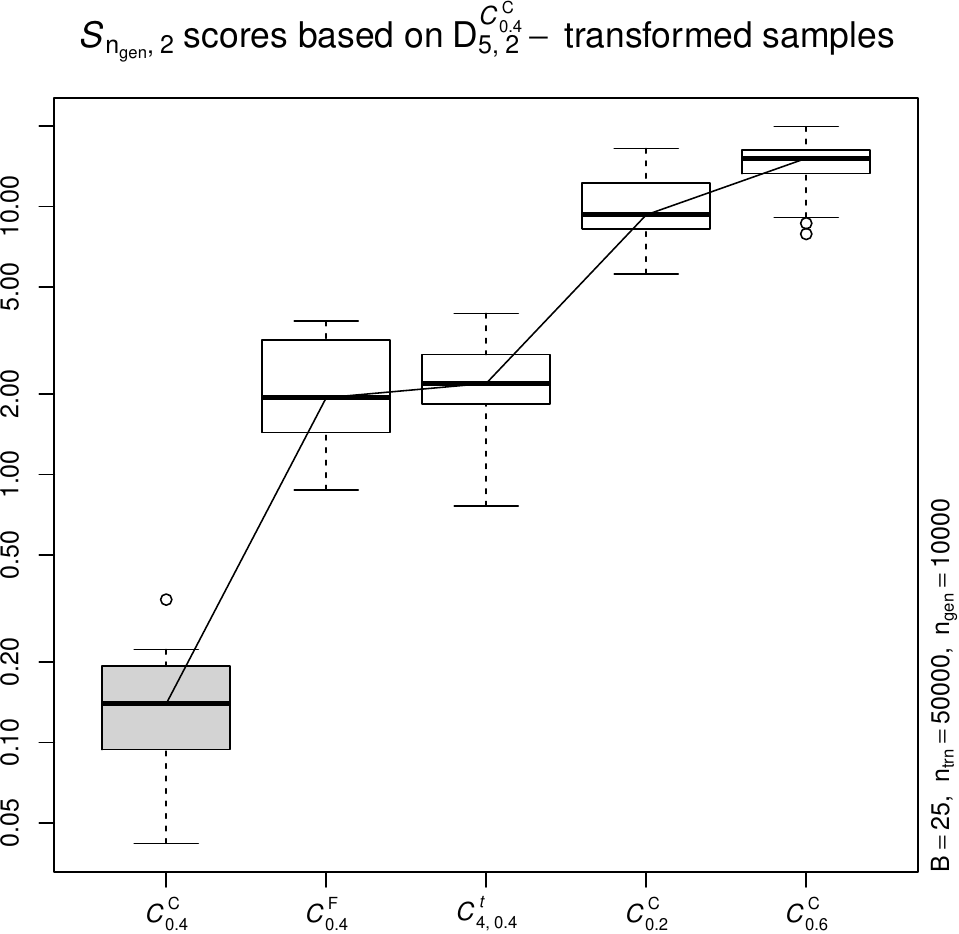}\hfill
  \includegraphics[width=0.30\textwidth]{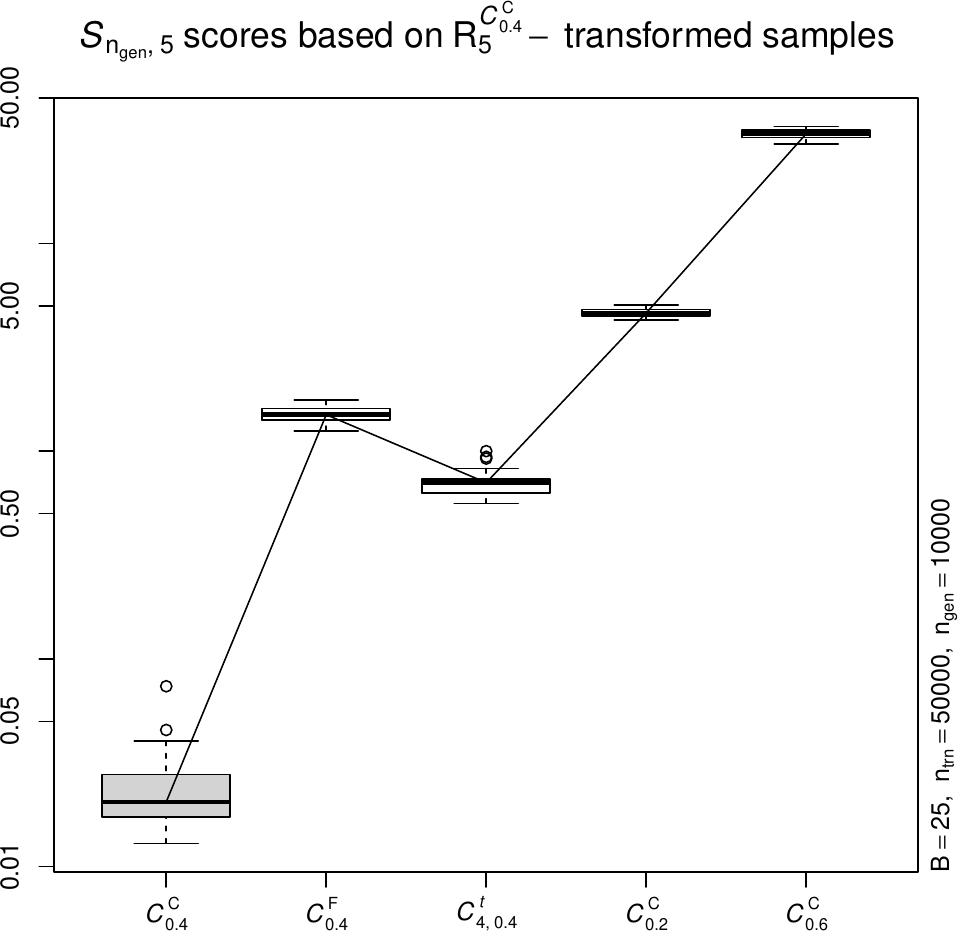}\hfill
  \includegraphics[width=0.30\textwidth]{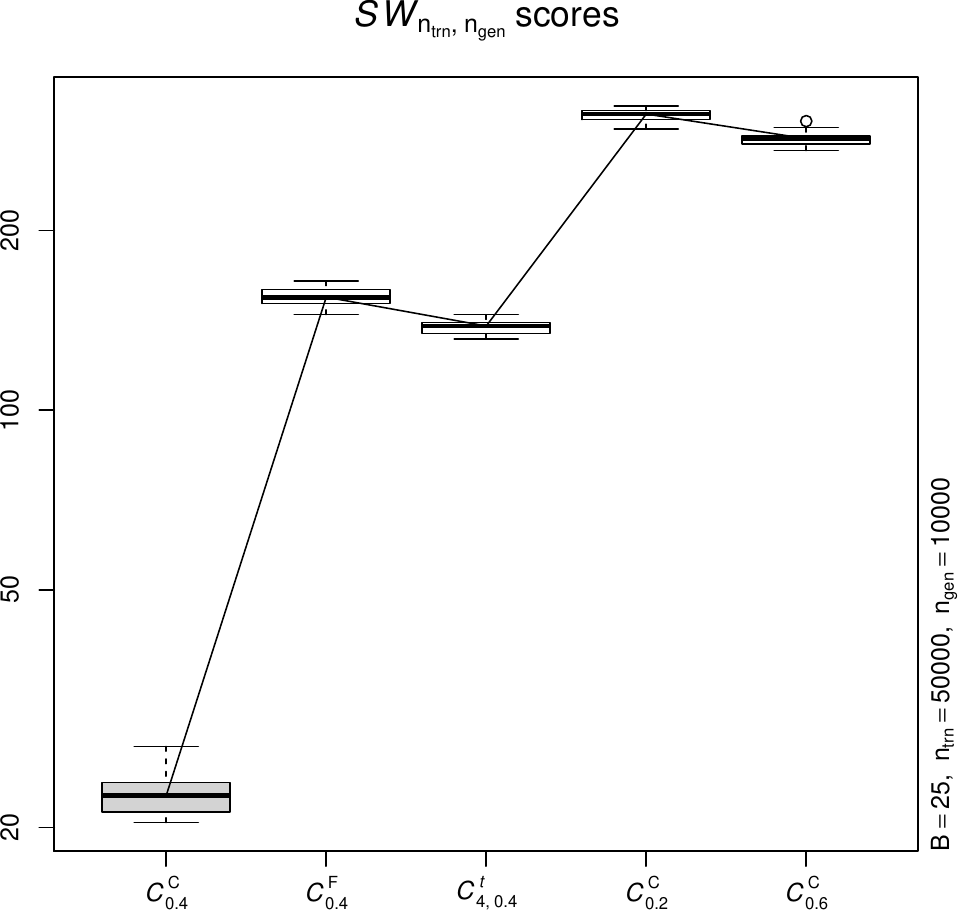}\hspace*{\fill}
  \\[4mm]
  \hspace*{\fill}
  \includegraphics[width=0.30\textwidth]{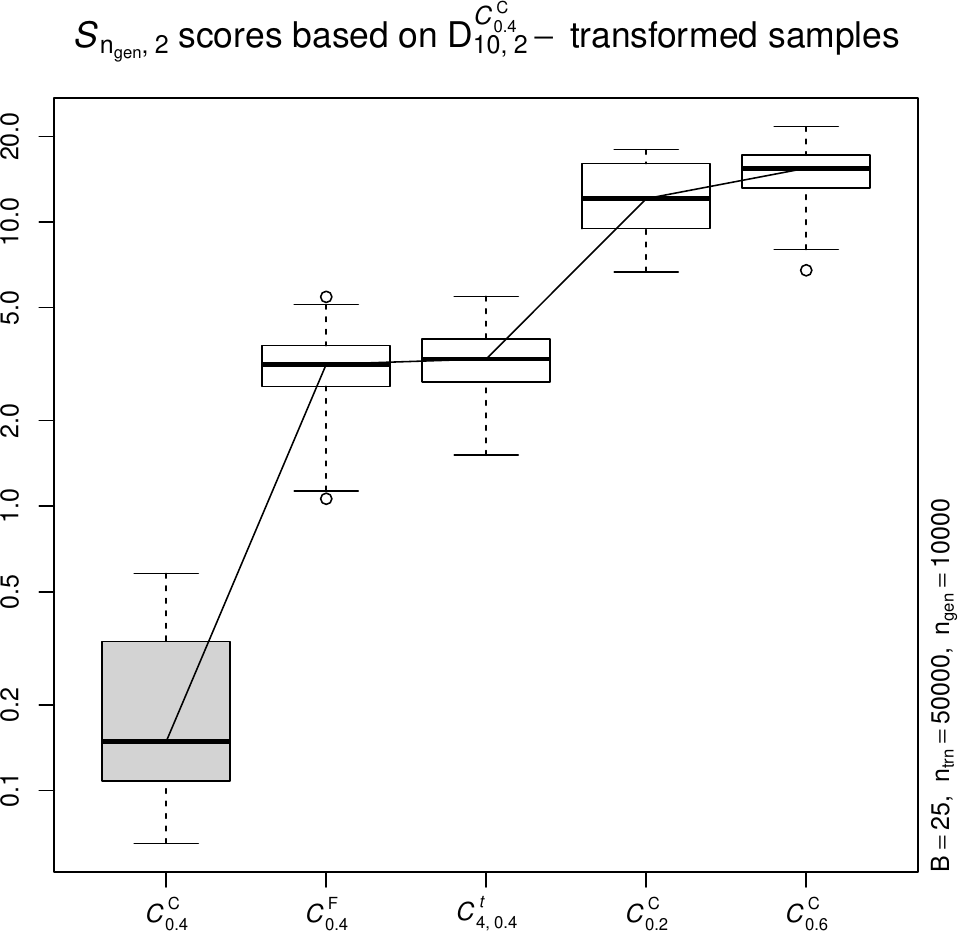}\hfill
  \includegraphics[width=0.30\textwidth]{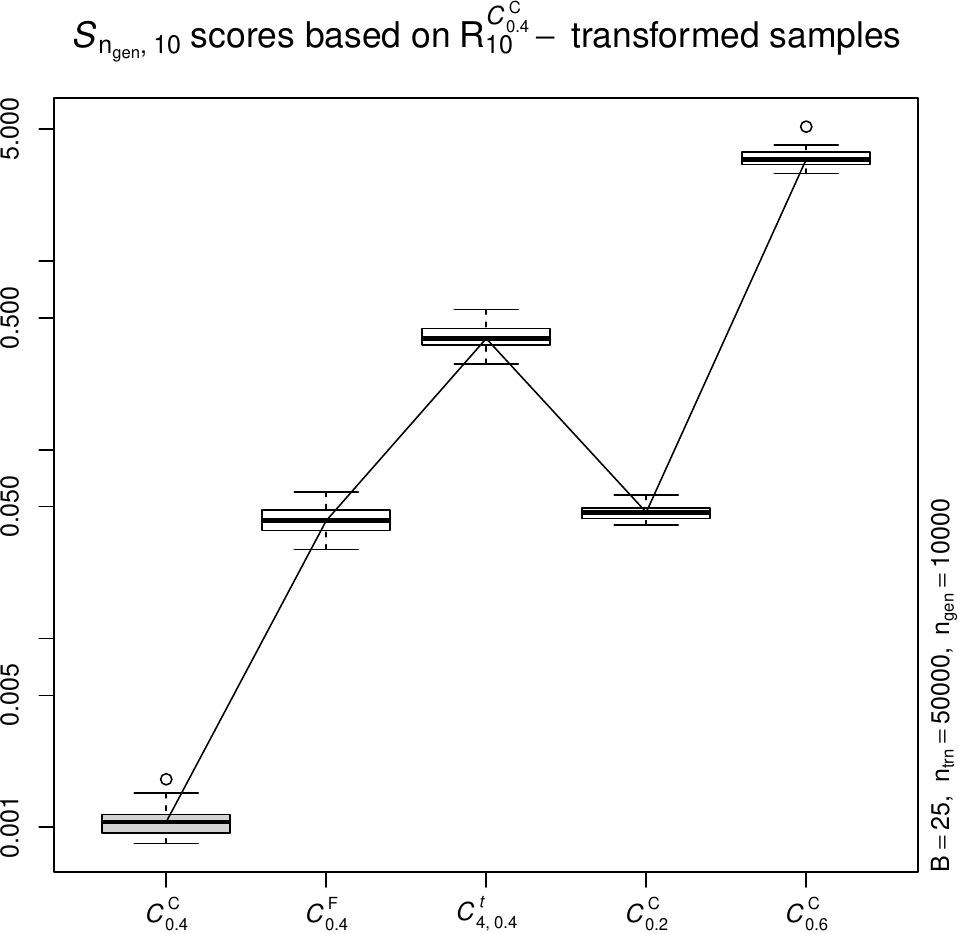}\hfill
  \includegraphics[width=0.30\textwidth]{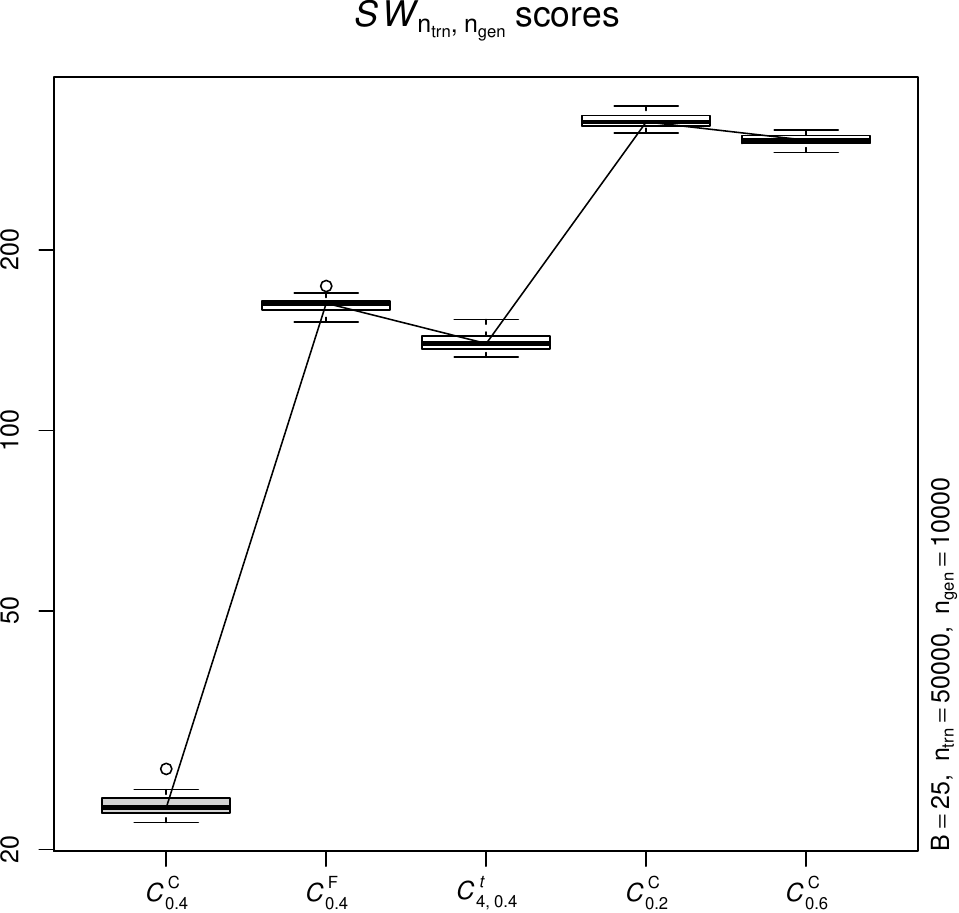}\hspace*{\fill}

  \caption{The left column shows box plots of CvM scores $S_{\ngen,2}$
      based on $B=25$ $D^{C^{\text{C}}_{0.4}}_{d,2}$-transformed samples of size
      $\ngen=10\,000$ from trivariate ($d=3$; top), five-dimensional ($d=5$;
      middle) and ten-dimensional ($d=10$; bottom) copulas $C^{\text{C}}_{0.4}$,
      $C^{\text{F}}_{0.4}$, $C^{t}_{4,0.4}$, $C^{\text{C}}_{0.2}$ and
      $C^{\text{C}}_{0.6}$; see Algorithm~\ref{algo:num:assess} for details.
      The middle column shows the same for $R^{C^{\text{C}}_{0.4}}_d$-transformed samples.
      The right column shows box plots of sliced Wasserstein
      scores $SW_{\ntrn,\ngen}$, computed between the training samples and
      generated samples from the aforementioned copulas.}
  \label{fig:num:clayton}
\end{figure}

In the first setting, we consider $C_{0.4}^{\text{C}}$ as true copula $C$ in
Algorithm~\ref{algo:num:assess}, and $C_{0.4}^{\text{F}}$, $C_{4,0.4}^t$,
$C_{0.2}^{\text{C}}$ and $C_{0.6}^{\text{C}}$ as candidate copulas. The
left-hand side of Figure~\ref{fig:num:clayton} shows box plots of the CvM scores
according to Algorithm~\ref{algo:num:assess} for $d=3$ (top), $d=5$ (middle) and
$d=10$ (bottom).
The middle includes similar plots but obtained from applying the Rosenblatt
transform $R^{C^{\text{C}}_{0.4}}_{d}$ instead of a DecoupleNet
$D^{C^{\text{C}}_{0.4}}_{d,2}$. In particular, recall that the Rosenblatt
transform maps to $d'=d>2$ dimensions so the values of the CvM scores
are not directly comparable. Nevertheless, apart from $C^{t}_{4,0.4}$ (for
$d\in\{3,5\}$) and $C^{\text{C}}_{0.2}$ (for $d=10$), the rankings of the
candidate models are the same. A comparison with the box plot of the true copula
$C_{0.4}^{\text{C}}$ also correctly reveals that based on both
$D^{C^{\text{C}}_{0.4}}_{d,2}$ and $R^{C^{\text{C}}_{0.4}}_{d}$, none of the
candidate copulas is adequate. Finally, the right column of plots in
  Figure~\ref{fig:num:clayton} shows box plots of the sliced Wasserstein score
  $SW_{\ntrn,\ngen}$; see \cite{bonneelrabinpeyrepfister2015}. This score is
  computed between the training samples
  $\hat{U}=\{\hat{\bm{U}}_i\}_{i=1}^{\ntrn}$ and generated samples
  $\tilde{U}=\{\tilde{\bm{U}}_i\}_{i=1}^{\ngen}$ from the aforementioned
  (true or candidate) copulas via
  \begin{align*}
    SW_{\ntrn,\ngen}=\biggl(\frac{1}{\nprj}\sum_{i=1}^{\nprj} \sum_{k=1}^m\biggl|\hat{F}_{\hat{U}\bm{P}_i,m}^{-1}\biggl(\frac{k-1/2}{m}\biggr) - \hat{F}_{\tilde{U}\bm{P}_i,m}^{-1}\biggl(\frac{k-1/2}{m}\biggr)\biggr|^p\biggr)^{1/p},
  \end{align*}
  where $\bm{P}_i=\bm{Z}_i/\lVert\bm{Z}_i\rVert_2$ for
  $\bm{Z}_i\isim\N_d(\bm{0},I_d)$, $i=1,\dots,\nprj=1000$, are random
  projections, $\hat{U}\bm{P}_i$ (respectively $\tilde{U}\bm{P}_i$) denotes the
  univariate dataset with empirical quantile function
  $\hat{F}_{\hat{U}\bm{P}_i,m}^{-1}$ (respectively
  $\hat{F}_{\tilde{U}\bm{P}_i,m}^{-1}$) resulting from projecting $U$
  (respectively $\tilde{U}$) onto $\bm{P}_i$, and $m=\min\{\ntrn,\ngen\}$. We can see that
  the ranking according to this metric (computed without transforming samples to
  multivariate uniformity first) is mostly in line with the previous rankings.

\begin{figure}[htbp]
	\hspace*{\fill}
  \includegraphics[width=0.30\textwidth]{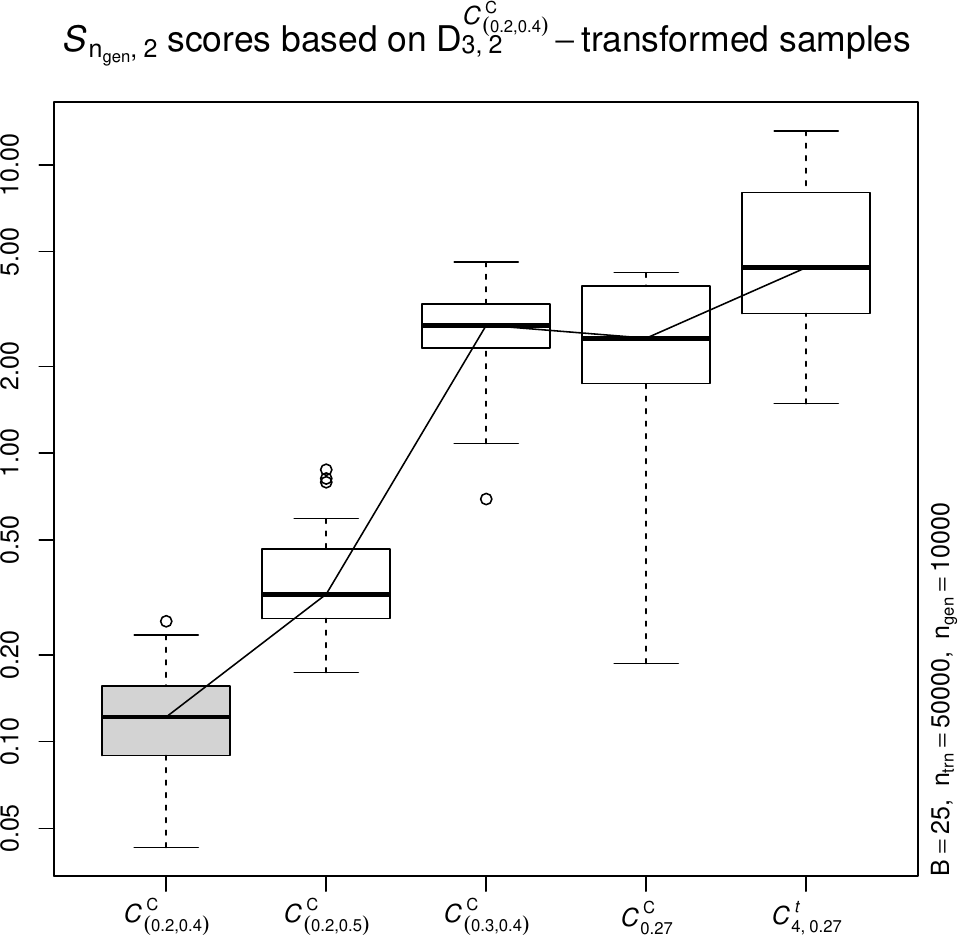}\hfill
  \includegraphics[width=0.30\textwidth]{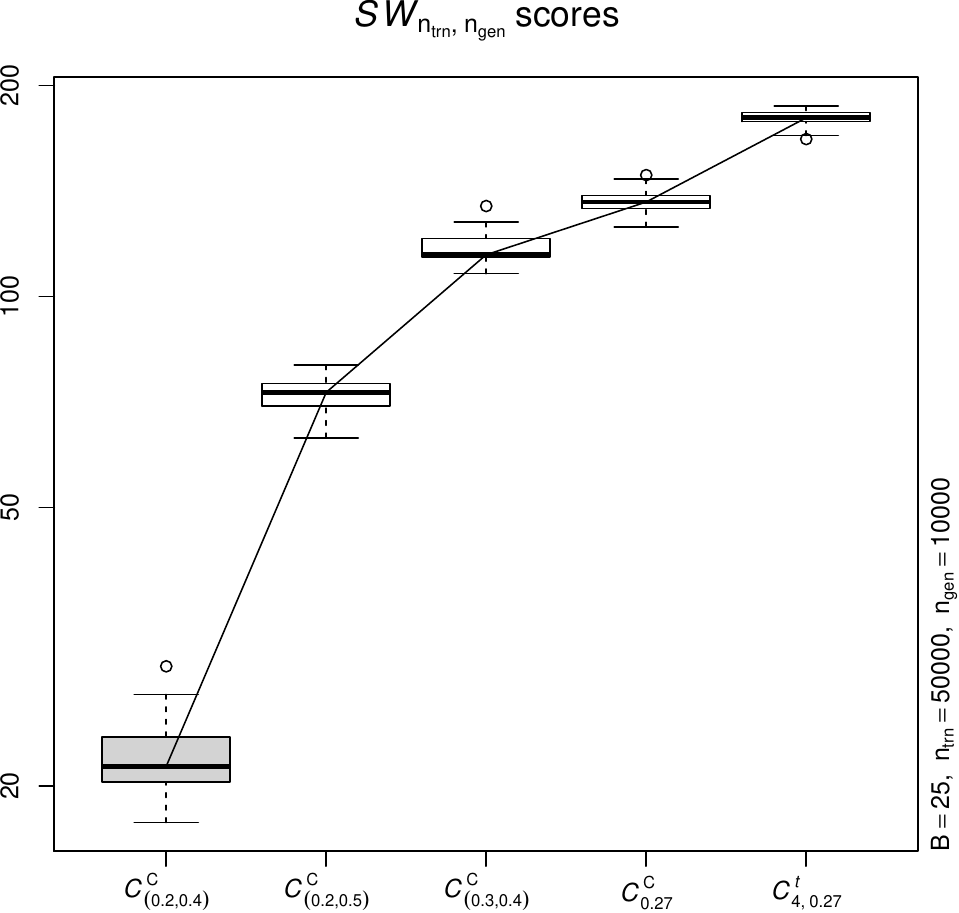} \hspace*{\fill}
  \\[4mm]
  \hspace*{\fill}
  \includegraphics[width=0.30\textwidth]{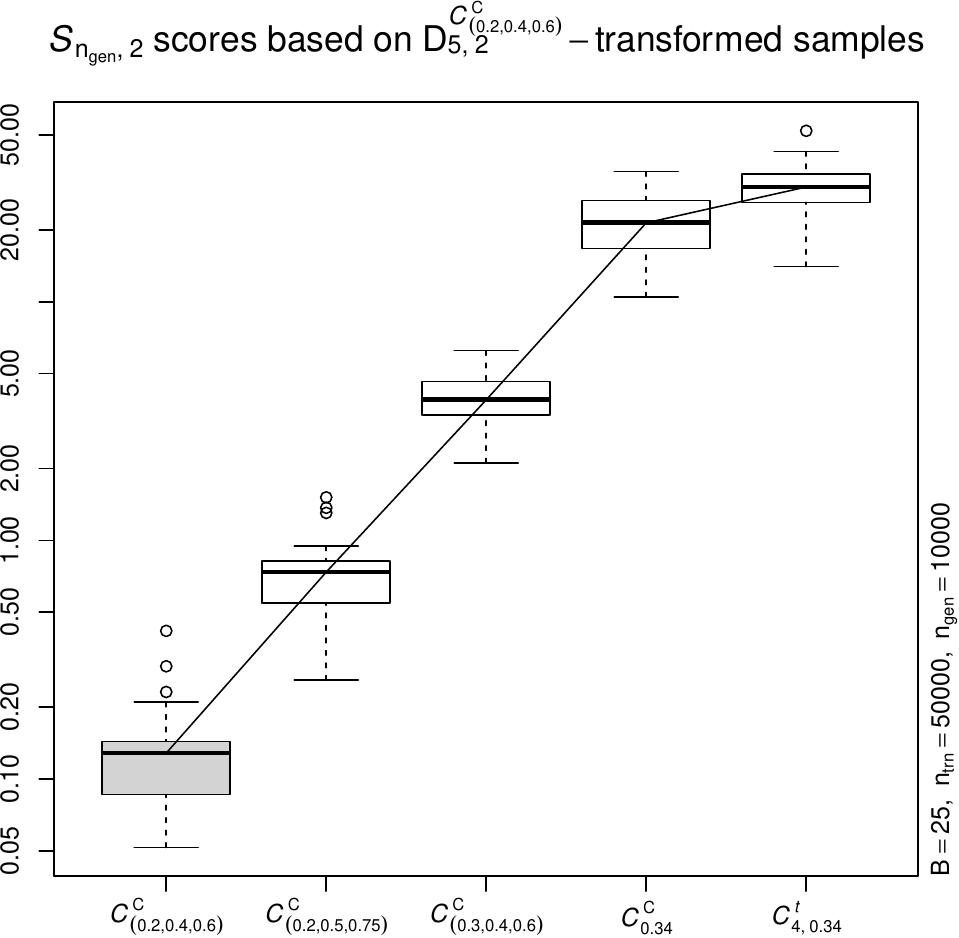}\hfill
  \includegraphics[width=0.30\textwidth]{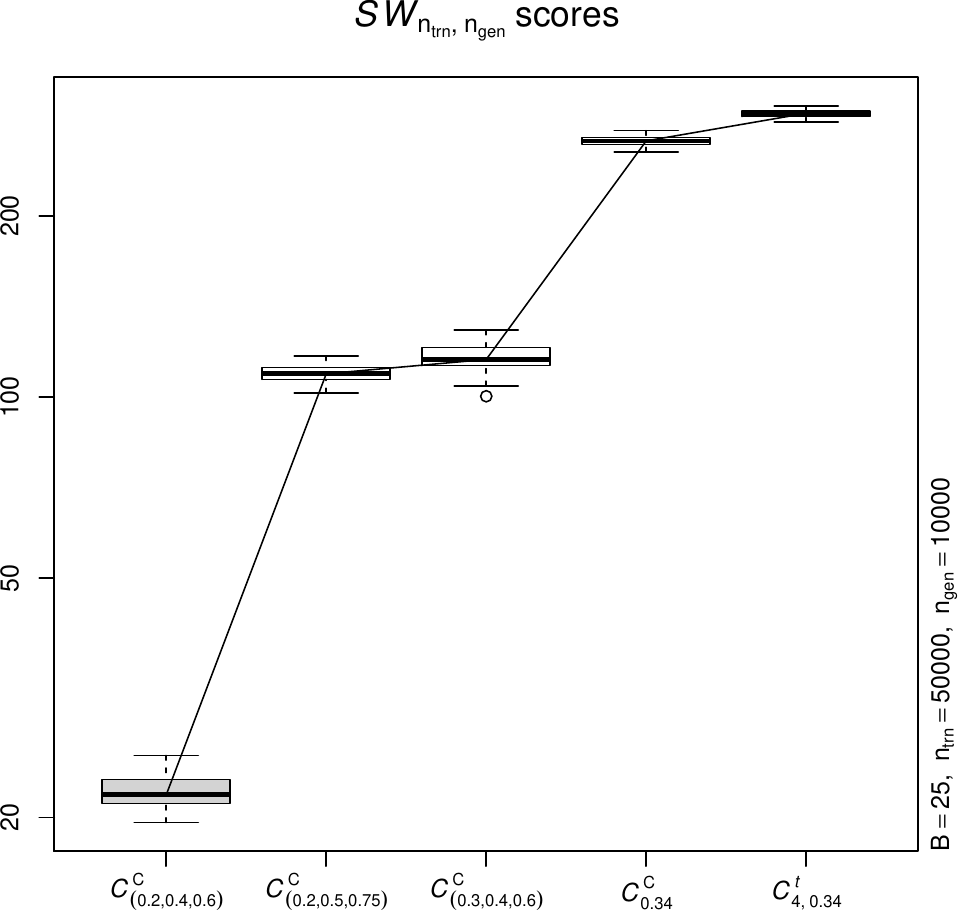}\hspace*{\fill}
  \\[4mm]
  \hspace*{\fill}
  \includegraphics[width=0.30\textwidth]{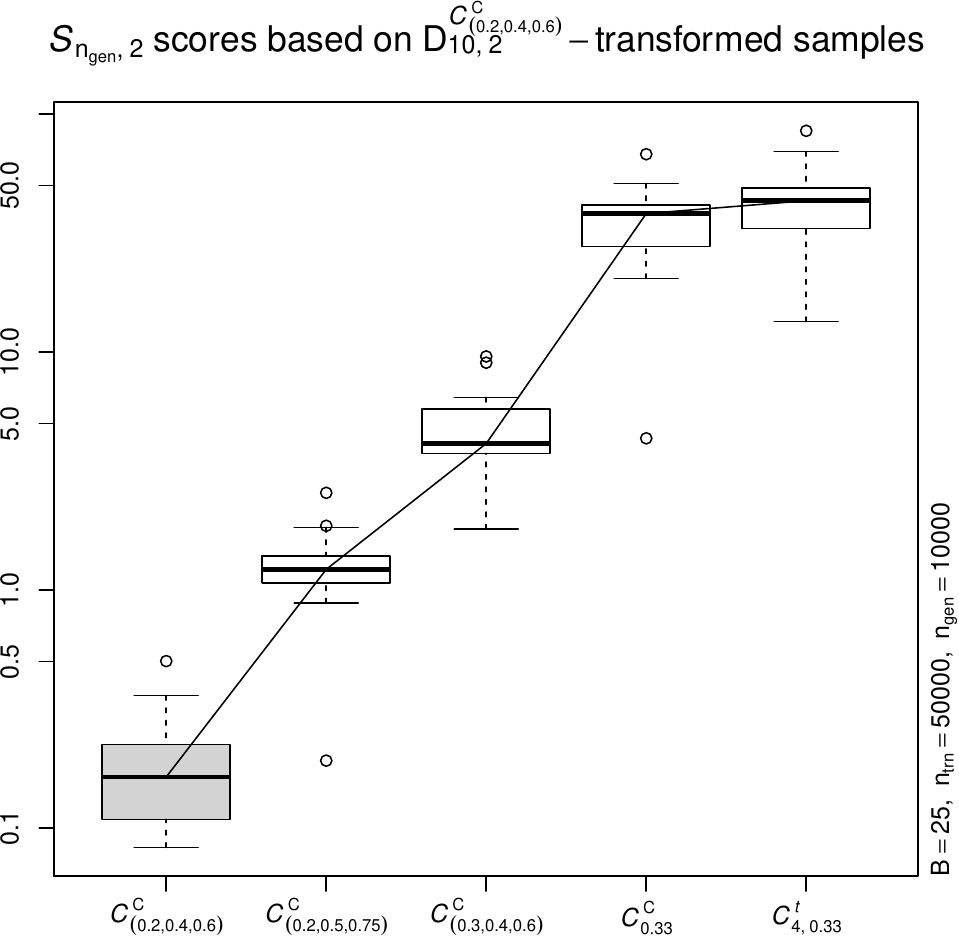}\hfill
  \includegraphics[width=0.30\textwidth]{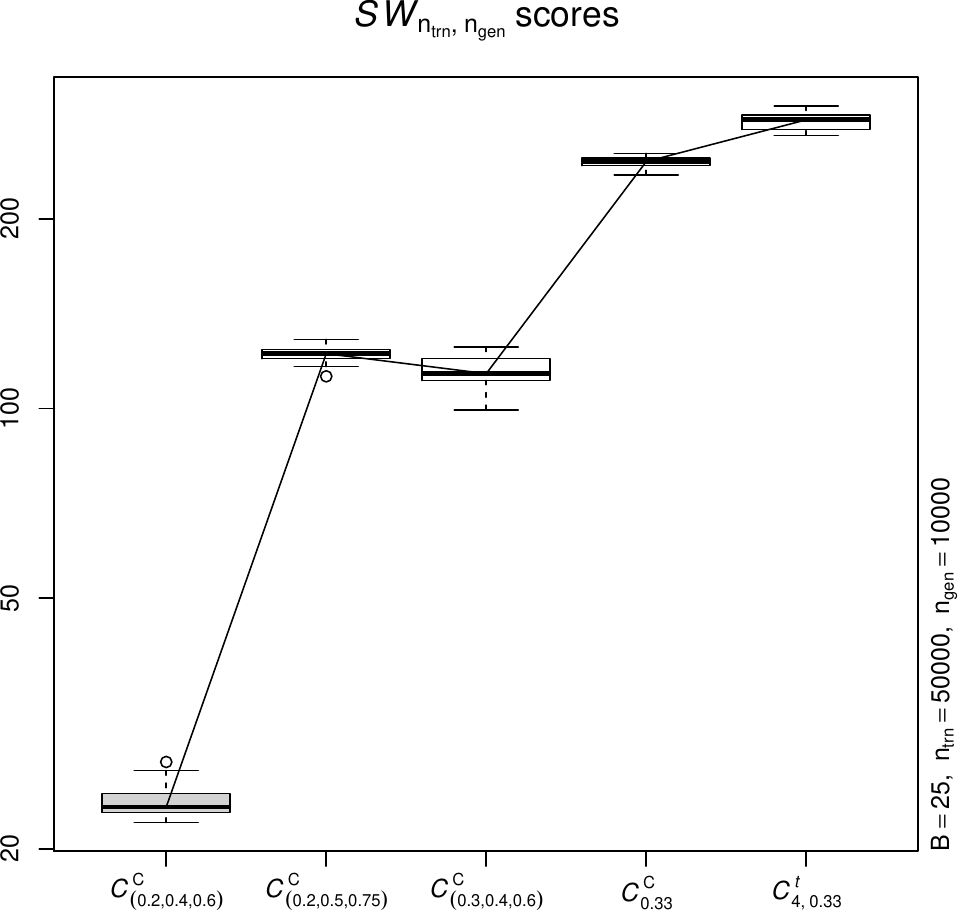}\hspace*{\fill}
  \caption{The left column shows box plots of CvM scores $S_{\ngen,2}$
      based on $\nrep=25$ $D^{C^{\text{C}}_{(0.2,0.4)}}_{3,2}$-transformed
      samples of size $\ngen=10\,000$ from trivariate copulas
      $C^{\text{C}}_{(0.2,0.4)}$,
      $C^{\text{C}}_{(0.2,0.5)}$, $C^{\text{C}}_{(0.3,0.4)}$,
      $C^{\text{C}}_{0.27}$, $C^{t}_{4,0.27}$ (top);
      $D^{C^{\text{C}}_{(0.2,0.4,0.6)}}_{5,2}$-transformed samples of the same
      size from five-dimensional copulas $C^{\text{C}}_{(0.2,0.4,0.6)}$,
      $C^{\text{C}}_{(0.2,0.5,0.75)}$, $C^{\text{C}}_{(0.3,0.4,0.6)}$,
      $C^{\text{C}}_{0.34}$, $C^{t}_{4,0.34}$ (middle); and
      $D^{C^{\text{C}}_{(0.2,0.4,0.6)}}_{10,2}$-transformed samples of the same
      size from ten-dimensional copulas $C^{\text{C}}_{(0.2,0.4,0.6)}$,
      $C^{\text{C}}_{(0.2,0.5,0.75)}$, $C^{\text{C}}_{(0.3,0.4,0.6)}$,
      $C^{\text{C}}_{0.33}$, $C^{t}_{4,0.33}$ (bottom); see
      Algorithm~\ref{algo:num:assess} for details. The right column shows box
      plots of sliced Wasserstein scores $SW_{\ntrn,\ngen}$, computed between
      the training samples and generated samples from the aforementioned
      copulas.}
  \label{fig:num:NC}
\end{figure}

In the second setting, we consider nested Clayton copulas as true copula $C$ in
Algorithm~\ref{algo:num:assess}. To this end let $C_k$, $k=0,1,2$, be a Clayton
copula with parameter chosen such that Kendall's tau equals $\tau_k$.  For $d=3$
we choose a $(2,1)$-nested Clayton copula $C_0(C_1(u_1,u_2),u_3)$ with
$(\tau_0,\tau_1)=(0.2,0.4)$, denoted by $C^{\text{C}}_{(0.2,0.4)}$.  Besides
this copula as true copula, we consider the trivariate candidate models
$C^{\text{C}}_{(0.2,0.5)}$, $C^{\text{C}}_{(0.3,0.4)}$, $C^{\text{C}}_{0.27}$
and $C^{t}_{4,0.27}$. The top left plot of Figure~\ref{fig:num:NC}
shows the box plots of the CvM scores according to
Algorithm~\ref{algo:num:assess}.
For $d=5$ we choose a $(2,3)$-nested Clayton copula
$C_0(C_1(u_1,u_2),C_2(u_3,u_4,u_5))$ with
$(\tau_0,\tau_1,\tau_2)=(0.2,0.4,0.6)$, denoted by
$C^{\text{C}}_{(0.2,0.4,0.6)}$.  Besides this copula as true copula, we consider
the five-dimensional candidate models $C^{\text{C}}_{(0.2,0.5,0.75)}$,
$C^{\text{C}}_{(0.3,0.4,0.6)}$, $C^{\text{C}}_{0.34}$ and $C^{t}_{4,0.34}$.  The
resulting box plots of the CvM scores according to
Algorithm~\ref{algo:num:assess} are shown in the middle left
of Figure~\ref{fig:num:NC}. And for $d=10$ we choose a $(5,5)$-nested Clayton
copula $C_0(C_1(u_1,\dots,u_5),C_2(u_6,\dots,u_{10}))$ with
$(\tau_0,\tau_1,\tau_2)=(0.2,0.4,0.6)$, also denoted by
$C^{\text{C}}_{(0.2,0.4,0.6)}$, and ten-dimensional candidate models
$C^{\text{C}}_{(0.2,0.5,0.75)}$, $C^{\text{C}}_{(0.3,0.4,0.6)}$,
$C^{\text{C}}_{0.33}$ and $C^{t}_{4,0.33}$. The resulting box plots of the CvM
scores are shown in the bottom left of
Figure~\ref{fig:num:NC}. Among the candidate models, the first two are also of
hierarchical nature, just with different parameters, whereas the other candidate
models are exchangeable with parameters chosen to match the average pairwise
dependence. That is, for $d=3$, $d=5$ and $d=10$ copulas, we set
$\tau=\frac{1}{{3 \choose 2}}(2\tau_0+\tau_1)$,
$\tau=\frac{1}{{5\choose 2}}(6\tau_0+\tau_1+3\tau_2)$ and
$\tau=\frac{1}{{10\choose 2}}(25\tau_0+10\tau_1+10\tau_2)$, respectively.
As results, we clearly see from Figure~\ref{fig:num:NC} that none of the
exchangeable or nested candidate models are adequate, which aligns
with intuition. Moreover, from the rankings of the two nested models, we see
that the deviation in $\tau_0$ is more important than deviations in both
$\tau_1$ or $\tau_2$. This is due to the fact that there exist more pairwise
marginal copula with Kendall's tau $\tau_0$ than those with Kendall's tau
$\tau_1$ and Kendall's tau $\tau_2$ combined. The right column of Figure~\ref{fig:num:NC}
shows the corresponding box plots of the sliced Wasserstein score; again, the
ranking according to this metric is mostly in line with those obtained
via the DecoupleNet transformed samples.

\begin{figure}[htbp]
  \hspace*{\fill}
  \includegraphics[width=0.30\textwidth]{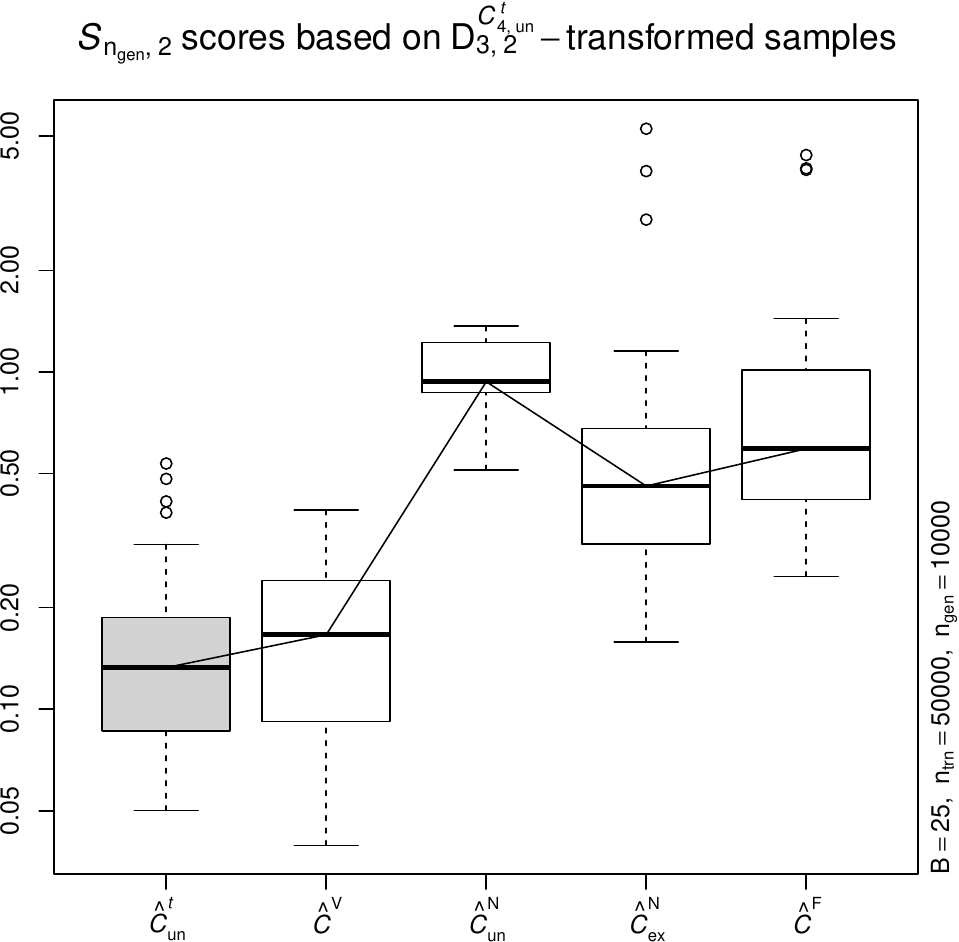}\hfill
  \includegraphics[width=0.30\textwidth]{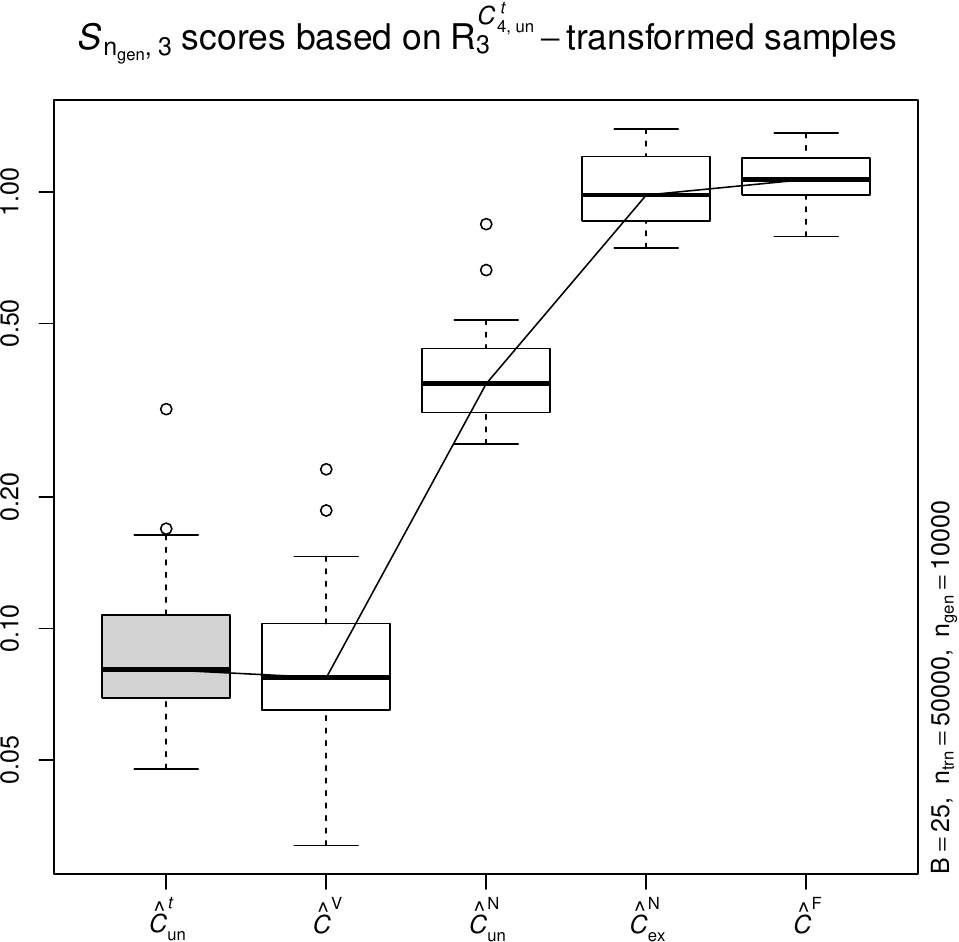}\hfill
  \includegraphics[width=0.30\textwidth]{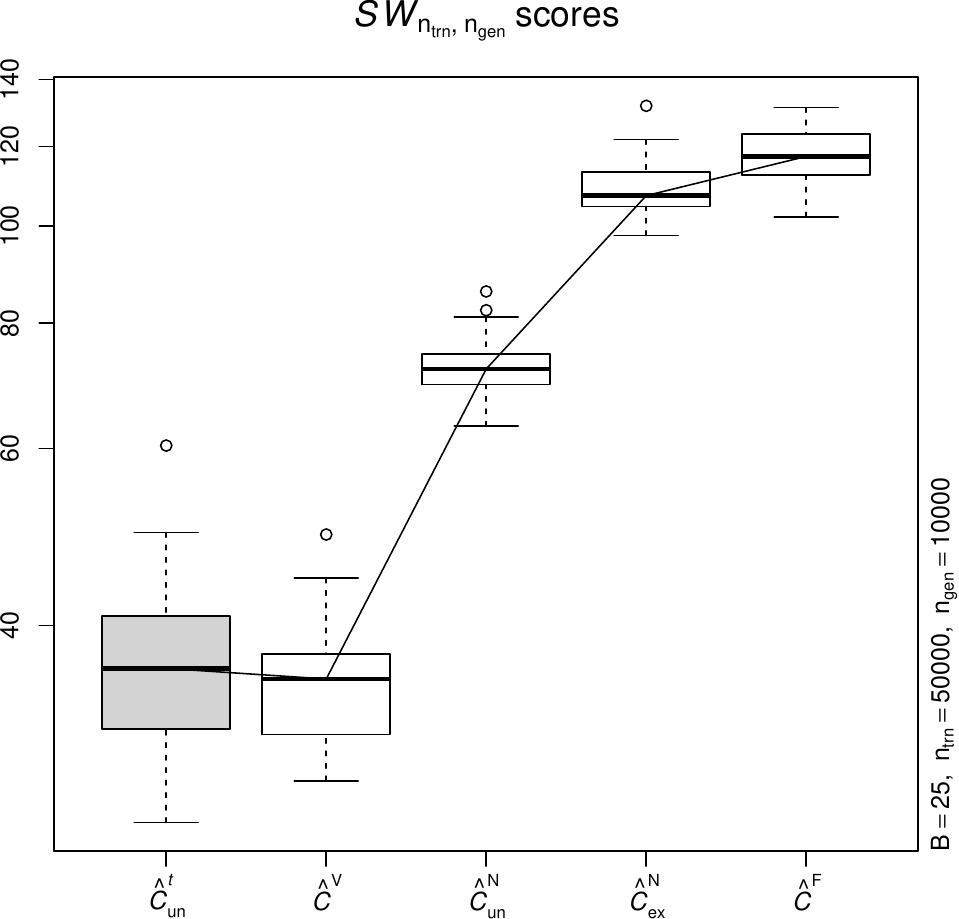}\hspace*{\fill}
  \\[4mm]
  \hspace*{\fill}
  \includegraphics[width=0.30\textwidth]{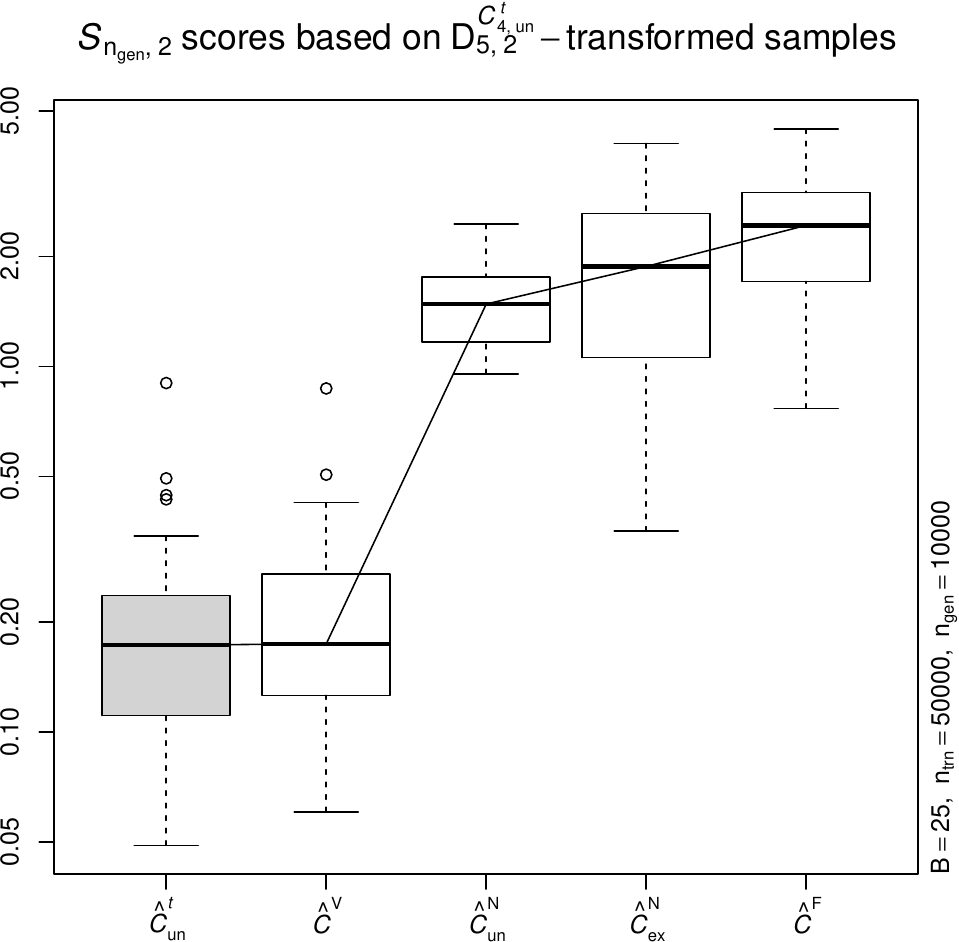}\hfill
  \includegraphics[width=0.30\textwidth]{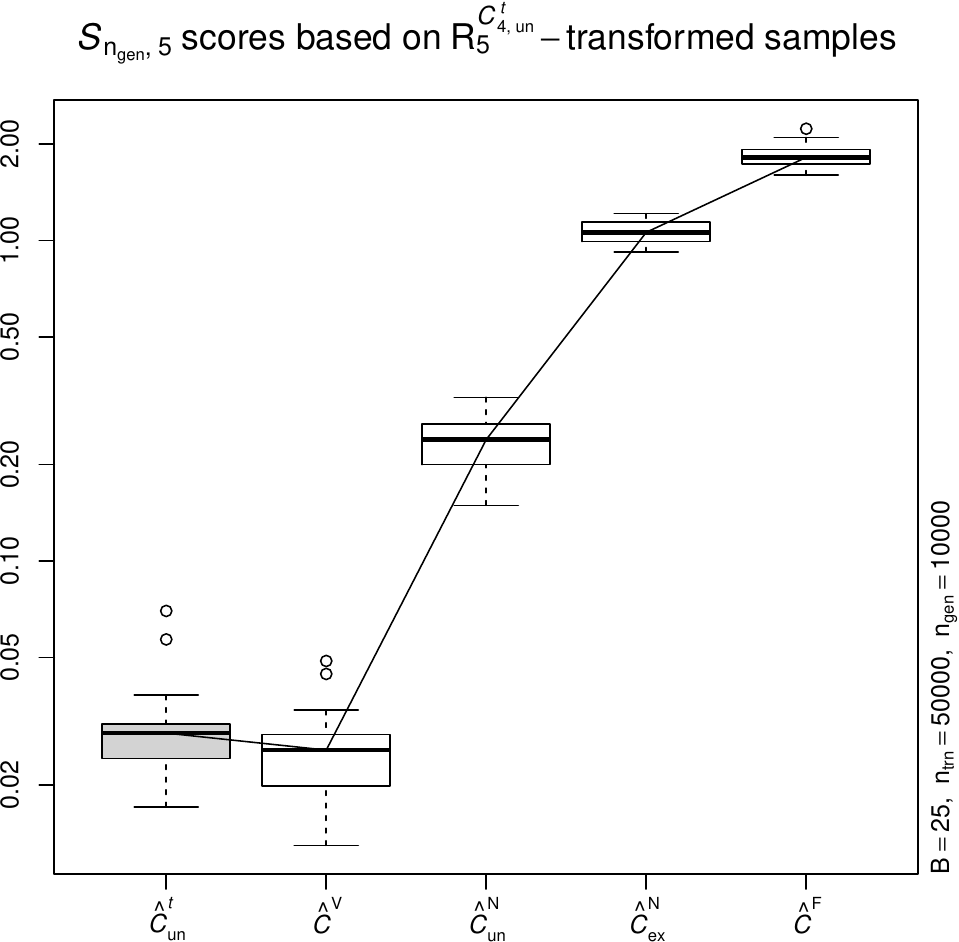}\hfill
  \includegraphics[width=0.30\textwidth]{fig_boxplot_sliced_was_cop_dim_3_ngen_10000_t_un_t_un_vine_norm_un_norm_ex_frank.pdf}\hspace*{\fill}
  \\[4mm]
  \hspace*{\fill}
  \includegraphics[width=0.30\textwidth]{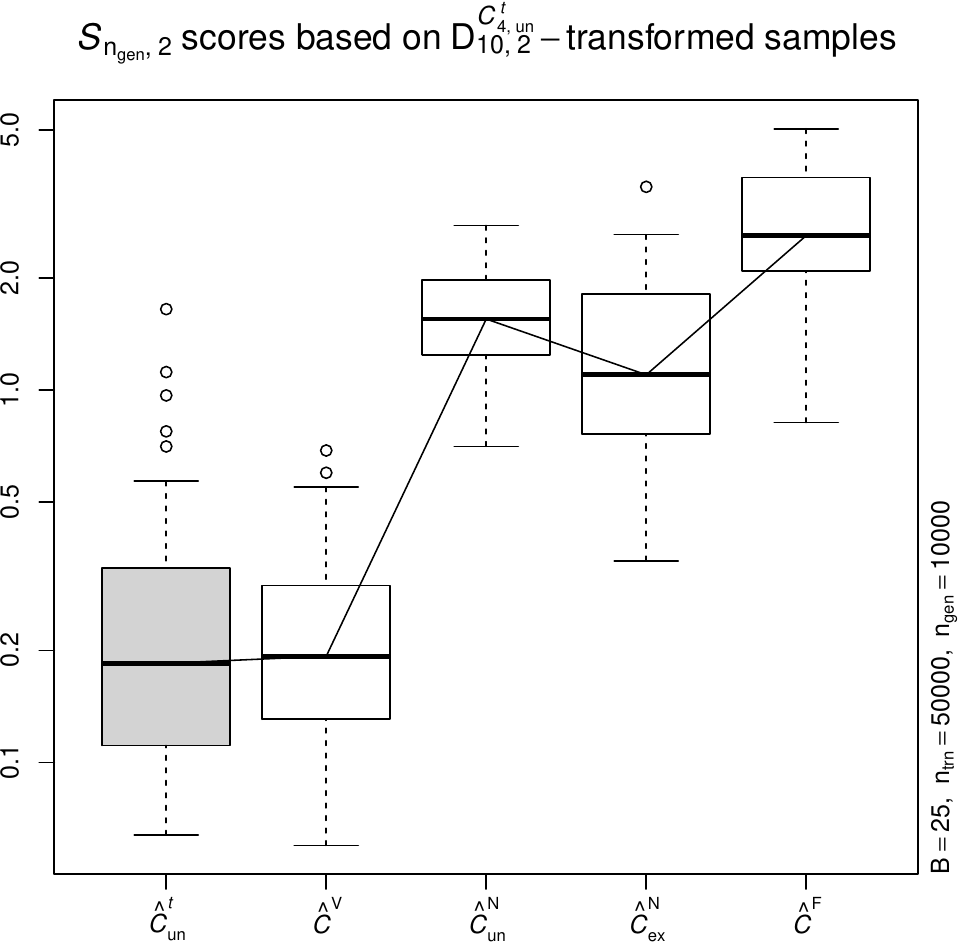}\hfill
  \includegraphics[width=0.30\textwidth]{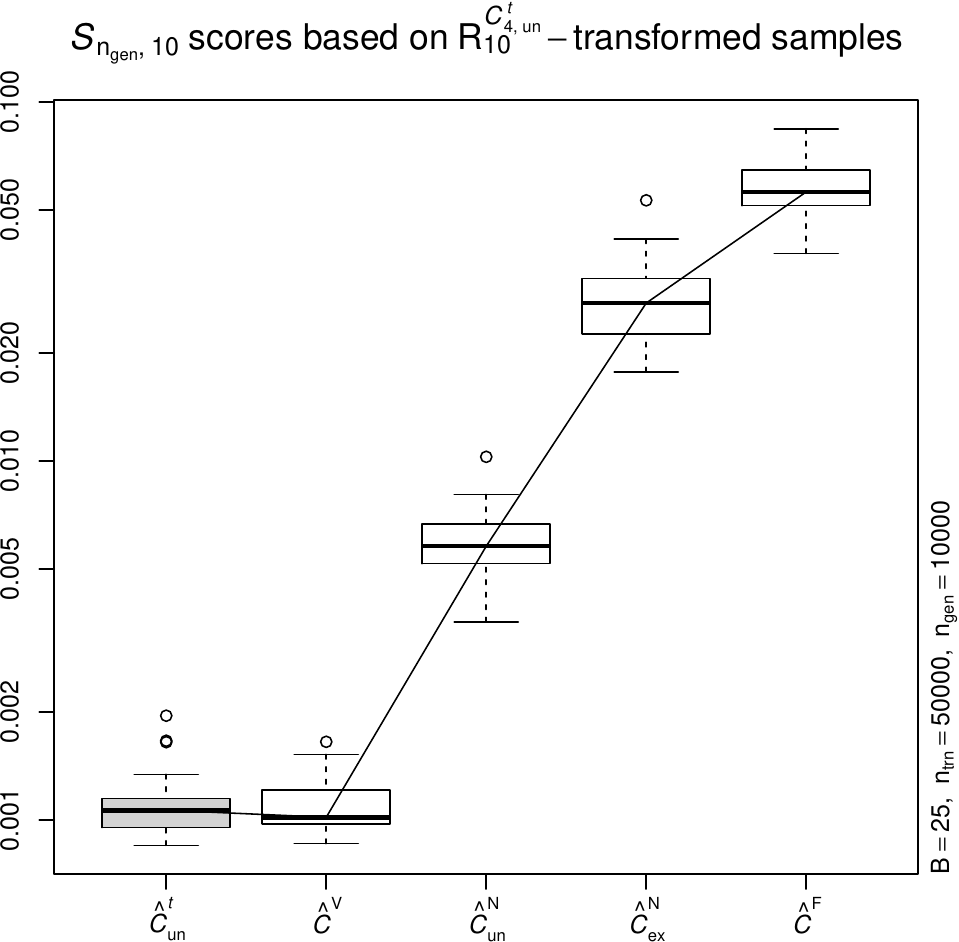}\hfill
  \includegraphics[width=0.30\textwidth]{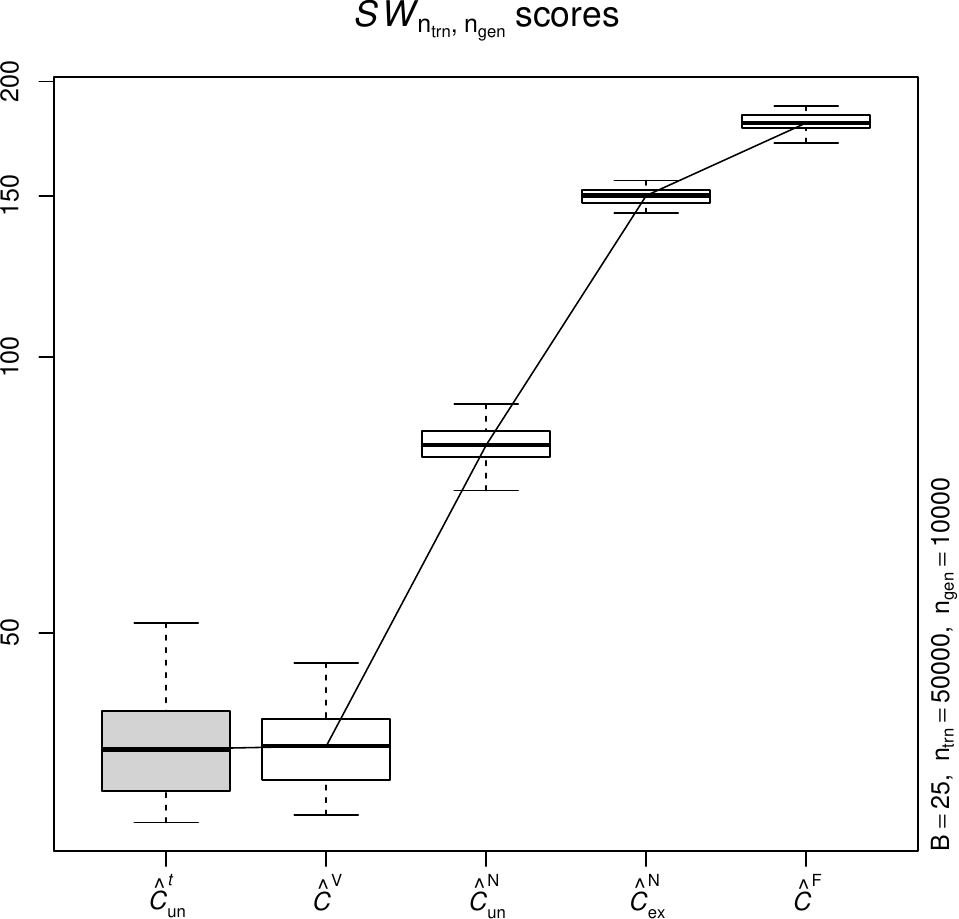}\hspace*{\fill}
  \caption{The left column shows box plots of CvM scores $S_{\ngen,2}$
      based on $B=25$ $D^{C^{t}_{4,\text{un}}}_{d,2}$-transformed samples of
      size $\ngen=10\,000$ from trivariate ($d=3$; top), five-dimensional
      ($d=5$; middle) and ten-dimensional ($d=10$; bottom) fitted copulas
      $\hat{C}^t_{\text{un}}$, $\hat{C}^{\text{V}}$,
      $\hat{C}^{\text{N}}_{\text{un}}$, $\hat{C}^{\text{N}}_{\text{ex}}$ and
      $\hat{C}^{\text{F}}$; see Algorithm~\ref{algo:num:assess} for details.
      The middle column shows the same for $R^{C^{t}_{4,\text{un}}}_d$-transformed samples.
      The right column shows box plots of sliced
      Wasserstein scores $SW_{\ntrn,\ngen}$, computed between the training
      samples and generated samples from the aforementioned copulas.
    }
  \label{fig:num:tun}
\end{figure}

In the third and final setting, we consider an unstructured $t$ copula with
$\nu=4$ degrees of freedom and random correlation
matrix %
as true copula, in $d=3$, $d=5$ and $d=10$ dimensions. As benchmark we include a
fitted (unstructured) $t$ copula $\hat{C}^t_{\text{un}}$.  As candidate copulas
we include a fitted vine copula $\hat{C}^{\text{V}}$ (fitted with
\code{RvineStructureSelect()} from the \R\ package \code{VineCopula} with tree
structure selected using Dissman's algorithm in \cite{dissmann2013} and AIC to
select the pair-copula families), a fitted unstructured normal copula
$\hat{C}^{\text{N}}_{\text{un}}$, a fitted exchangeable normal copula
$\hat{C}^{\text{N}}_{\text{ex}}$ and a fitted Frank copula $\hat{C}^{\text{F}}$.
The left-hand column of Figure~\ref{fig:num:tun} shows the box plots of the CvM
scores according to Algorithm~\ref{algo:num:assess} for $d=3$ (top),
$d=5$ (middle) and $d=10$ (bottom).
The center column includes similar plots but obtained from applying the
Rosenblatt transform $R^{C^t_{4,\text{un}}}_{d}$ instead of a DecoupleNet
$D^{C^{t}_{4,\text{un}}}_{d,2}$. And the right column contains the box plots
based on the sliced Wasserstein score. We observe here that the rankings
produced from these different scores are fairly comparable as
well. In particular, the sliced Wasserstein score (again, not involving a
  transformation to uniformity first) does not do a better job at distinguishing
  $\hat{C}^{\text{V}}$ from $\hat{C}^t_{\text{un}}$; note that the slightly
  larger variance for the $D^{C^{t}_{4,\text{un}}}_{d,2}$-transformed samples
  does not come as a surprise due to the retraining of the DecoupleNet $B$
  times, a price one has to pay for the gain in flexibility.

\section{Model assessment and selection based on real world data}\label{sec:applications}
In this section we apply our DecoupleNet approach to two real world datasets.
The first contains pseudo-observations of the water-level heights of two rivers;
the second consists of two sets of foreign exchange rates.

\subsection{Danube data}
We consider the dataset \code{danube} from the \R\ package \code{lcopula},
referred to as the ``Danube data'' in what follows.  It consists of $659$
pseudo-observations of prewhitened monthly average water-level heights of the
Danube river at Nagyramos (Hungary) and those of the Inn river at Sch\"arding
(Austria); for more information about the Danube data, including the type of
prewhitening applied, see the help page of \code{danube} in \code{lcopula}.
With the Inn being a tributary to the Danube, the two water-level heights are
naturally dependent, and \cite[Section~5.2.5]{hofertkojadinovicmaechleryan2018}
showed that there is no strong evidence against the hypothesis that this
dependence is Gumbel.

\begin{figure}[htbp]
  \includegraphics[width=0.16\textwidth]{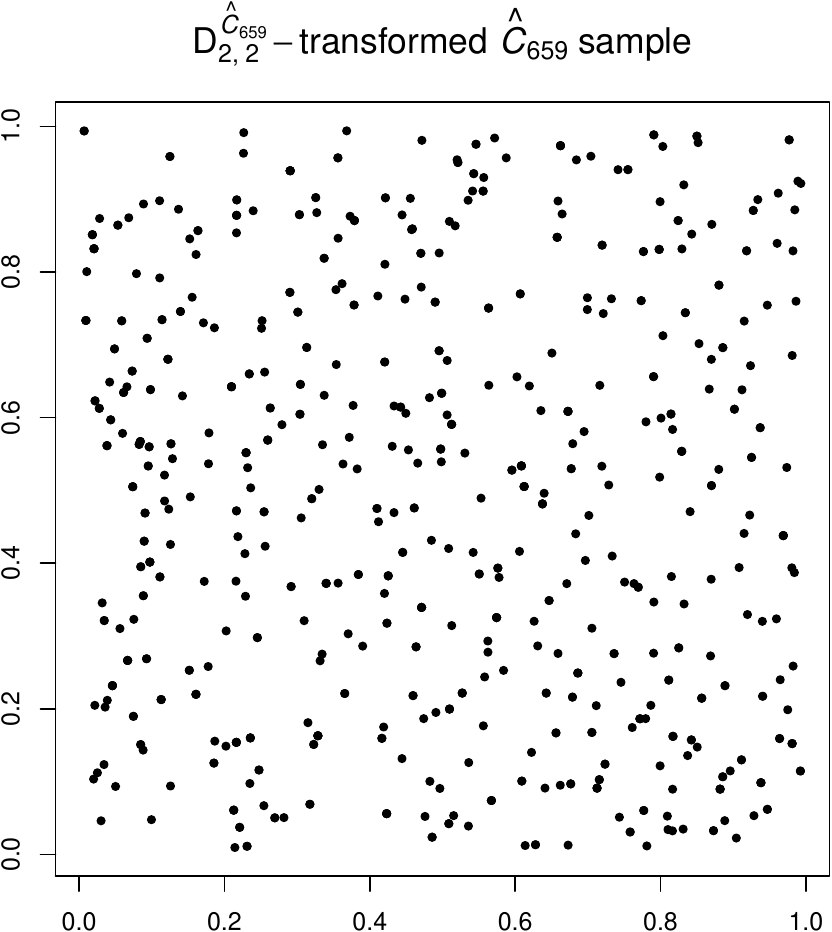}\hfill
  \includegraphics[width=0.16\textwidth]{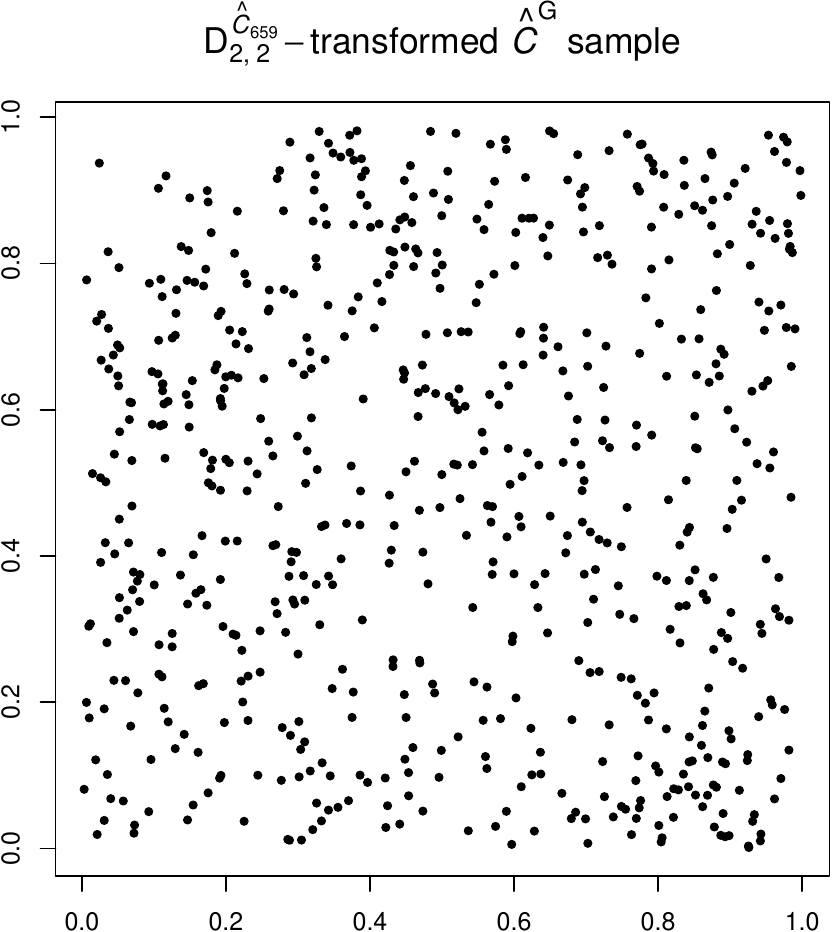}\hfill
  \includegraphics[width=0.16\textwidth]{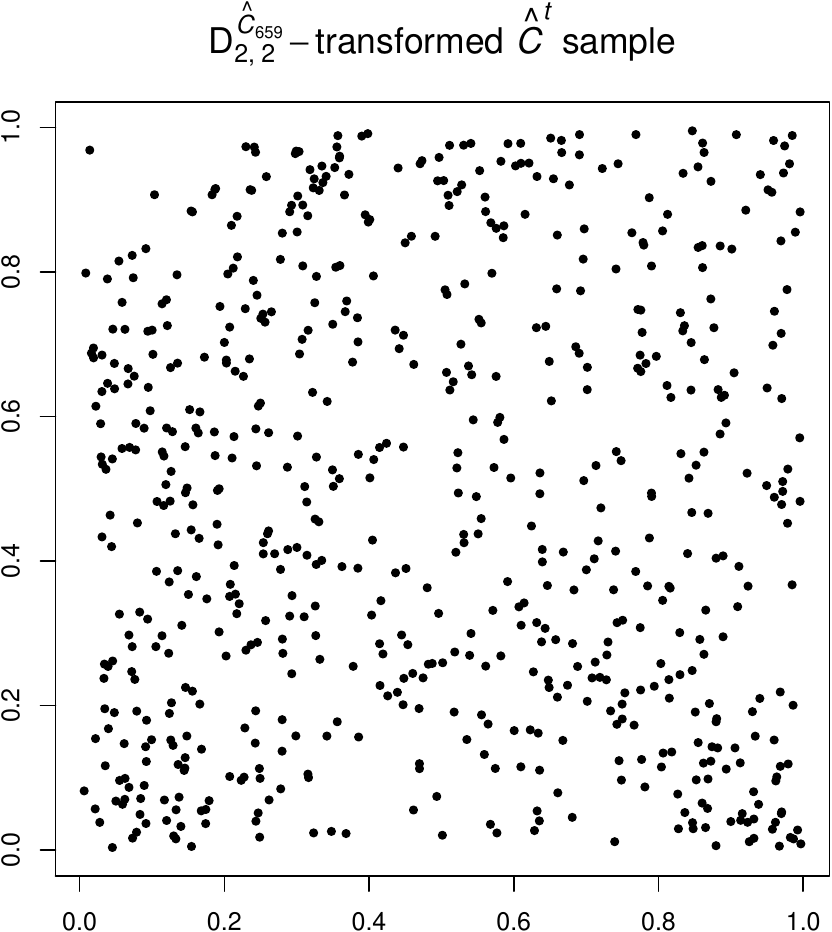}\hfill
  \includegraphics[width=0.16\textwidth]{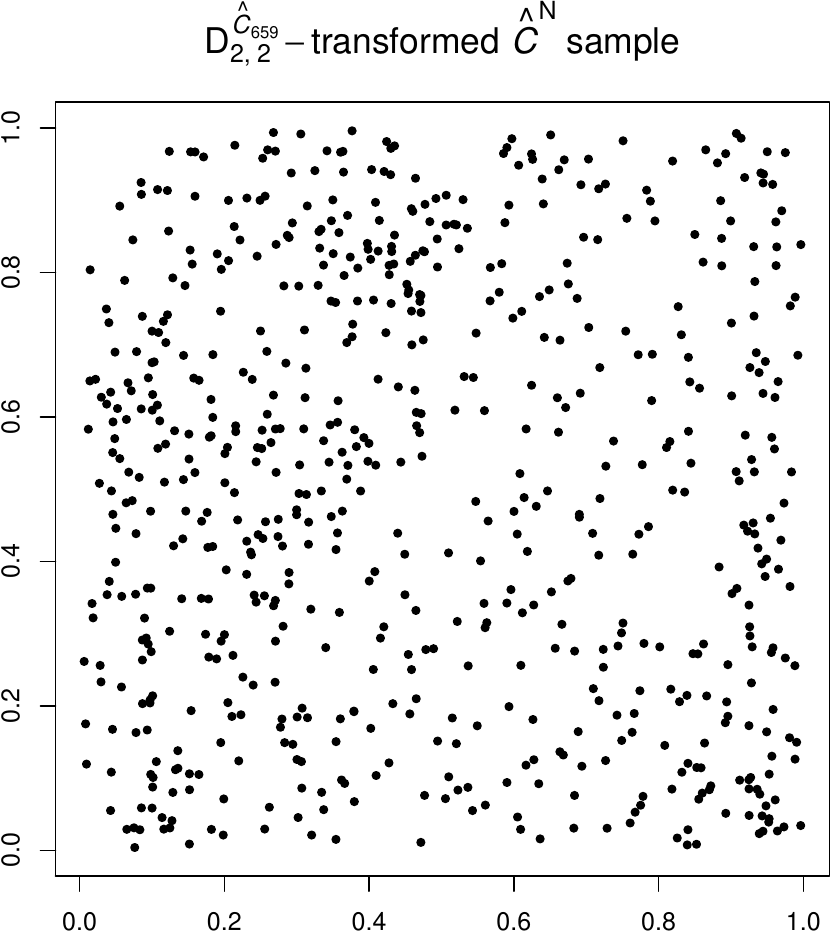}\hfill
  \includegraphics[width=0.16\textwidth]{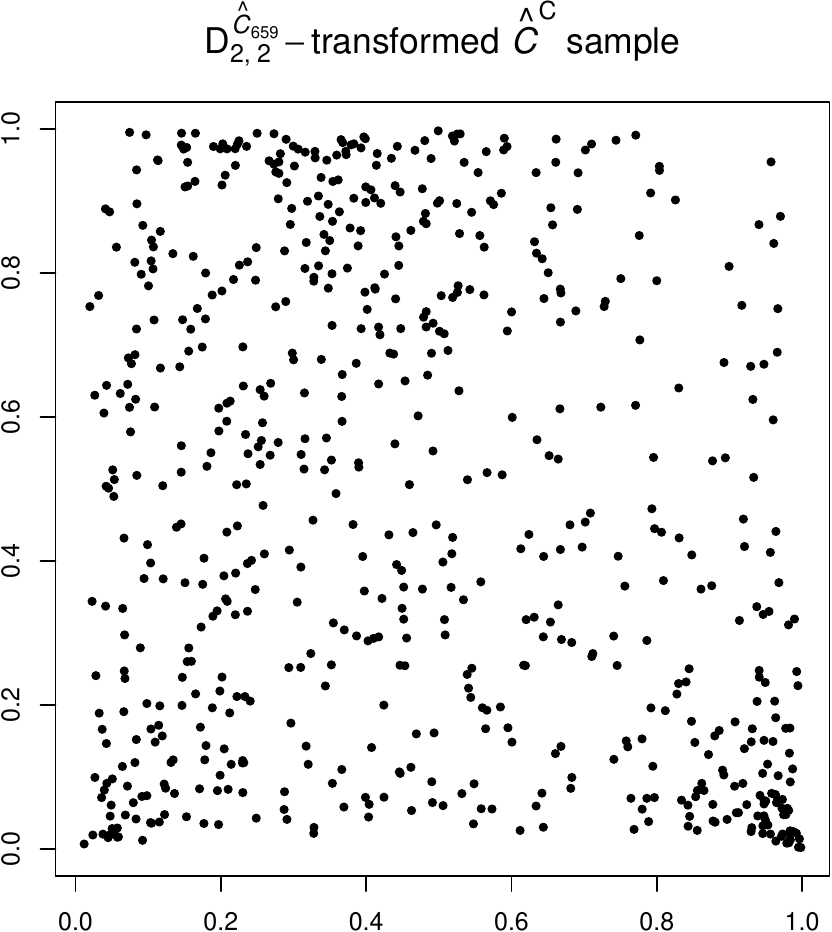}\hfill
  \includegraphics[width=0.16\textwidth]{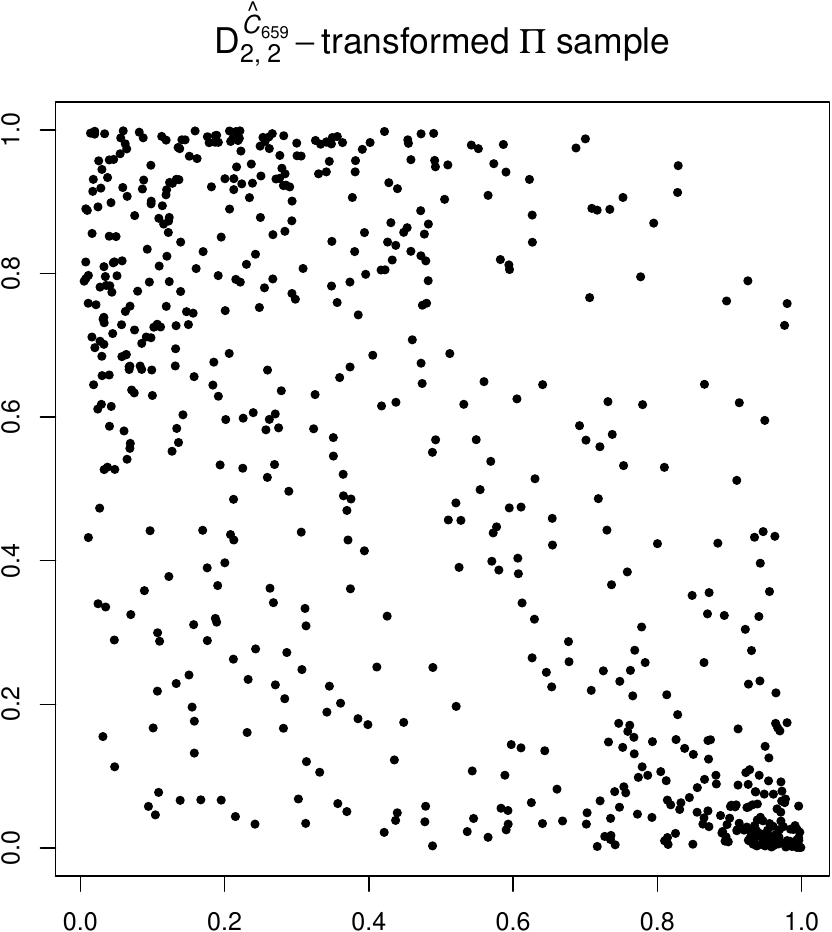}
  \\[2mm]
  \includegraphics[width=0.16\textwidth]{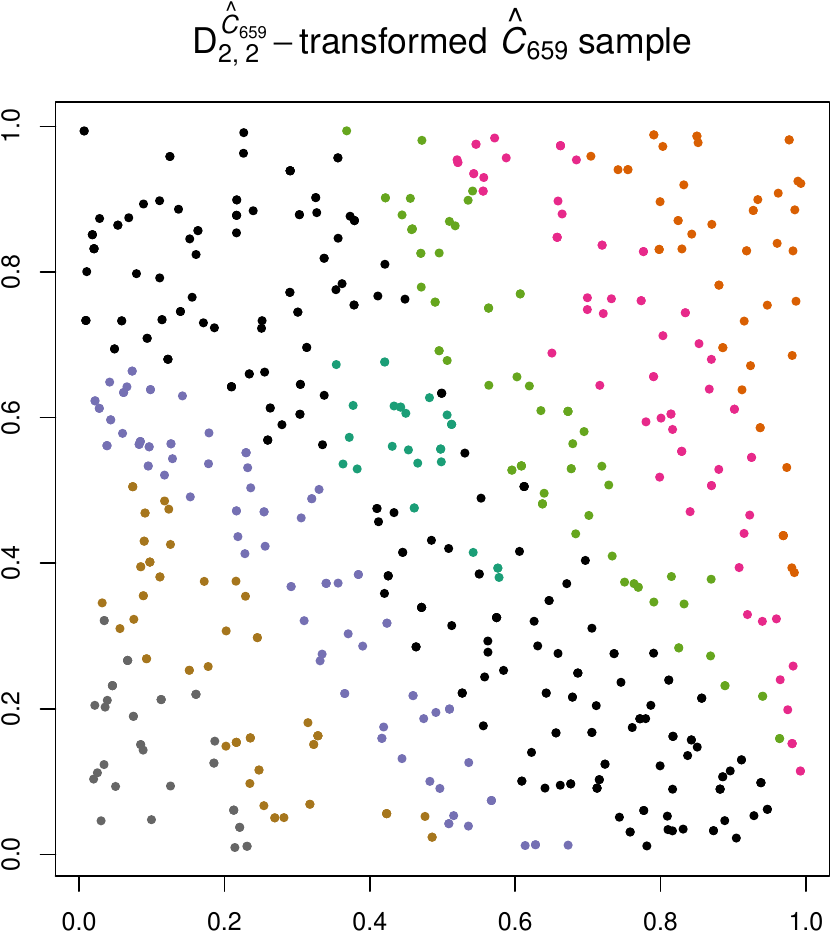}\hfill
  \includegraphics[width=0.16\textwidth]{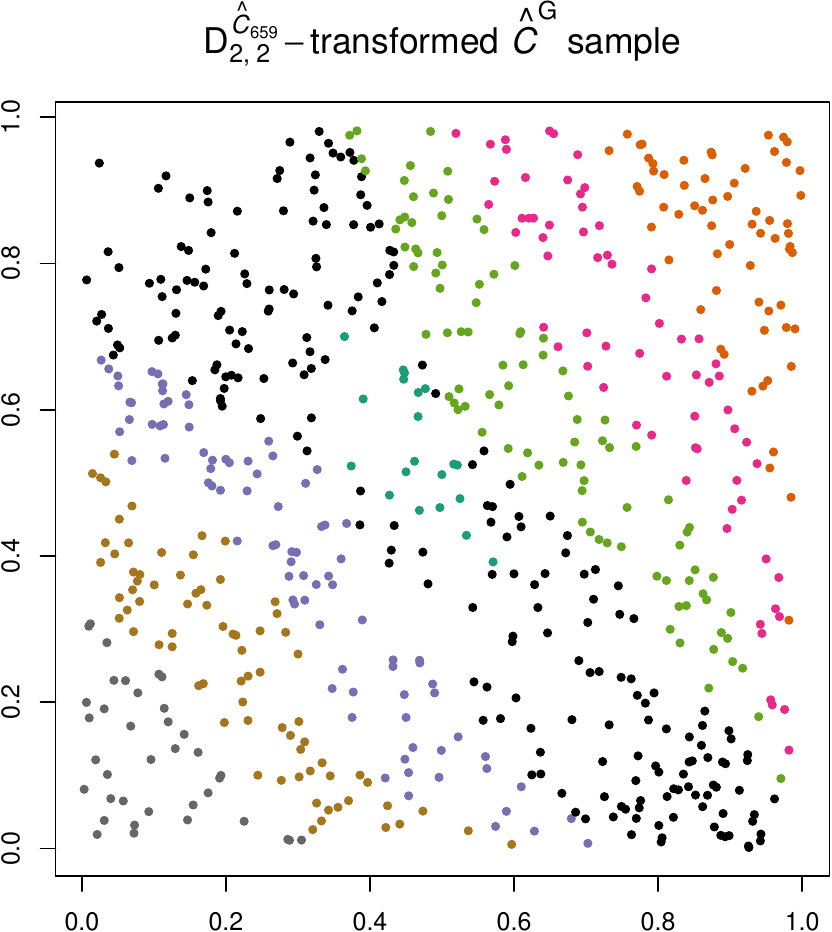}\hfill
  \includegraphics[width=0.16\textwidth]{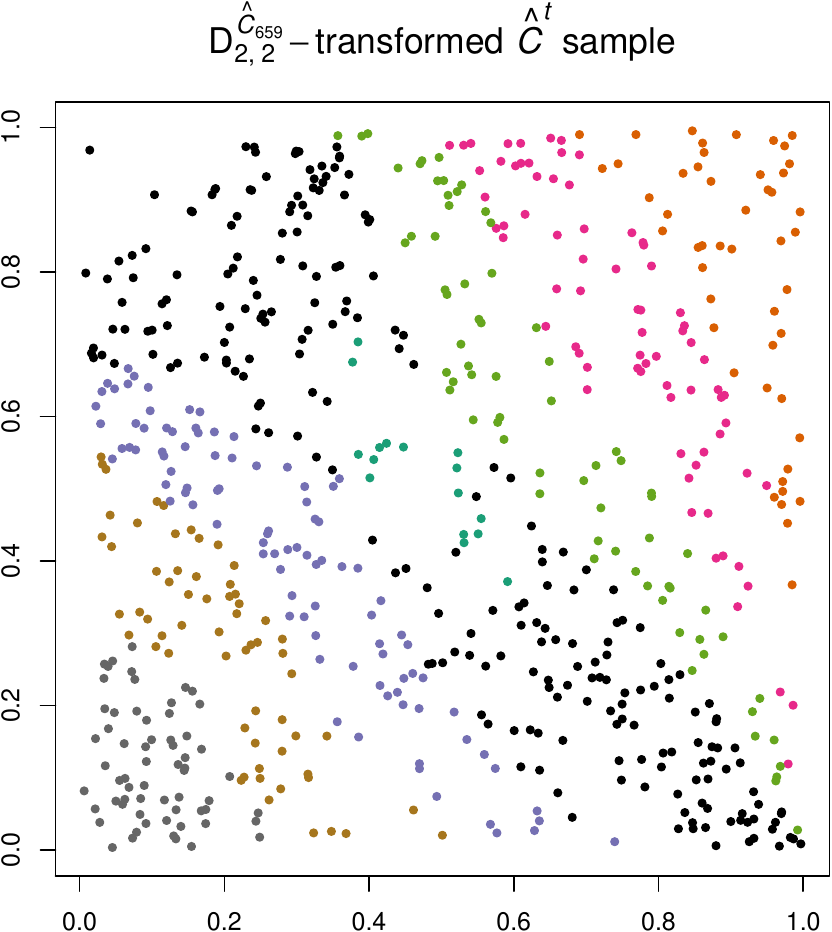}\hfill
  \includegraphics[width=0.16\textwidth]{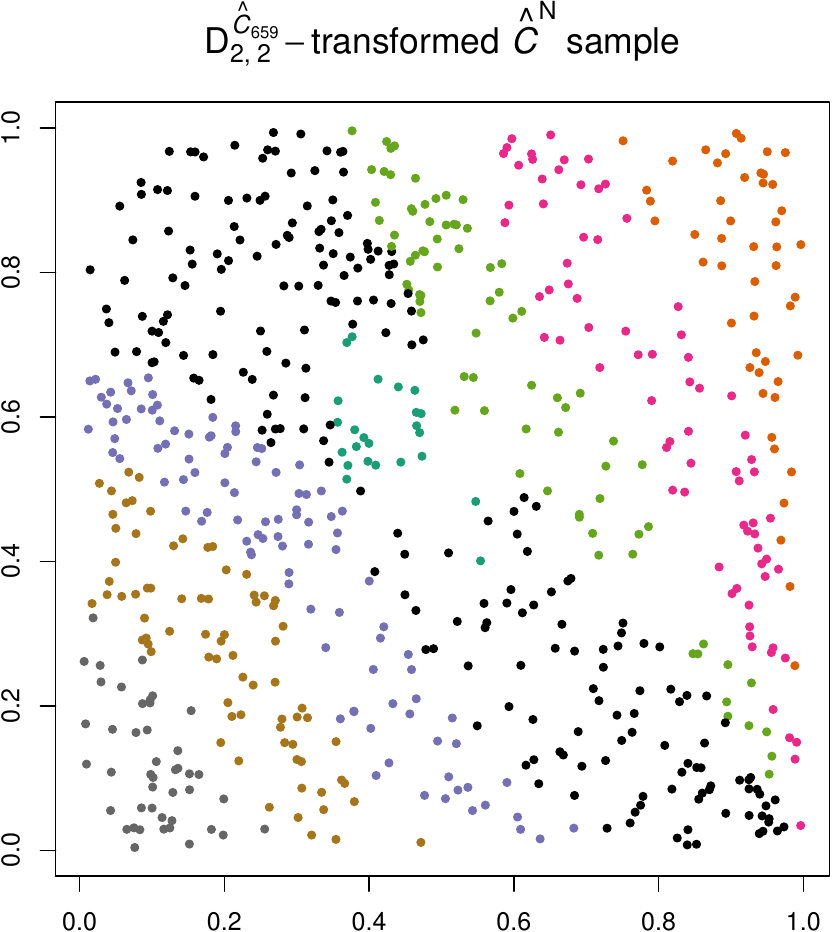}\hfill
  \includegraphics[width=0.16\textwidth]{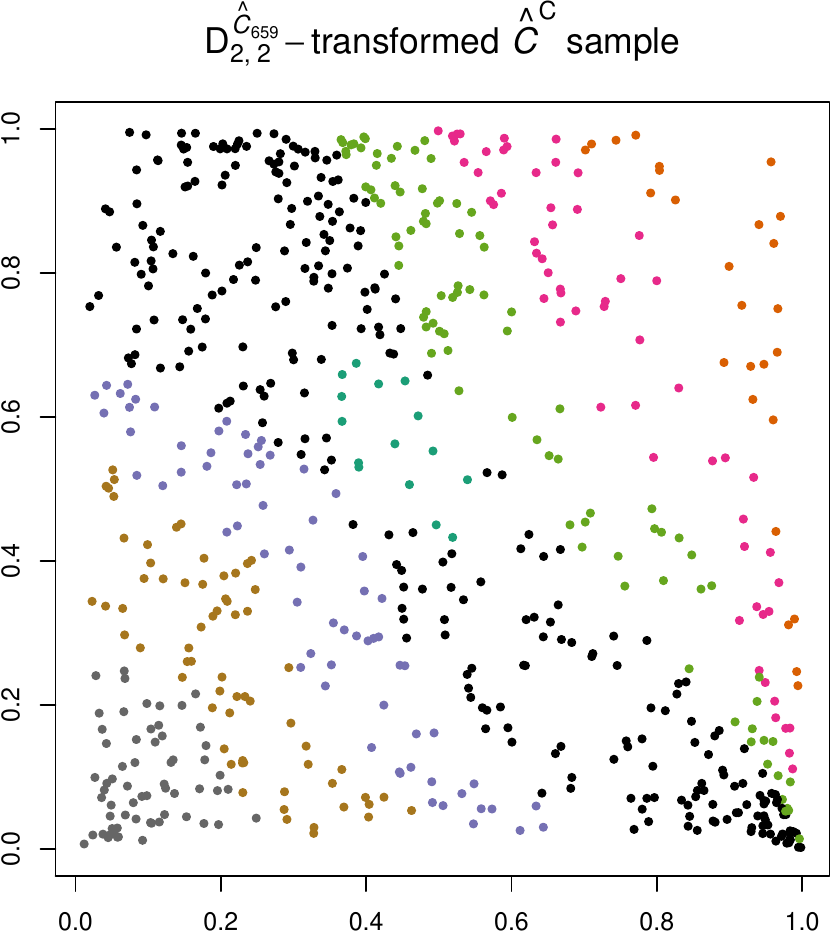}\hfill
  \includegraphics[width=0.16\textwidth]{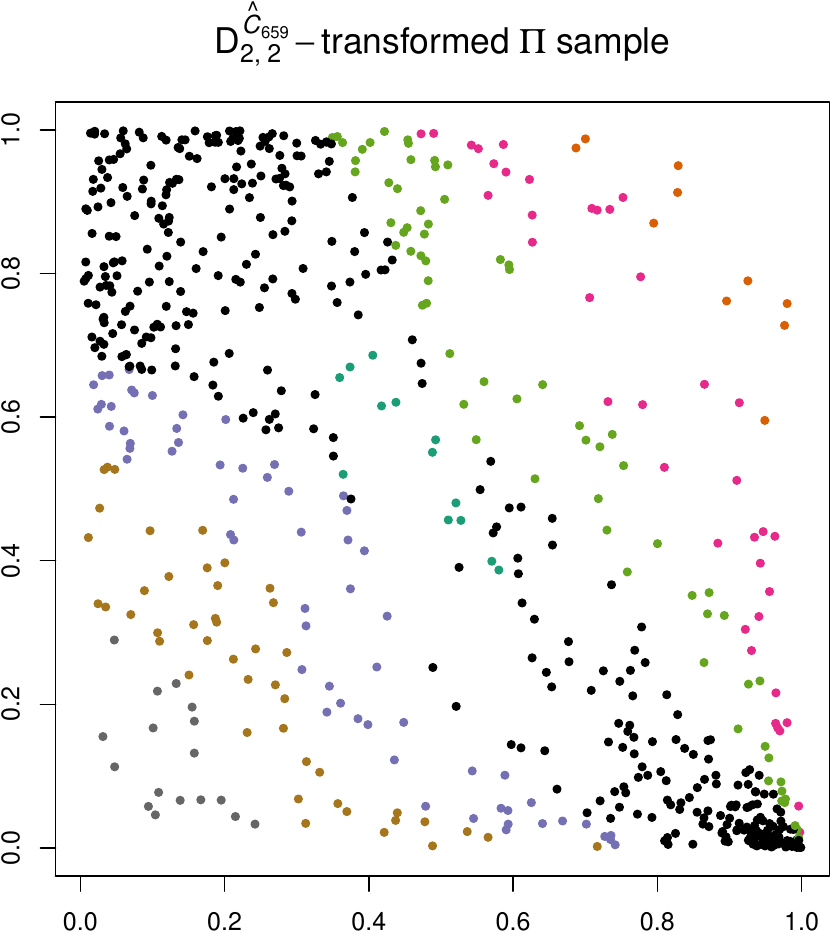}
  \caption{$D^{\hat{C}_n}_{2,2}$-transformed samples of size $\ngen=659$ from
    the bivariate empirical copula $\hat{C}_n$ of the Danube data for $n=659$, a
    fitted Gumbel, $t$, normal, Clayton and the independence copula (top row, from left
    to right), and the same samples with colors (bottom row).}
  \label{fig:scatter:Danube}
\end{figure}

For demonstrating our graphical assessment and selection procedure, we train a
DecoupleNet $D^{\hat{C}_n}_{2,2}$ on the Danube data. We then pass $\ngen=659$
samples from various copulas through $D^{\hat{C}_n}_{d,2}$. As benchmark, we
include $\hat{C}_n$; sampling from $\hat{C}_n$ is done in the usual way, by
drawing pseudo-observations at random with replacement. As candidate models,
we include a Gumbel copula, a normal copula, a $t$ copula, a Clayton copula and
the independence copula.  All parameters of the candidate models were estimated
from the Danube data.  The top row of Figure~\ref{fig:scatter:Danube} shows
scatter plots of the $D^{\hat{C}_n}_{d,2}$-transformed samples for the Danube
data, and the bottom row shows the samples colored with the same color scheme as
before, so, for example, samples with bright colors are decoupled samples from
the joint right tail of the input sample.
The graphical assessment and selection procedure is a bit more challenging to
apply in this case due to the small sample size $n=659$ of the dataset. We
cannot select a clear winner among the fitted Gumbel, $t$ or normal
copulas. Nevertheless, we see (more) non-uniformity for the fitted Clayton and
the independence copula. Similarly for the corresponding colored plots in the
bottom row of Figure~\ref{fig:scatter:Danube}.

\begin{figure}[htbp]
  \centering
  \includegraphics[width=0.48\textwidth]{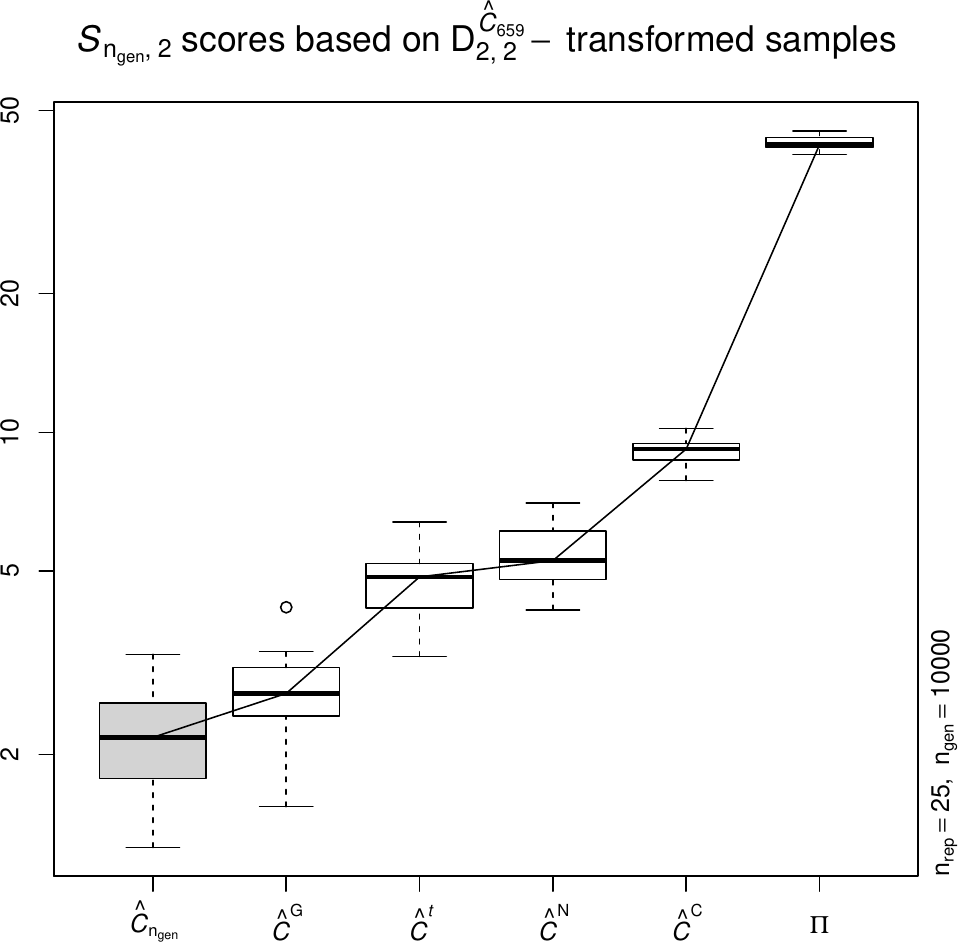}
  \caption{Box plots of CvM scores $S_{\ngen,2}$ based on $\nrep=25$
    $D_{2,2}^{\hat{C}_n}$-transformed samples of size $\ngen=10\,000$ from
    bivariate copulas $\hat{C}_{\ngen}$, $\hat{C}^{\text{G}}$ $\hat{C}^t$,
    $\hat{C}^{\text{N}}$, $\hat{C}^{\text{C}}$ fitted to the Danube data with
    sample size $n=659$. Also included is the independence copula $\Pi$.
    See Algorithm~\ref{algo:model:selection} for details.}
  \label{fig:box:Danube}
\end{figure}

We can additionally compare the numerical summary for the same set of
models. For each one, we generate $\nrep=25$
samples of size $\ngen=10\,000$ and pass them through the DecoupleNet
$D_{2,2}^{\hat{C}_n}$. We then compute the corresponding CvM scores
$S_{\ngen,2}$; see~\eqref{CvM:score}. The resulting box plots are shown in
Figure~\ref{fig:box:Danube}.
The numerical summaries based on the CvM scores here indeed reveal
the fitted Gumbel copula as an adequate dependence model
and best among all candidate copulas.

\subsection{Exchange rate data}
Here we consider two datasets of foreign exchange rates (FX) with the goal of
investigating the dependence for each of these datasets, an important task from
the realm of risk management. The data can be found in the \R\ package
\code{qrmdata}. The first dataset consists of daily exchange rates of Canadian
dollar (CAD), Pound sterling (GBP), Euro (EUR), Swiss Franc (CHF) and Japanese
yen (JPY) with respect to the US dollar (USD). And the second consists of daily
exchange rates of CAD, USD, EUR, CHF, JPY and the Chinese Yuan (CNY) with
respect to the GBP. The considered trading days are from 2000-01-01 to
2015-12-31, resulting in $5844$ five-dimensional ($d=5$) and six-dimensional
($d=6$) observations for the USD and the GBP FX datasets, respectively.
For each of the two
datasets, negative log-returns were formed and deGARCHed; see
\cite{hofertprasadzhu2021} for more details. This leaves us with $n=5843$
observations per dataset.

\begin{figure}[htbp]
  \includegraphics[width=0.16\textwidth]{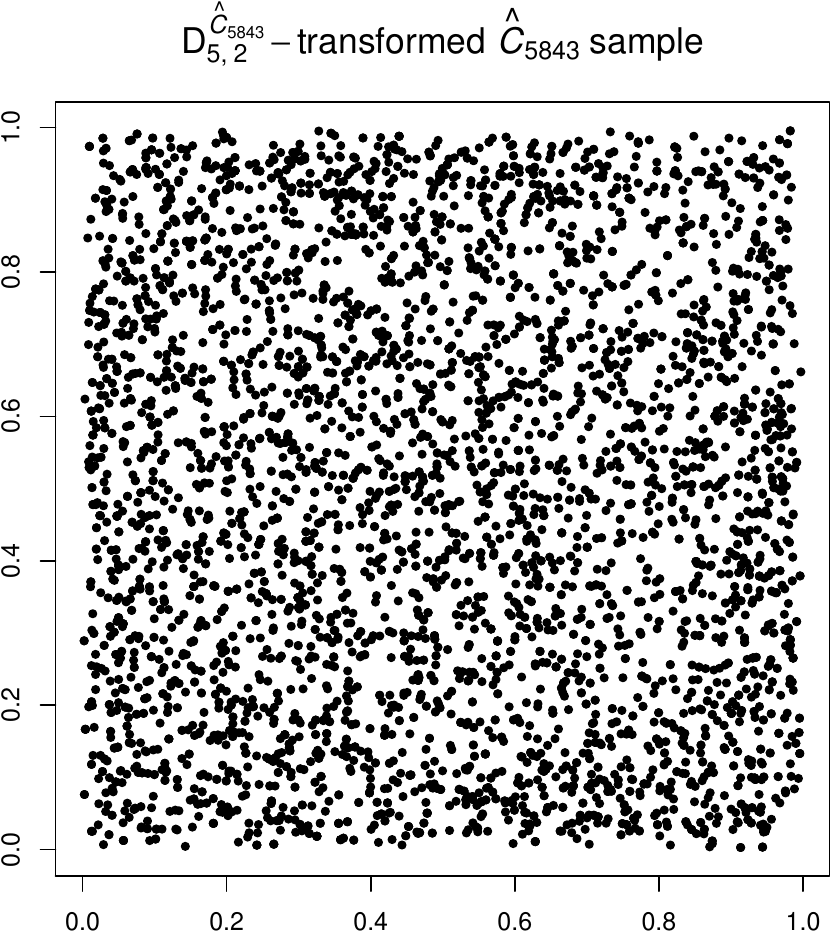}\hfill
  \includegraphics[width=0.16\textwidth]{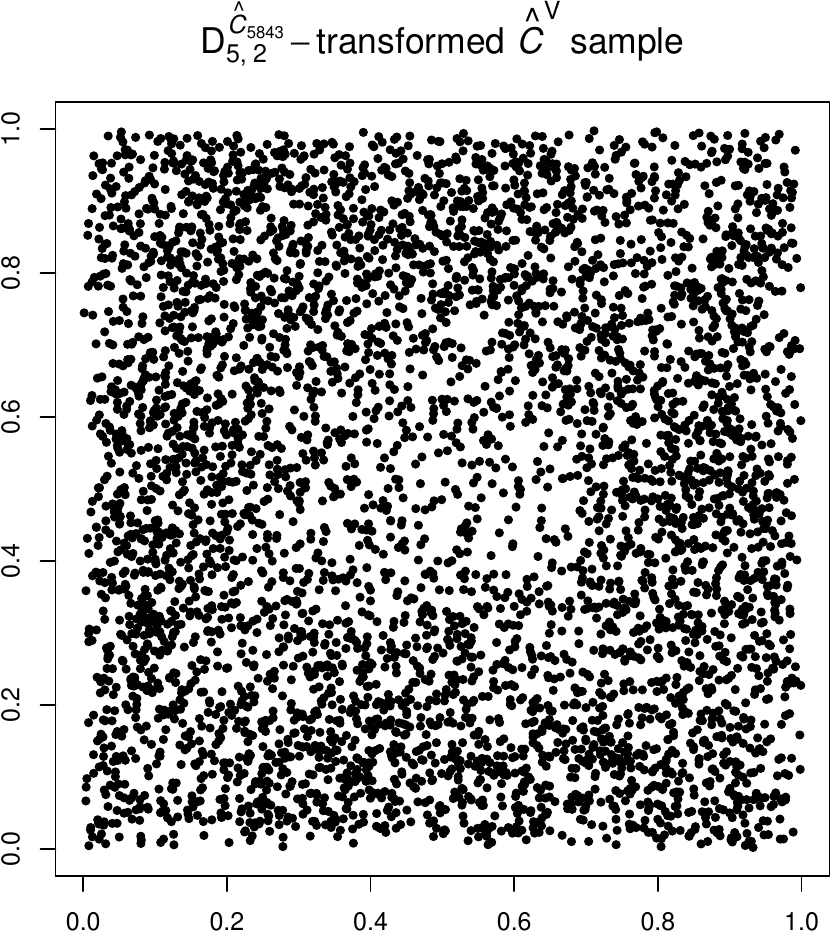}\hfill
  \includegraphics[width=0.16\textwidth]{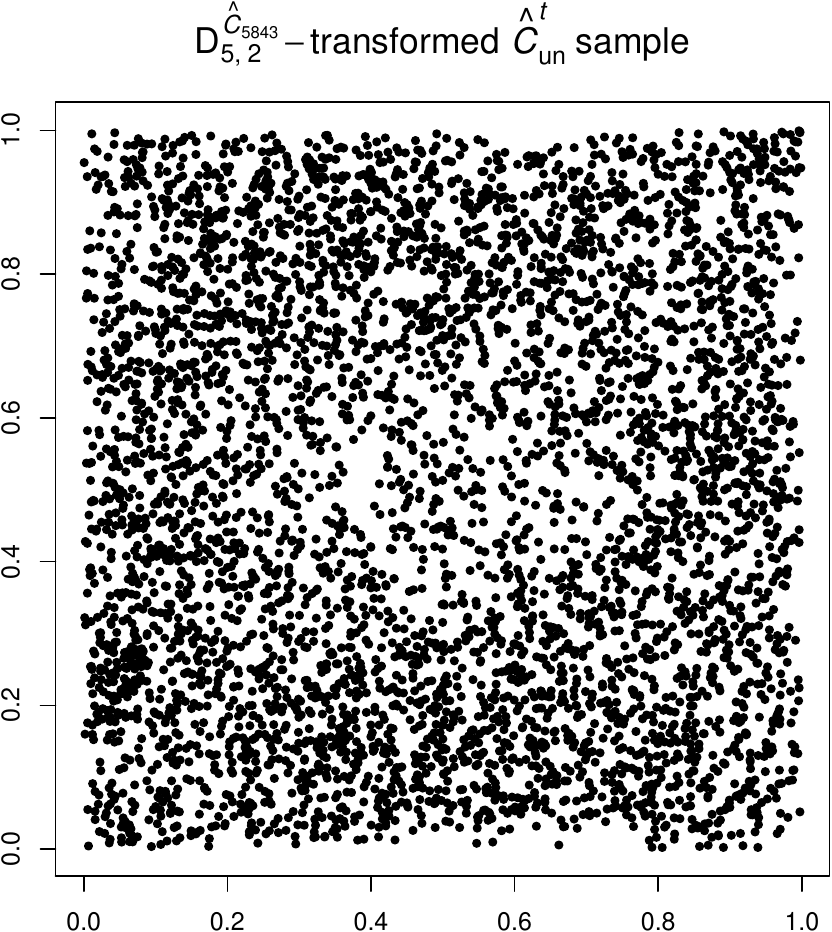}\hfill
  \includegraphics[width=0.16\textwidth]{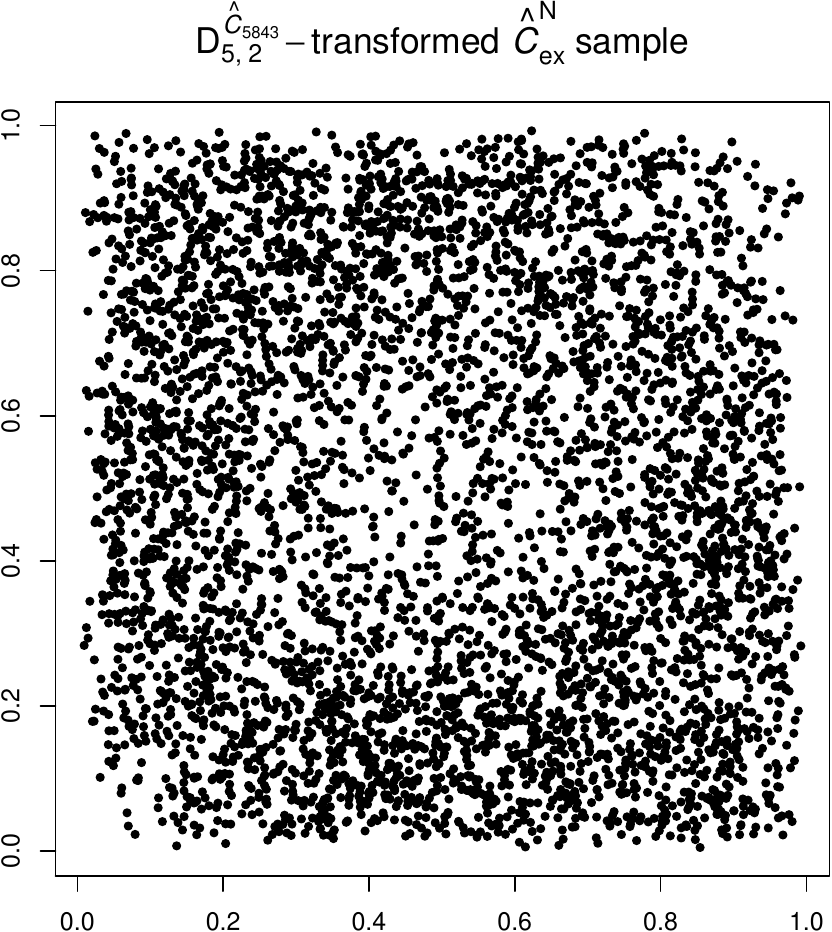}\hfill
  \includegraphics[width=0.16\textwidth]{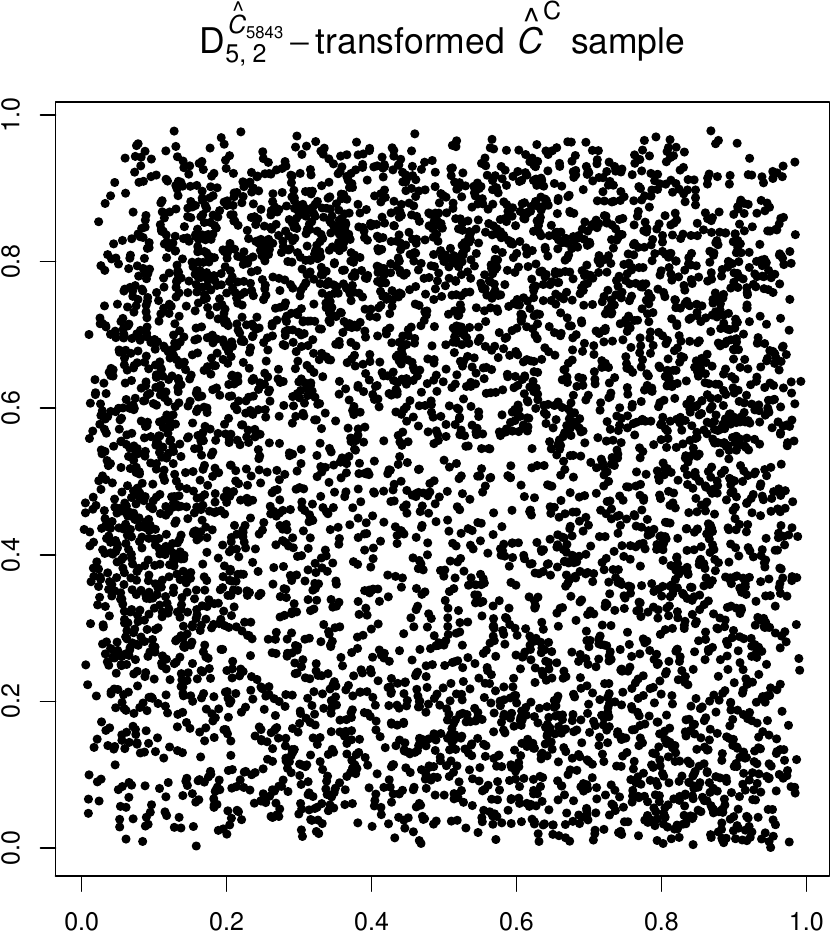}\hfill
  \includegraphics[width=0.16\textwidth]{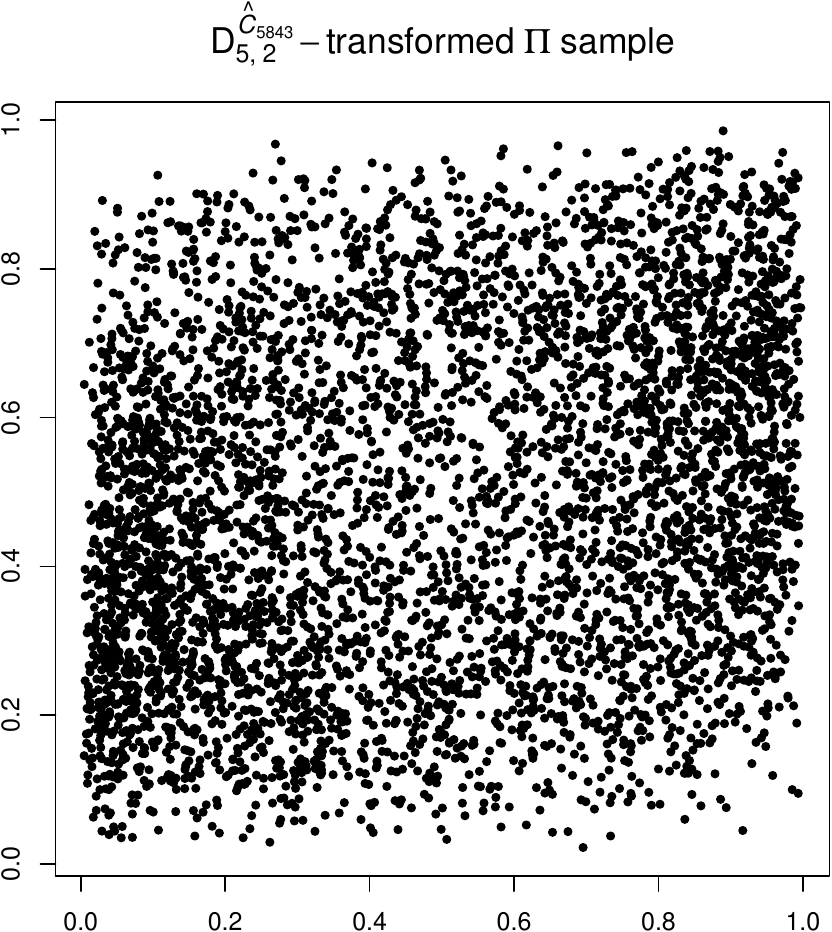}
  \\[2mm]
  \includegraphics[width=0.16\textwidth]{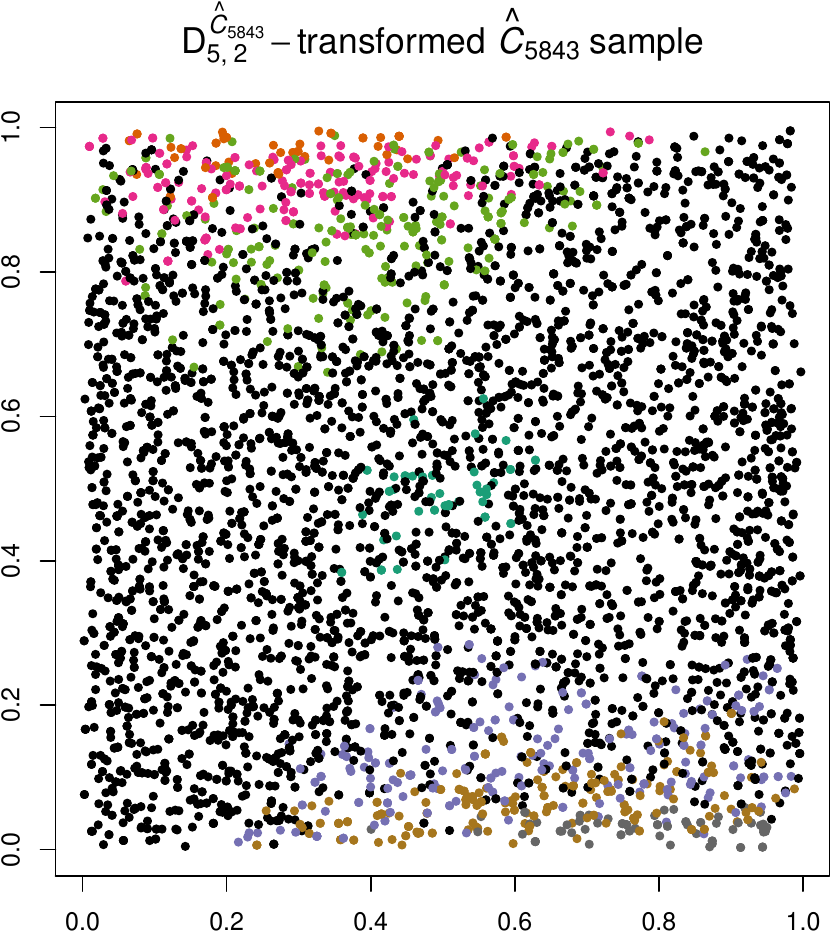}\hfill
  \includegraphics[width=0.16\textwidth]{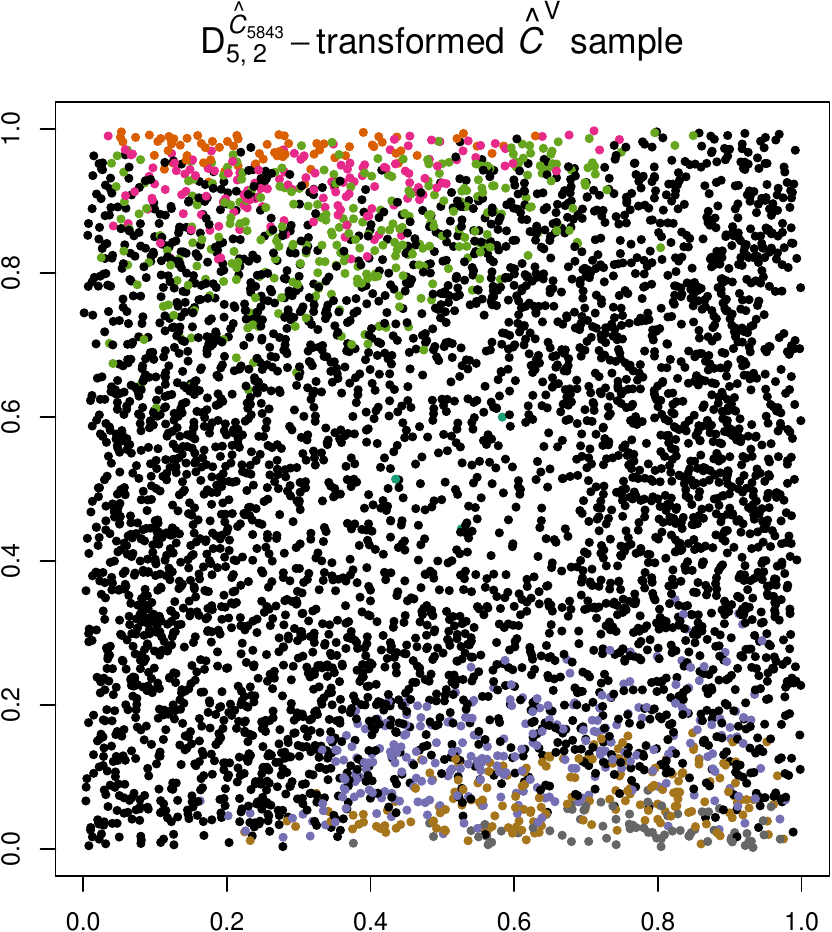}\hfill
  \includegraphics[width=0.16\textwidth]{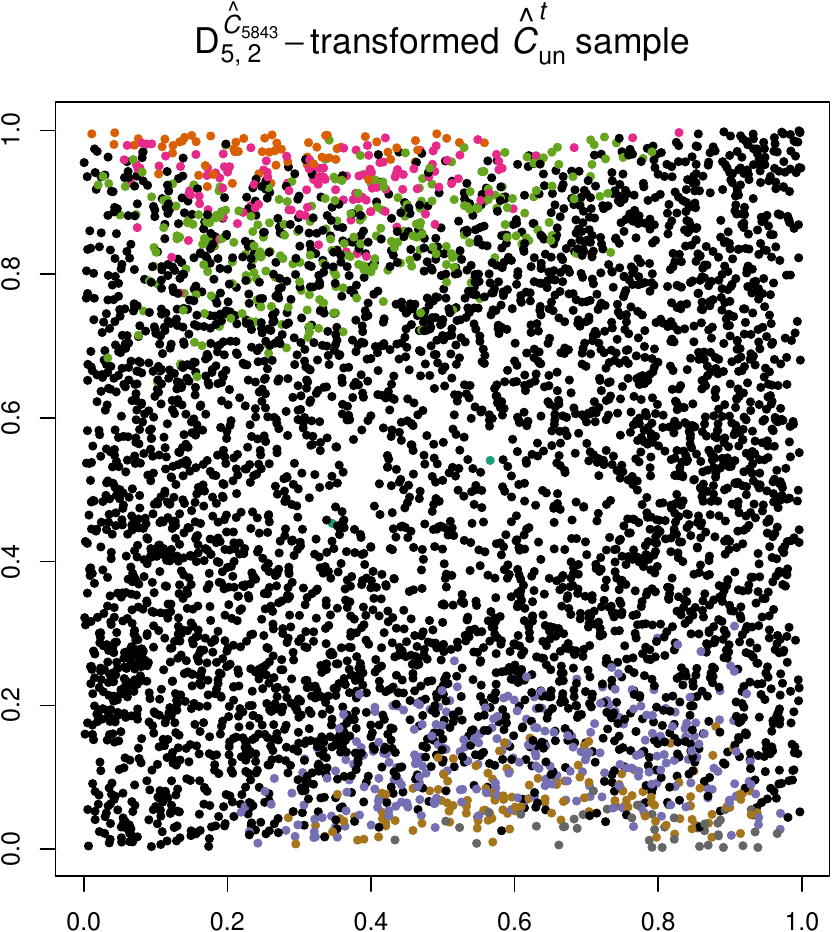}\hfill
  \includegraphics[width=0.16\textwidth]{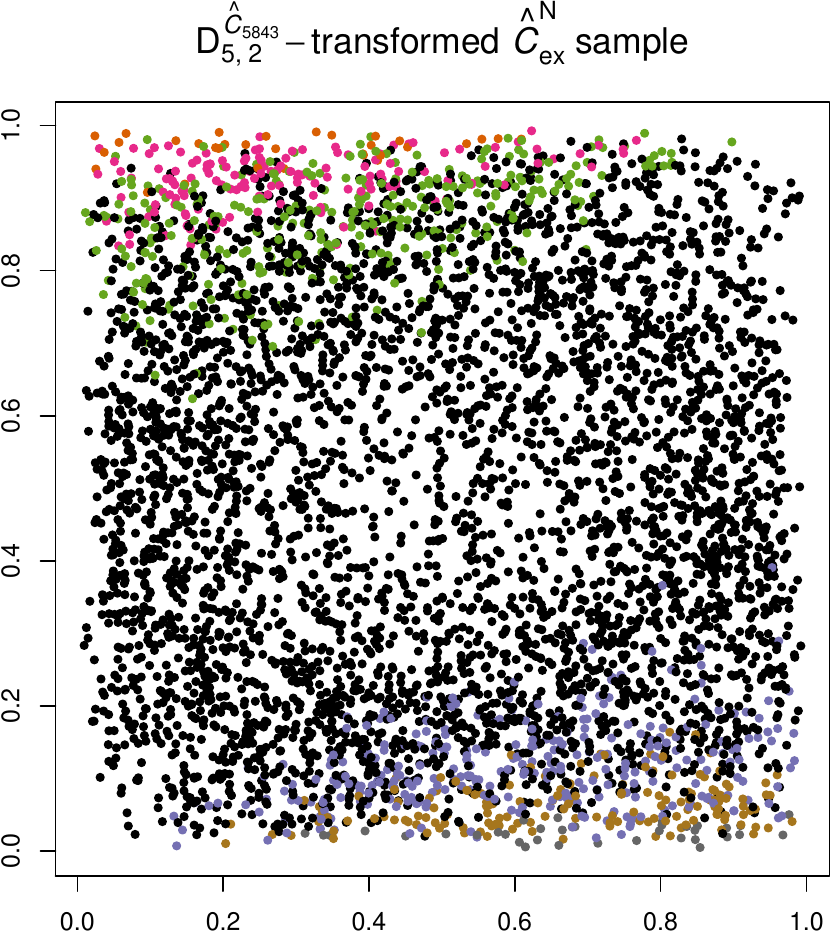}\hfill
  \includegraphics[width=0.16\textwidth]{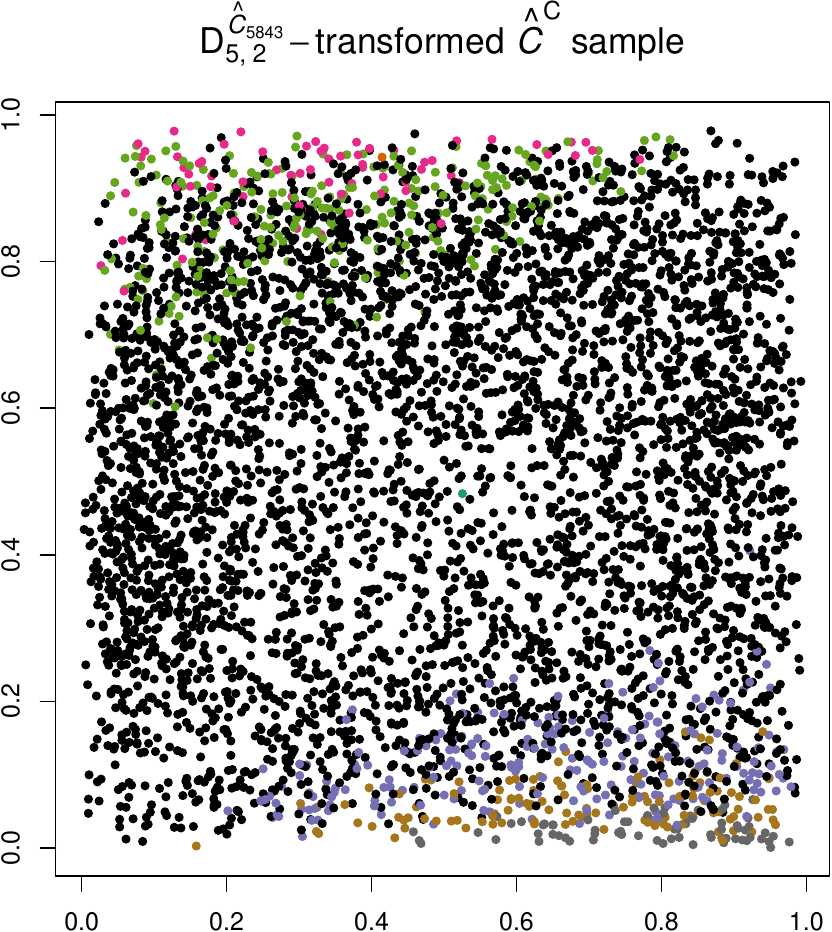}\hfill
  \includegraphics[width=0.16\textwidth]{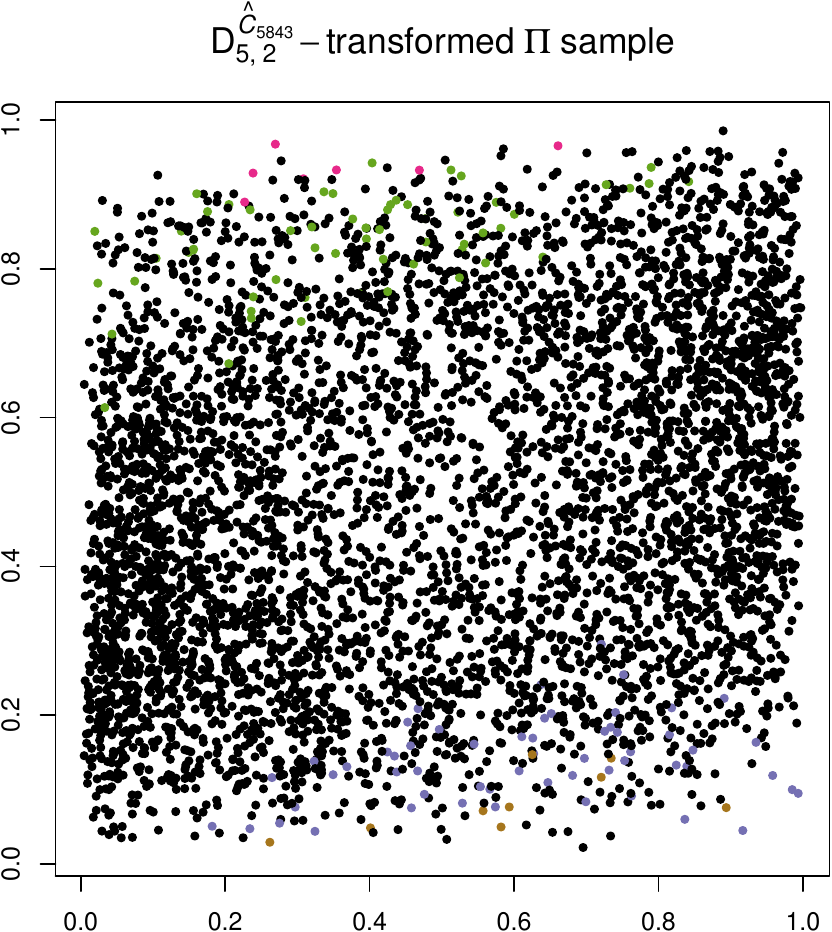}
  \\[6mm]
  \includegraphics[width=0.16\textwidth]{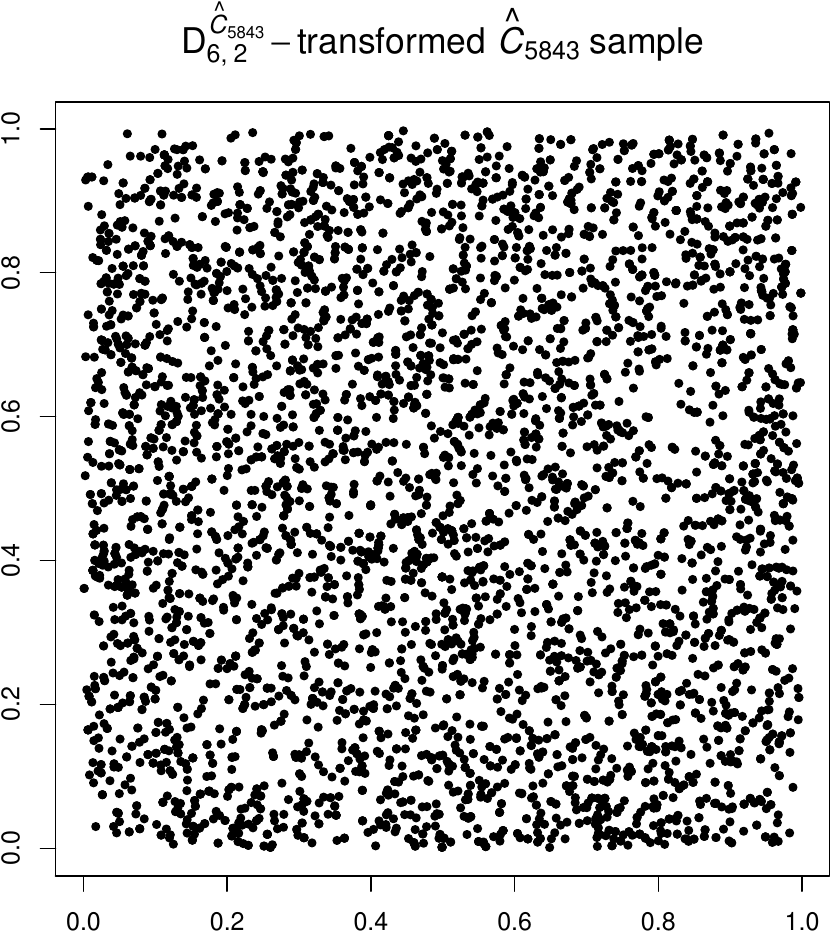}\hfill
  \includegraphics[width=0.16\textwidth]{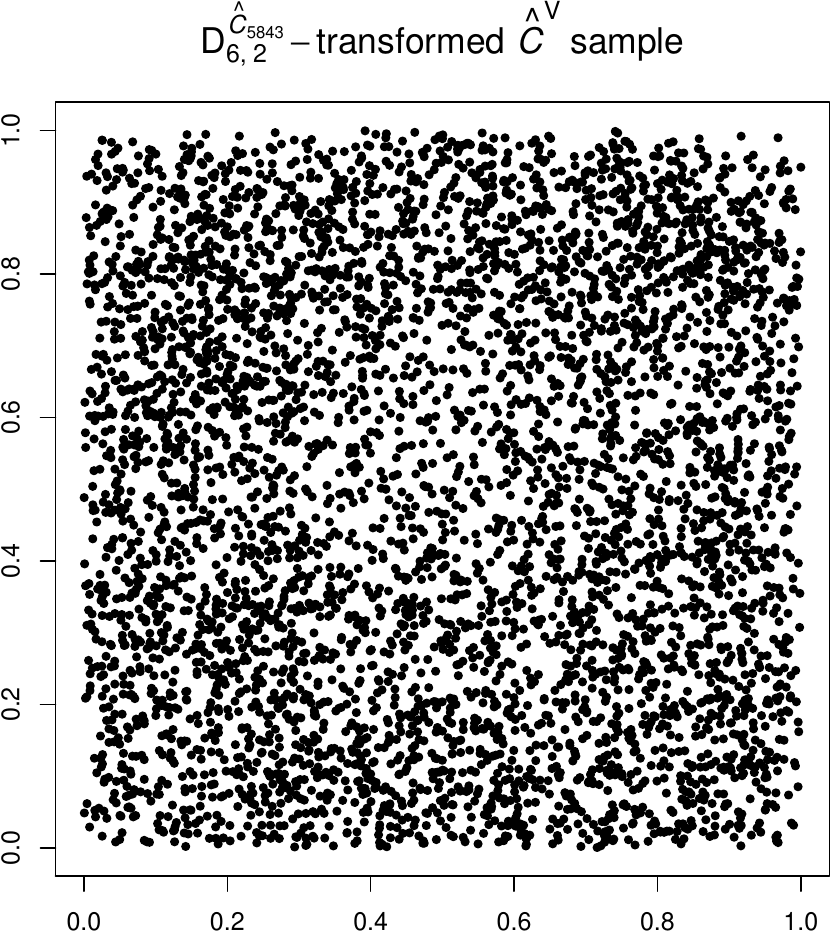}\hfill
  \includegraphics[width=0.16\textwidth]{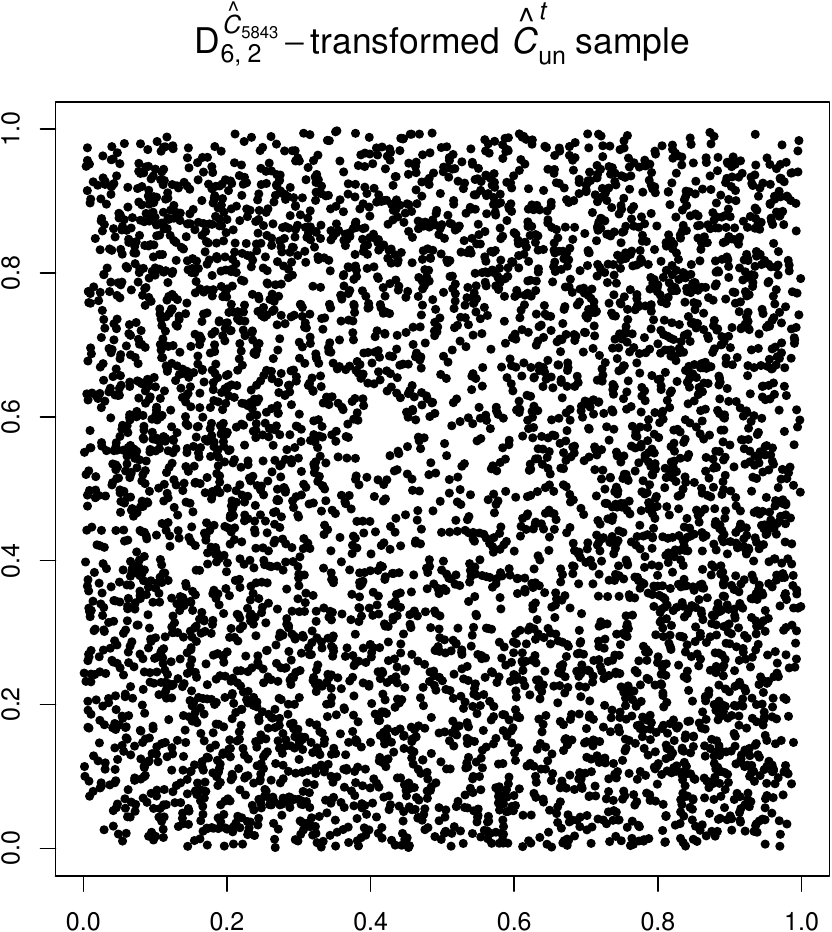}\hfill
  \includegraphics[width=0.16\textwidth]{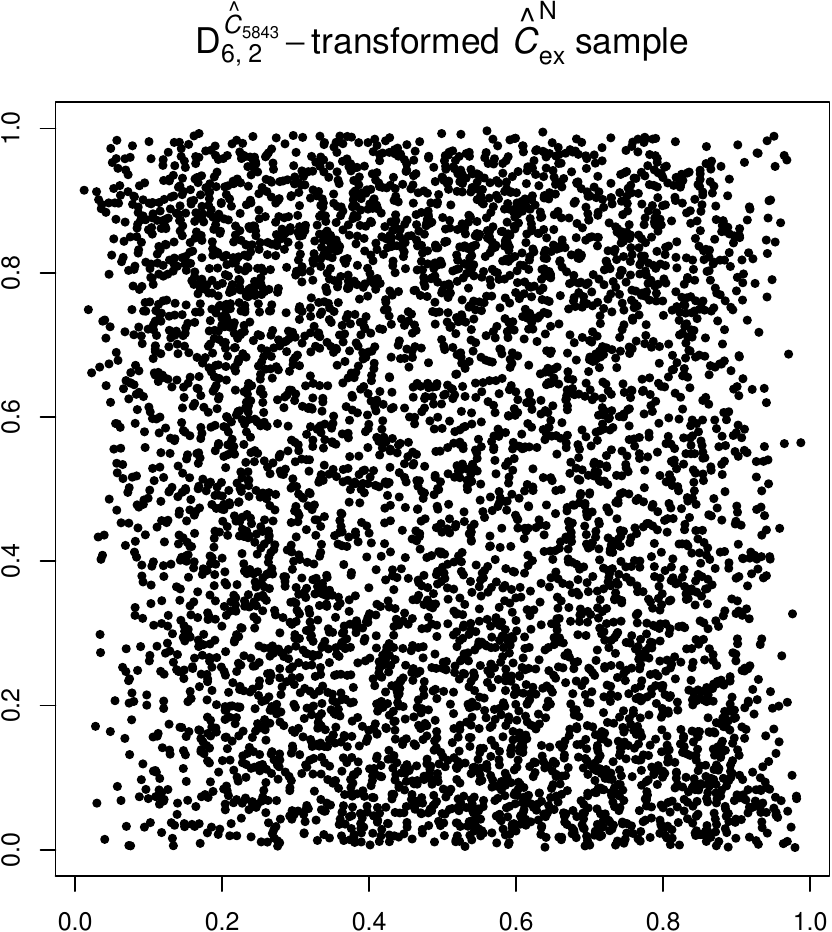}\hfill
  \includegraphics[width=0.16\textwidth]{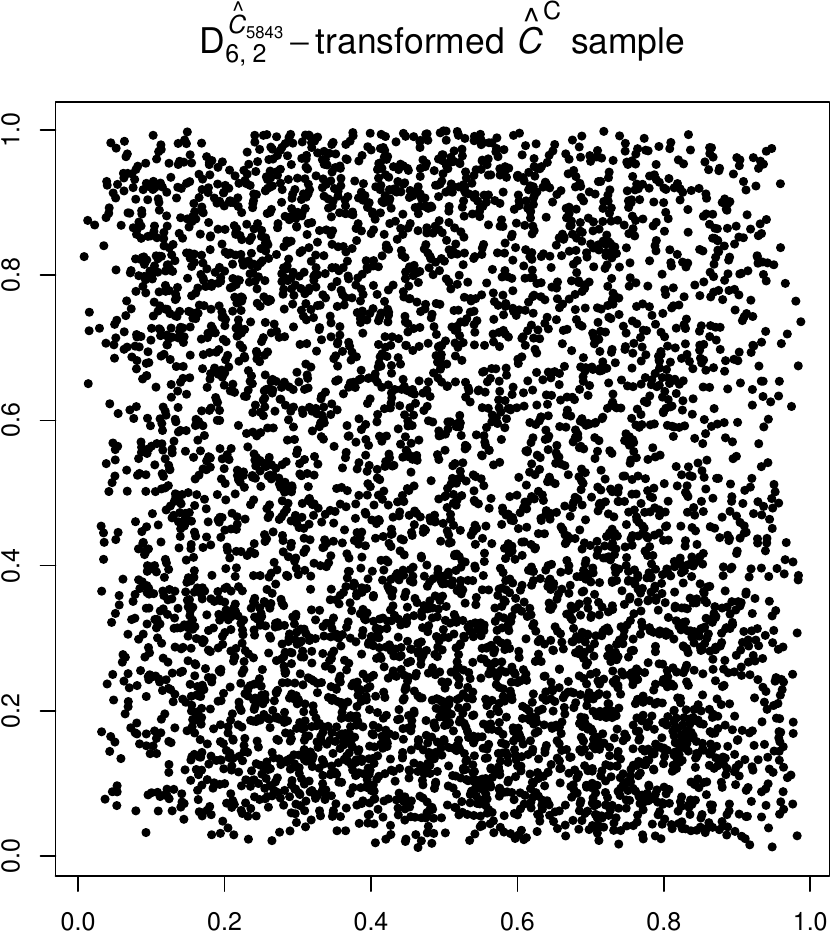}\hfill
  \includegraphics[width=0.16\textwidth]{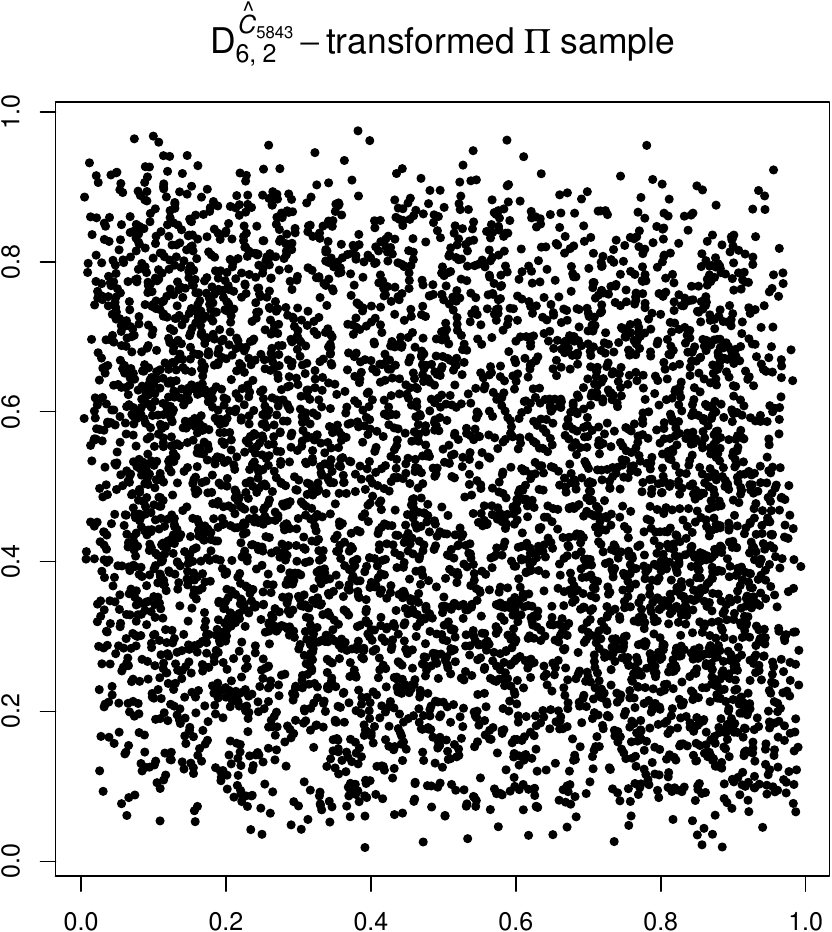}
  \\[2mm]
  \includegraphics[width=0.16\textwidth]{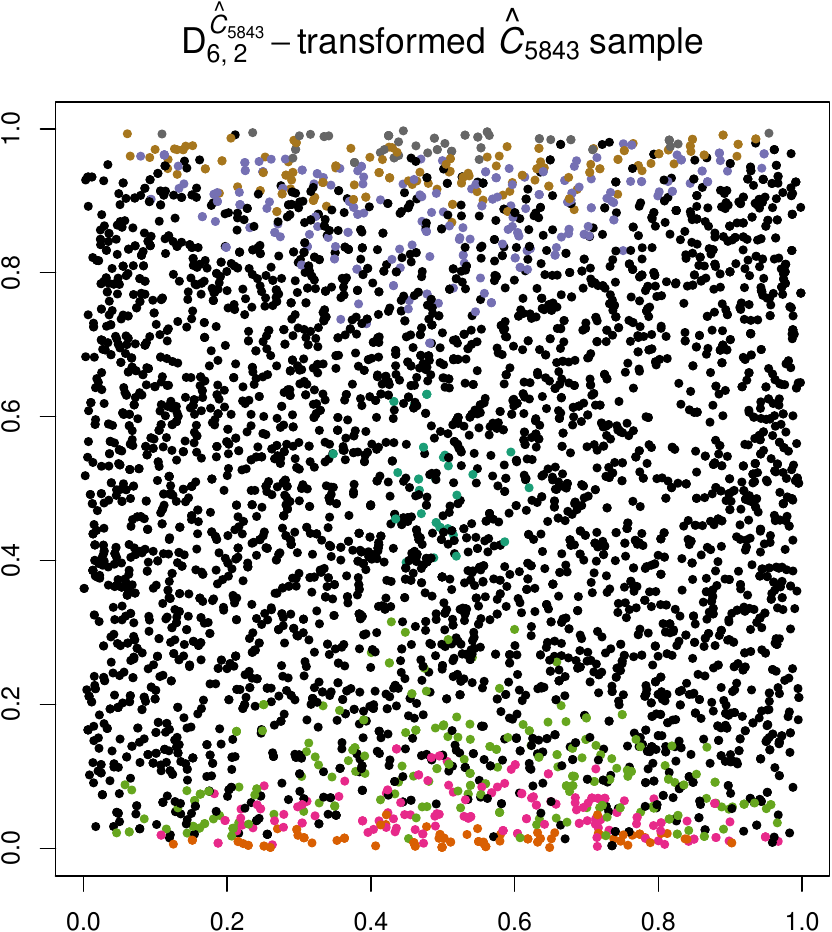}\hfill
  \includegraphics[width=0.16\textwidth]{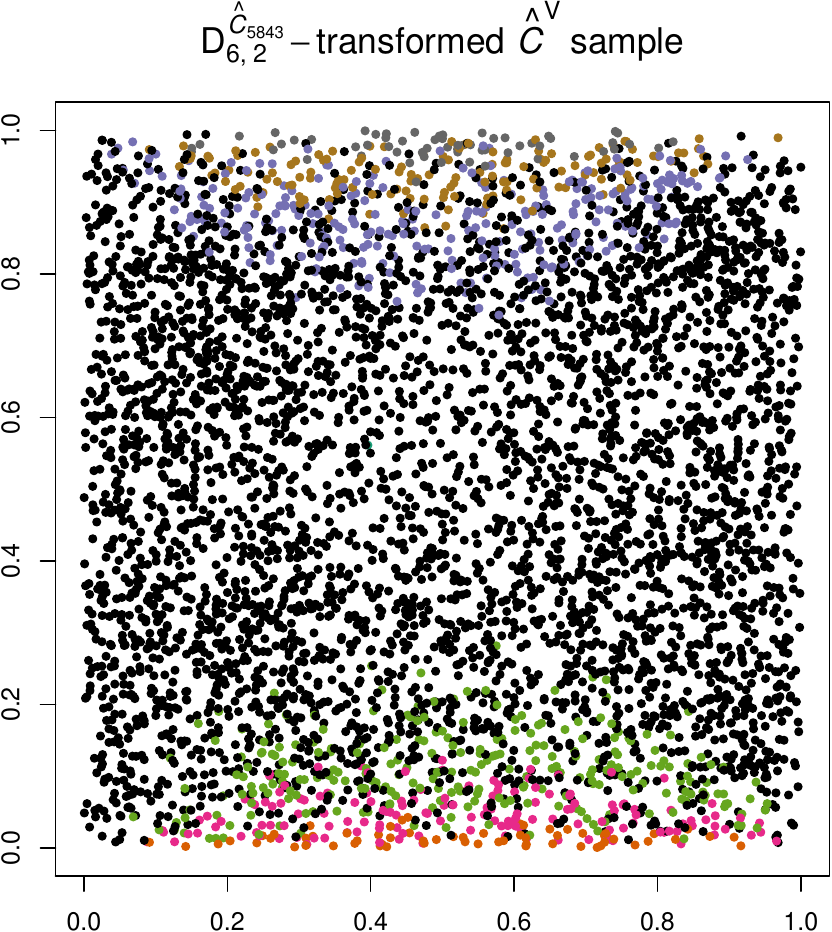}\hfill
  \includegraphics[width=0.16\textwidth]{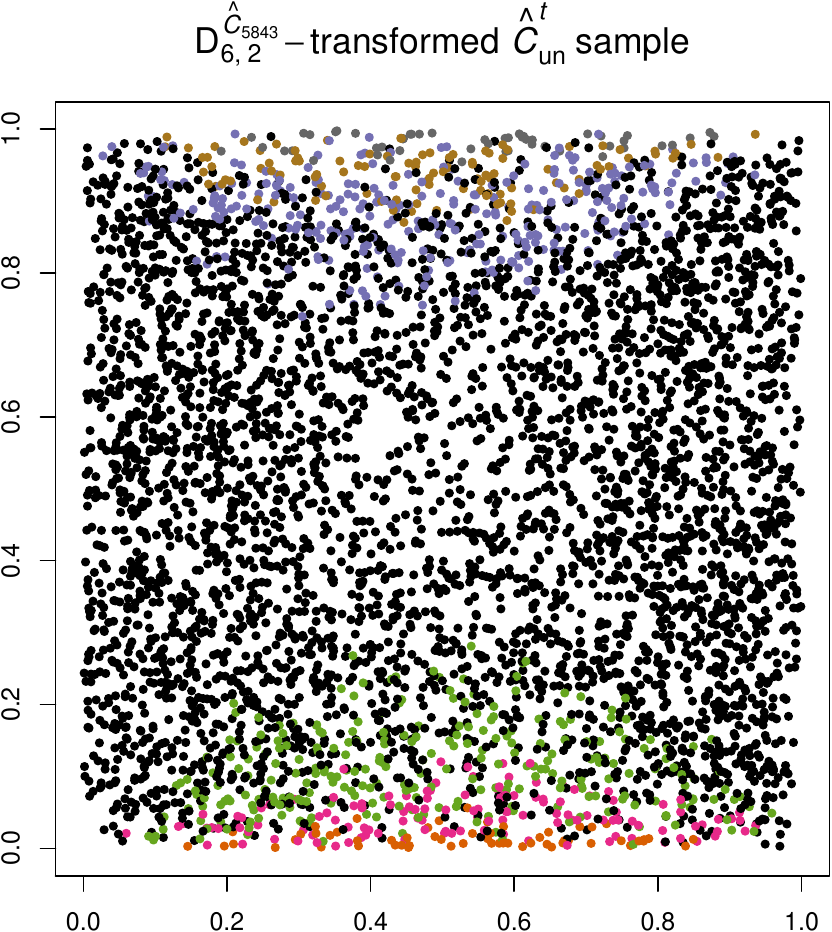}\hfill
  \includegraphics[width=0.16\textwidth]{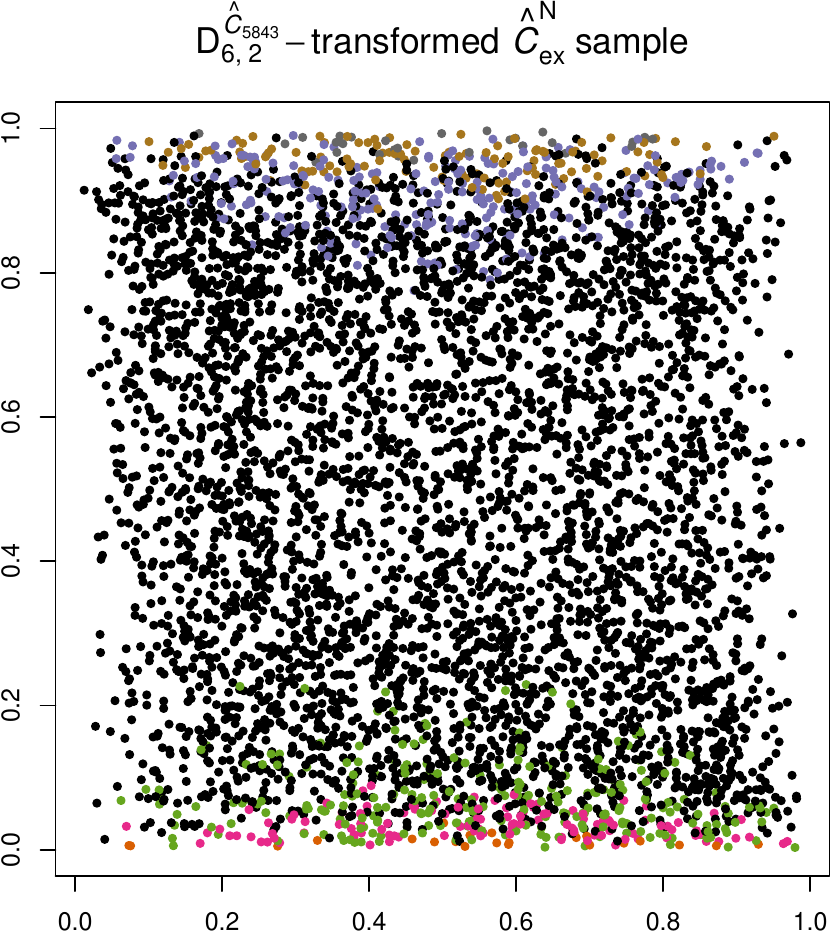}\hfill
  \includegraphics[width=0.16\textwidth]{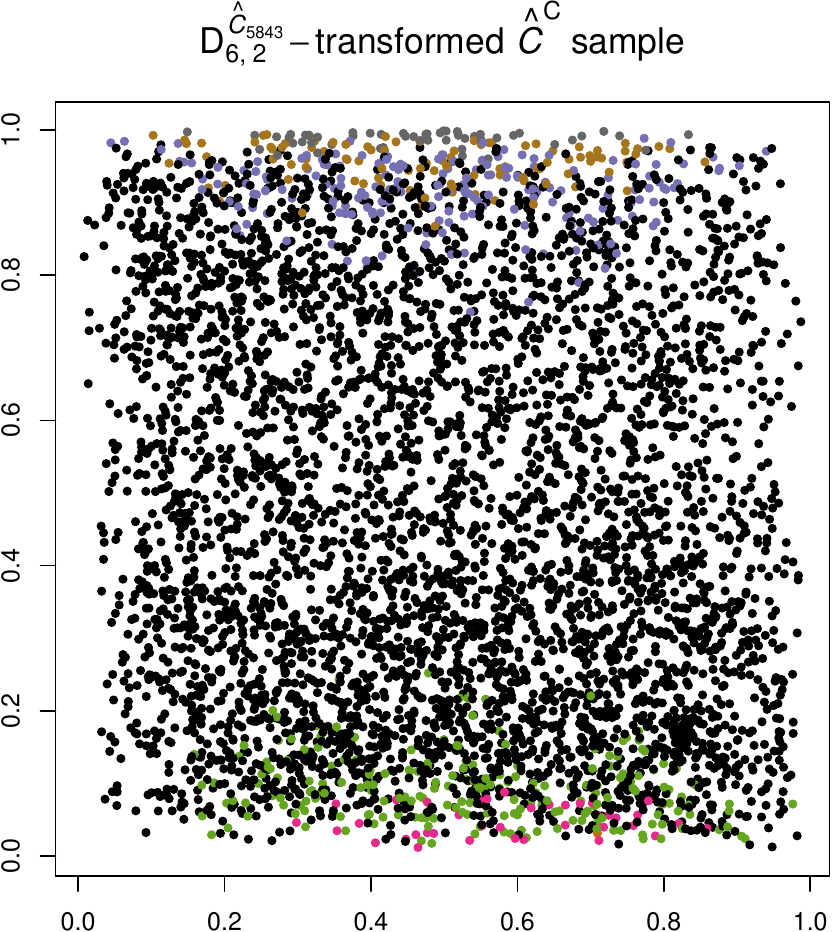}\hfill
  \includegraphics[width=0.16\textwidth]{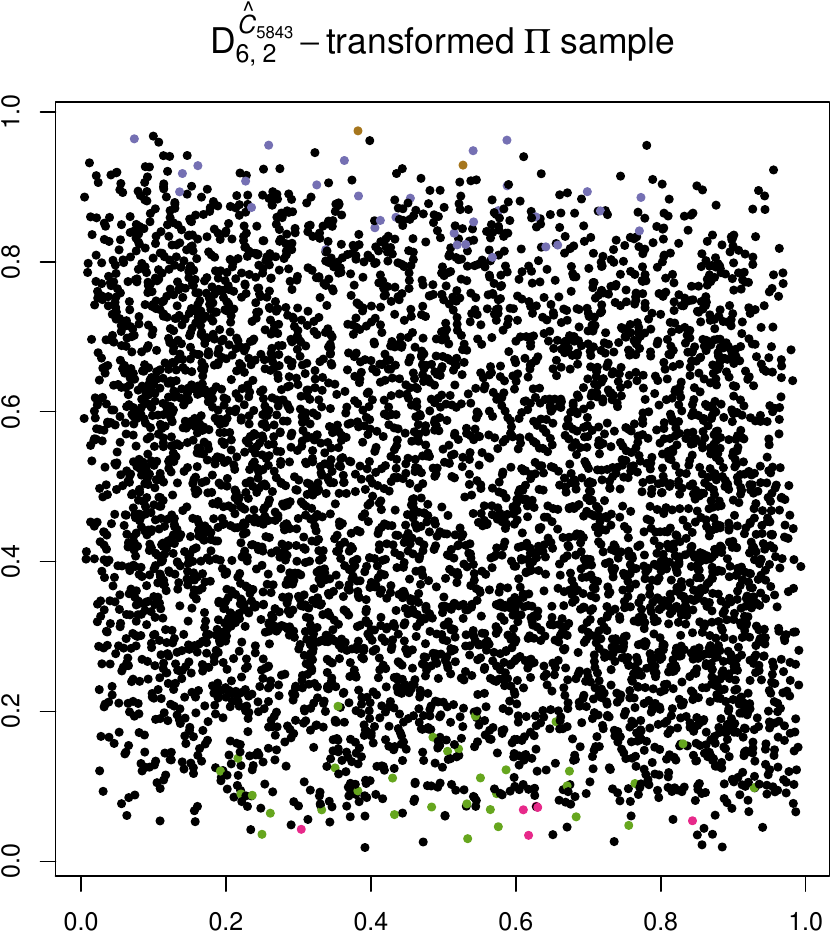}
  \caption{First row: $D^{\hat{C}_n}_{d,2}$-transformed samples of size $\ngen=5000$ from
    the $d$-dimensional (with $d=5$) empirical copula $\hat{C}_n$ of the FX USD
    data for $n=5843$, a fitted vine, unstructured $t$, exchangeable normal, Clayton and the
    independence copula (from left to right). Second row: the same samples
    with colors. Third and forth rows: same as first two rows, but
    now for the $d$-dimensional (with $d=6$) GBP FX data. The parameters of the
    candidate copulas in the center four columns were estimated.}
  \label{fig:scatter:FX}
\end{figure}

For demonstrating our graphical assessment and selection procedure, we consider
the pseudo-observations with corresponding $d$-dimensional empirical copula
$\hat{C}_n$ for both datasets. For each dataset, we trained a DecoupleNet
$D^{\hat{C}_n}_{d,2}$ on these pseudo-observations. We then pass $\ngen=5000$
samples from various copulas through $D^{\hat{C}_n}_{d,2}$. As benchmark, we
include $\hat{C}_n$. And as candidate models, we include
a vine copula, a $t$ copula with unstructured correlation matrix, an exchangeable normal
copula with homogeneous correlation matrix, a
Clayton copula and the independence copula. All parameters of the candidate
models were estimated from the respective pseudo-observations. The first row of
Figure~\ref{fig:scatter:FX} shows scatter plots of the
$D^{\hat{C}_n}_{d,2}$-transformed samples for the USD FX data ($d=5$), and the
second row shows the samples colored with the same color scheme as before.
Rows three and four of Figure~\ref{fig:scatter:FX} show similar plots as rows
one and two, respectively, but now for the GBP FX data ($d=6$). The first column
shows uniformity of the DecoupleNet-transformed samples of the empirical copula
$\hat{C}_n$, so training of the two DecoupleNets worked well. The samples
corresponding to all candidate models show non-uniformity, though, so none of
them seems to fit the respective dataset well, supporting our statement in the
first paragraph of Section~\ref{sec:intro}. From the (barely visible) mid-range
colored samples in row two and four, we can identify that none of the candidate
models fits well in the body of the underlying $d$-dimensional distribution.
Judging from the fits in the joint right tail (bright colors), both the fitted
vine and the fitted $t$ copulas seem adequate for capturing the dependence in
this region.

\begin{figure}[htbp]
  \includegraphics[width=0.48\textwidth]{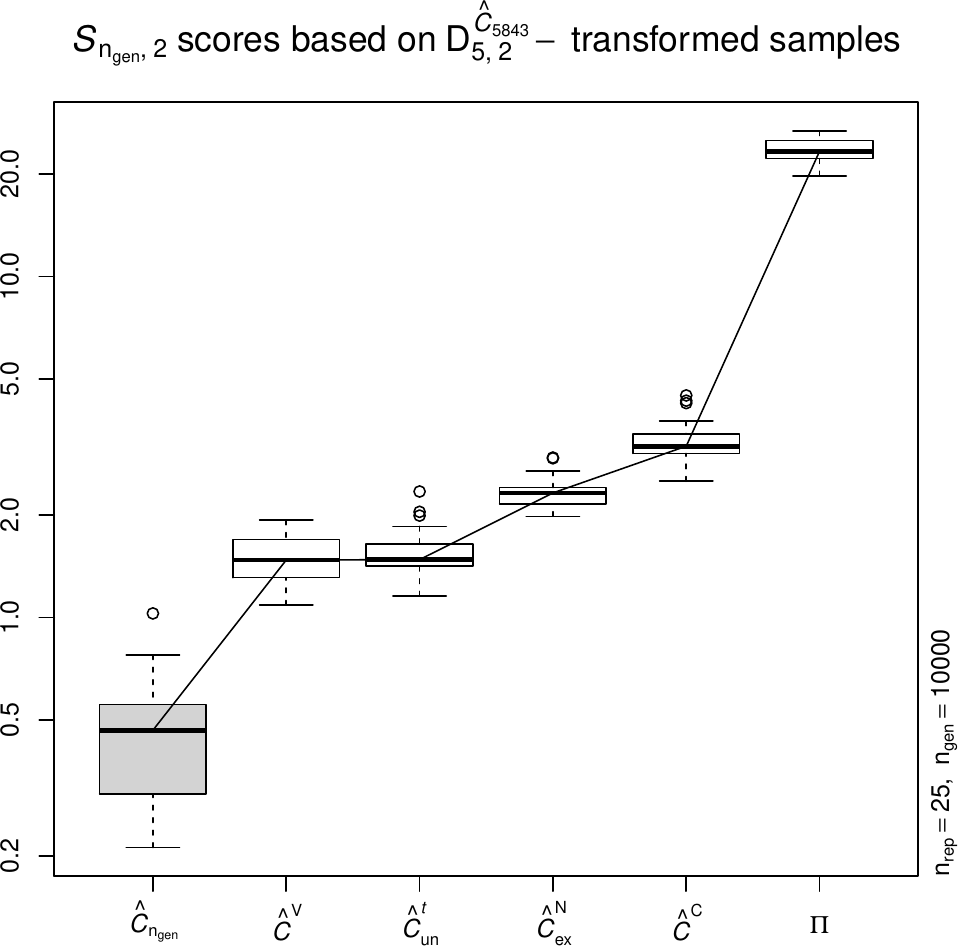}\hfill
  \includegraphics[width=0.48\textwidth]{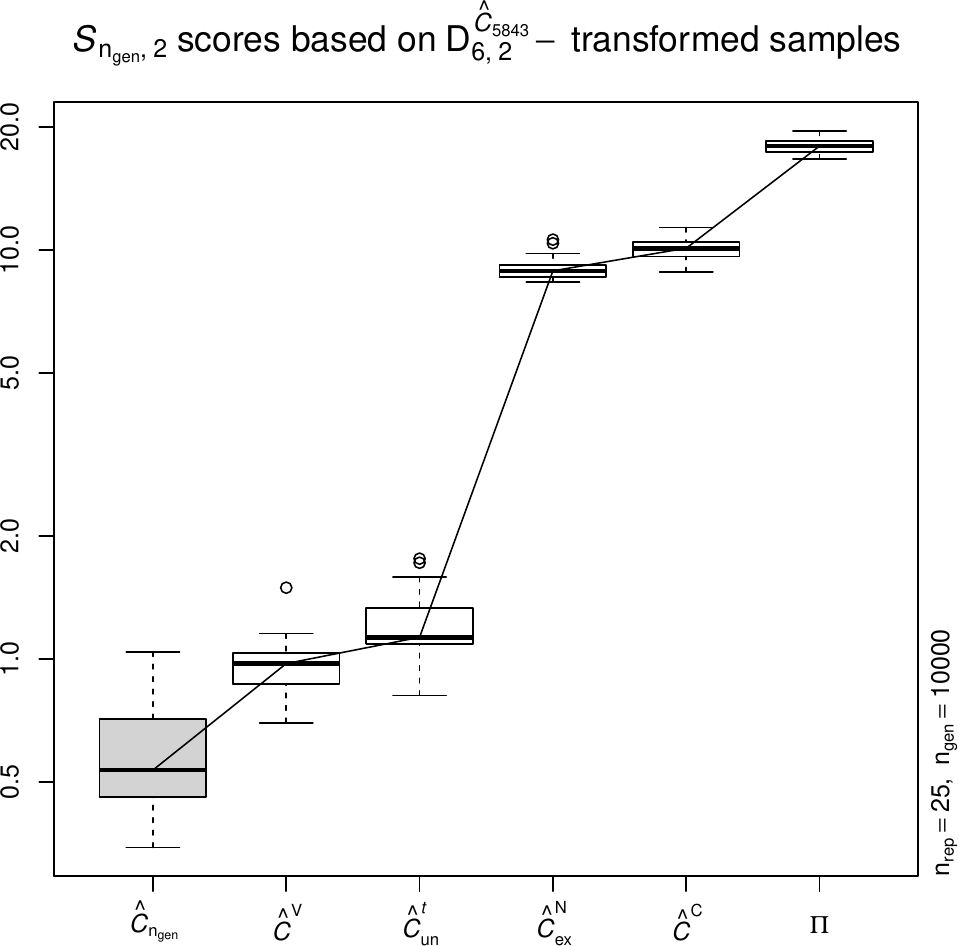}
  \caption{Box plots of CvM scores $S_{\ngen,2}$ based on $\nrep=25$
    $D_{d,2}^{\hat{C}_n}$-transformed samples of size $\ngen=10\,000$ from
    $d$-dimensional copulas $\hat{C}_{\ngen}$, $\hat{C}^{\text{V}}$,
    $\hat{C}^t_{\text{un}}$, $\hat{C}^{\text{N}}_{\text{ex}}$,
    $\hat{C}^{\text{C}}$ fitted to the FX USD data for $d=5$ (left) and the FX
    GBP data for $d=6$ (right) with sample size $n=5843$. Also included is the
    independence copula $\Pi$. See Algorithm~\ref{algo:model:selection} for
    details.}
  \label{fig:box:FX}
\end{figure}

We can also compare the numerical summary for all candidate models as per Section~\ref{sec:num:approach}.
For each of the datasets and models considered, we generate $\nrep=25$
samples of size $\ngen=10\,000$ and pass them through the respective DecoupleNet
$D_{5,2}^{\hat{C}_n}$ (for the USD FX data) or $D_{6,2}^{\hat{C}_n}$ (for the
GBP FX data). We then compute the corresponding CvM scores $S_{\ngen,2}$;
see~\eqref{CvM:score}. The resulting box plots are shown in
Figure~\ref{fig:box:FX}.
Our conclusion from the numerical summary is the same as from the graphical
approach. We see from the box plots that none of the candidate models are
particularly good for the respective data set, with vine and $t$ copulas
performing best on both the USD and the GBP FX data.

\section{Conclusion}\label{sec:concl}
We introduced DecoupleNets for dependence model assessment and selection.  A
DecoupleNet is a neural network based transformation of a random vector from a
copula to a random vector from a standard uniform distribution. A DecoupleNet
can be trained on samples from a known copula or, more importantly, on
pseudo-observations from a given multivariate dataset for which no copula is
known. A candidate copula for the given dataset can then be assessed by
computing a DecoupleNet-transformed sample from the candidate model and
assessing its (non-)uniformity. Model selection can be done by comparing the
(non-)uniformity of DecoupleNet-transformed samples from the candidate models
and selecting the one producing the least non-uniform output. For both tasks,
the flexibility of neural networks is a main advantage and allows DecoupleNets
to be trained on and applied to any copula sample. Another advantage is that
DecoupleNets can map to the (bivariate) unit square, which is computationally
advantageous and, especially, allows for a graphical approach to assess and select dependence
models. In particular, coloring input samples and
corresponding DecoupleNet-transformed output samples even allows one to assess
and select dependence models based on particular regions of interest, a
  fact particularly important for practical applications in which dependence
  models often turn out to be inadequate as models overall, but are only of
  interest in specific regions such as the tails.

\Urlmuskip=0mu plus 1mu\relax%
\printbibliography[heading=bibintoc]
\end{document}

%
%
%
%
